\documentclass[journal,comsoc]{IEEEtran}
\usepackage{amsmath,amsfonts}
\usepackage{algorithmic}
\usepackage[linesnumbered,ruled,vlined]{algorithm2e}
\usepackage{array}
\usepackage{textcomp}
\usepackage{stfloats}
\usepackage{url}
\usepackage{verbatim}
\usepackage{graphicx}
\usepackage{mathtools}
\usepackage{cite}
\usepackage{subcaption}
\usepackage{multirow}
\usepackage{bbm}
\usepackage{caption}
\usepackage{xcolor}
\allowdisplaybreaks

\newtheorem{assumption}{\textbf{Assumption}}
\newtheorem{theorem}{\textbf{Theorem}}
\newtheorem{proposition}{\textbf{Proposition}}
\newtheorem{definition}{\textbf{Definition}}
\newtheorem{lemma}{\textbf{Lemma}}

\newtheorem{remark}{\textbf{Remark}}

\begin{document}

\title{Towards Communication-efficient Federated Learning via Sparse and Aligned Adaptive Optimization}

\author{Xiumei~Deng,~Jun~Li,~\IEEEmembership{Fellow,~IEEE},~Kang~Wei,~\IEEEmembership{Member,~IEEE},~Long~Shi,~\IEEEmembership{Senior Member,~IEEE},~Zehui~Xiong,~\IEEEmembership{Senior Member,~IEEE},~Ming~Ding,~\IEEEmembership{Senior Member,~IEEE},~Wen~Chen,~\IEEEmembership{Senior Member,~IEEE},~Shi~Jin,~\IEEEmembership{Fellow,~IEEE},~and~H.~Vincent~Poor,~\IEEEmembership{Life Fellow,~IEEE}

\thanks{X. Deng is with the Pillar of Information Systems Technology and Design, Singapore University of Technology and Design, Singapore (e-mail: xiumei\_deng@sutd.edu.sg).}
\thanks{J. Li is with the School of Information Science and Engineering, K. Wei is with the School of Cyber Science and Engineering, and S. Jin is with the National Mobile Communications Research Laboratory, Southeast University, Nanjing 210096, China (e-mail: \{jun.li, kang.wei, jinshi\}@seu.edu.cn).}
\thanks{L. Shi is with School of Electronic and Optical Engineering, Nanjing University of Science and Technology, Nanjing 210094, China (e-mail: slong1007@gmail.com).}
\thanks{Z. Xiong is with the School of Electronics, Electrical Engineering and Computer Science (EEECS), Queen's University Belfast, Belfast, BT7 1NN, U.K. (z.xiong@qub.ac.uk).}
\thanks{M. Ding is with Data61, CSIRO, Sydney, NSW 2015, Australia (e-mail: ming.ding@data61.csiro.au).}
\thanks{W. Chen is with Department of Electronics Engineering, Shanghai Jiao Tong University, Shanghai 200240, China (e-mail: wenchen@sjtu.edu.cn).}
\thanks{H. V. Poor is with Department of Electrical and Computer Engineering, Princeton University, NJ 08544, USA (e-mail: poor@princeton.edu).}
%\thanks{Manuscript received April 19, 2021; revised August 16, 2021.}
}

\markboth{Journal of \LaTeX\ Class Files,~Vol.~14, No.~8, August~2021}%
{Shell \MakeLowercase{\textit{et al.}}: A Sample Article Using IEEEtran.cls for IEEE Journals}


\maketitle

\begin{abstract}
Adaptive moment estimation (Adam), as a Stochastic Gradient Descent (SGD) variant, has gained widespread popularity in federated learning (FL) due to its fast convergence. However, federated Adam (FedAdam) algorithms suffer from a threefold increase in uplink communication overhead compared to federated SGD (FedSGD) algorithms, which arises from the necessity to transmit both local model updates and first and second moment estimates from distributed devices to the centralized server for aggregation. Driven by this issue, we propose a novel sparse FedAdam algorithm called FedAdam-SSM, wherein distributed devices sparsify the updates of local model parameters and moment estimates and subsequently upload the sparse representations to the centralized server. To further reduce the communication overhead, the updates of local model parameters and moment estimates incorporate a shared sparse mask (SSM) into the sparsification process, eliminating the need for three separate sparse masks. Theoretically, we develop an upper bound on the deviation between the local model trained by FedAdam-SSM and the target model trained by centralized Adam, which is related to sparsification error and imbalanced data distribution. By minimizing the deviation bound between the model trained by FedAdam-SSM and centralized Adam, we optimize the SSM to mitigate the learning performance degradation caused by sparsification error. Additionally, we provide convergence bounds for FedAdam-SSM in both convex and non-convex objective function settings, and investigate the impact of local epoch, learning rate and sparsification ratio on the convergence rate of FedAdam-SSM. Experimental results show that FedAdam-SSM outperforms baselines in terms of convergence rate (over 1.1$\times$ faster than the sparse FedAdam baselines) and test accuracy (over 14.5\% ahead of the quantized FedAdam baselines).
\end{abstract}


\section{Introduction}\label{sec:introduction}
\IEEEPARstart{R}{ecent} advances in Internet of Things (IoT) and Artificial Intelligence (AI) technologies have empowered next-generation smart devices to bring us higher levels of comfort, convenience, connectivity, and intelligence\cite{DBLP:journals/iotj/ChangLXCT21}. The explosion of IoT data, limited data transmission capacity, as well as potential data privacy and security threats, make it impractical to upload all the raw data to a remote cloud for centralized machine learning (ML)\cite{DBLP:journals/corr/abs-2202-03402}. To avoid collecting private data from distributed devices, federated learning (FL) has emerged as a privacy-enhancing distributed ML paradigm\cite{DBLP:journals/tpds/LiSWDMSHP22, DBLP:journals/titb/AliNTK23,DBLP:journals/jsac/DengLMWSDC23}, where distributed devices perform local model training on their private datasets, and a centralized server aggregates the uploaded local models to build a global model. 

In recent years, the size of ML models has dramatically increased to achieve high accuracy and adapt to dynamic environments in many application scenarios such as computer vision, natural language processing and speech recognition\cite{DBLP:journals/iotj/BianAXLLCWDG22,DBLP:journals/tkde/WangFHHW22}. In practice, ML models, especially deep learning models, are typically hundreds of megabytes or even gigabytes in size. As the size of state-of-the-art ML models continues to explode, the implementation of an efficient FL paradigm faces the following challenges. 1) \textit{Convergence speed.} Stochastic Gradient Descent (SGD) optimizer has been widely applied to FL training. Despite its simplicity, the vanilla SGD suffers from slow convergence, and is sensitive to variations in hyper-parameters (e.g., learning rate). To reduce the computational burden on distributed devices throughout the FL training process, it is paramount to deploy advanced optimizers to improve the convergence rate while maintaining model accuracy. 2) \textit{Communication overhead.} Since distributed devices in FL frequently upload/download ML models to/from the centralized server, large-scale ML models inevitably lead to unaffordable communication overhead. Especially in wireless networks, high traffic between distributed devices and the centralized server can overwhelm the limited transmission bandwidth, resulting in frequent interruptions, prolonged latencies and prohibitive costs. 

As a variant of SGD, adaptive moment estimation (Adam) optimizer has gained widespread popularity in FL due to its fast convergence and intuitive explanations behind the adaptive per-parameter learning rate\cite{DBLP:journals/corr/abs-2111-00856,DBLP:journals/jmlr/TranRMF22}. However, federated Adam (FedAdam for short) algorithms suffer from a threefold increase in uplink communication overhead compared to federated SGD (FedSGD for short) algorithms, which arises from the necessity to transmit not only local model updates but also first and second moment estimates (i.e., exponential moving averages of historical gradients and gradient squares) from distributed devices to the centralized server for aggregation. Despite current efforts to enhance the convergence rate of FedAdam, how to reduce this communication overhead remains open in the literature. Driven by this issue, we propose a novel sparsification method for communication-efficient FedAdam. The key contributions of this paper are summarized as follows: 

\begin{enumerate}
	\item \textbf{FedAdam-SSM}. We propose a sparse FedAdam algorithm called FedAdam-SSM, wherein distributed devices sparsify the updates of local model parameters and moment estimates and subsequently upload the sparse representations to the centralized server. The key innovation is the incorporation of a shared sparse mask (SSM) in the sparsification process of the updates of local model parameters and moment estimates updates, eliminating the need for separate sparse masks for each updates. Compared with the vanilla FedAdam, FedAdam-SSM effectively reduces the uplink communication overhead from $\mathcal{O}(3dq)$ to $\mathcal{O}(3kq+d)$.		
	\item \textbf{Shared sparse mask design}. We optimize the SSM to mitigate the learning performance degradation caused by the sparsification error. We first derive an upper bound on the deviation between the local model trained by FedAdam-SSM and the target model trained by centralized Adam. By minimizing this deviation bound, we devise an optimal SSM to minimize the sparsification error of FedAdam-SSM.
	\item \textbf{Convergence analysis}. We provide convergence bounds for FedAdam-SSM in both non-convex and convex objective function settings. We prove that the convergence rate of FedAdam-SSM achieves a linear speedup of $\mathcal{O}(\frac{1}{\sqrt{T}})$ in the non-convex setting, and can be further increased to $\mathcal{O}(\frac{1}{T})$ in the convex setting. Furthermore, we investigate the impact of local epoch, learning rate, sparsification ratio, and data imbalance on the convergence rate of FedAdam-SSM, and provide guidance for selecting appropriate values of local epoch and learning rate to improve the performance of FedAdam-SSM. 
	\item \textbf{Experimental results}. We conduct extensive experiments involving training of Convolutional Neural Network (CNN), Visual Geometry Group-11 (VGG-11) and Residual Network-18 (ResNet-18) over Fashion-MNIST, CIFAR-10 and SVHN datasets to investigate the training performance of the proposed FedAdam-SSM. The experimental results verify our theoretical analysis and show that FedAdam-SSM outperforms baselines in terms of convergence rate (over 1.1$\times$ faster than the best sparse FedAdam baseline) and test accuracy (over 14.5\% ahead of the best quantized FedAdam baseline).
\end{enumerate}

The remainder of this paper is organized as follows. Section \ref{section2} reviews related works in the literature. Section \ref{section3} briefly provides some background on FL, FedAdam and sparsification techniques. Section \ref{section4} presents our proposed FedAdam-SSM algorithm, followed by Section \ref{section:ssm} that details the optimization of the SSM to minimize the sparsification error. Section \ref{section5} conducts the convergence analysis, and Section \ref{section6} represents the experimental results. Section \ref{section7} concludes this paper.

\section{Significance and Related Work}\label{section2}
In this section, we provide a literature review along three lines that are most related to this paper, namely sparse FL, adaptive optimizers and communication-efficient FedAdam.

\subsection{Sparse Federated Learning}

To alleviate the significant communication overhead inherent in the FL training process, many studies have focused on integrating sparsification techniques into FL, wherein distributed devices sparsify the local model parameters or updates before transmission to the centralized server for aggregation. Mainstream sparsifiers employed in FL include 1) Rand-$k$ sparsifier \cite{DBLP:journals/chinaf/WeiLMDSZCZ24}, which randomly selects $k$ elements from the tensor $\mathbb{R}^d$ with a predefined sparsification ratio, 2) Top-$k$ sparsifier \cite{sami2024secure,DBLP:journals/compsec/LuLLGY23}, which retains the $k$ elements with the largest magnitudes within the tensor, and 3) Threshold-$v$ sparsifier \cite{DBLP:conf/iclr/LinHM0D18}, which preserves elements whose magnitudes exceed a predefined threshold $v$. Although sparse FL effectively reduces the communication overhead, the compression errors introduced by sparsification can lead to non-negligible degradation of FL performance in terms of convergence rate and model accuracy.

To address this challenge, recent research \cite{DBLP:conf/nips/SahuDABCK21,DBLP:conf/icml/XuH22,DBLP:journals/twc/OhLWNC24} has introduced error compensation mechanisms into sparse FL, wherein distributed devices mitigate sparsification errors by maintaining error residuals and incorporating them into local model updates in subsequent communication rounds. The authors in \cite{DBLP:conf/icdcs/HanWL20,DBLP:journals/tmc/JiangXXWLQQ24,DBLP:journals/tpds/TangSLC23,DBLP:journals/tvt/SunMH20} developed adaptive sparsification algorithms that dynamically adjust the sparsification ratio during the FL process, seeking to optimize the trade-off between communication efficiency and model performance. For example, \cite{DBLP:conf/icdcs/HanWL20} introduced a bidirectional Top-$k$ sparsification method that reduces both uplink and downlink communication overhead, and proposed a reinforcement learning-based optimization algorithm to adaptively adjust the sparsification ratio, minimizing overall FL training time. \cite{DBLP:journals/tmc/JiangXXWLQQ24} provided a theoretical analysis of the relationship between sparsification ratio and model performance, and developed a Multi-Armed Bandit-based online learning algorithm to optimize the sparsification ratios, specifically accounting for the heterogeneous computational and communication capabilities of distributed devices participating in FL training. Building upon Top-$k$ sparsification, \cite{DBLP:conf/globecom/ZhengDC23,DBLP:conf/ecai/ShiT0ZC20,DBLP:conf/iccspa/BeitollahiLL22} proposed layer-wise sparsification algorithms that optimize sparsification ratios across different network layers, reducing sparsification errors in critical layers to preserve model accuracy. In addition, the authors in \cite{DBLP:conf/ciss/JinDX23,hu2023flexible,chen2024latency} investigated the optimization of device scheduling and resource allocation in wireless networks, minimizing overall training latency while ensuring model accuracy in sparse FL.

\subsection{Adaptive Optimizers}

The vanilla SGD optimizer suffers from a slow convergence rate and considerable effort required for hyper-parameter tuning, which can impose a significant computational burden on distributed devices and thereby result in prolonged FL latency and suboptimal model performance. These limitations are further exacerbated by the recent emergence of state-of-the-art models, especially in Generative AI, where SGD has increasingly struggled to achieve target training efficiency. In this context, adaptive optimizers have emerged as the mainstream choice for model training, offering superiority in accelerating convergence while eliminating the need for manual hyperparameter tuning through their ability to automatically adjust per-parameter learning rates.

Recent research has introduced a variety of adaptive optimizers to speed up FL convergence. The authors in \cite{DBLP:journals/tpds/ZhouYL22} proposed FedNAG, a Nesterov Accelerated Gradient Descent (NAG) based FL paradigm. Although the theoretical analysis suggests that FedNAG exhibits the same convergence rate of $\mathcal{O}\left(\frac{1}{T}\right)$ as FedSGD, experimental results demonstrate a superior convergence speed of FedNAG compared to FedSGD. Using various variance-reduced SGD optimizers, the authors in \cite{DBLP:conf/icpp/DinhTNBZZ20} and \cite{DBLP:journals/tsp/WangFLZ23} accelerated the convergence speed of FL by gradually eliminating the inherent variance of the gradients. In addition to the NAG and variance-reduced SGD optimizers, the Adam optimizer is highlighted for its ability to speed up convergence rate by adaptively tuning per-parameter learning rates using both first and second moment estimates. Specifically, the first moment estimates enables Adam to build inertia in the search direction, mitigating oscillations caused by noisy gradients and facilitating progress across flat regions of the loss landscape. The second moment estimates amplify learning rates for parameters with less pronounced gradient history, facilitating parameter updates for sparse but critical input features. Building upon these advantages, \cite{DBLP:conf/aistats/ChenGSY21,DBLP:conf/icc/Li0WSLY20,DBLP:conf/icdm/XianHH22} developed various FedAdam algorithms, demonstrating theoretical guarantees for accelerated convergence rate and superior empirical performance.

This integration of adaptive optimizers into FL has yielded significant improvements in training efficiency, with FedAdam garnering particular attention for not only well-documented empirical performance but also theoretical guarantees for both faster escape from saddle points and accelerated convergence to second-order stationary points compared to FedSGD. However, the superior convergence rate of FedAdam comes at the cost of a threefold increase in the uplink communication overhead compared to FedSGD, as not only the local model parameters but also both the local first and second moment estimates have to be uploaded to the centralized server for aggregation. In this context, developing strategies to mitigate this significant increase in communication overhead is of considerable importance in improving the training efficiency of FedAdam.

\subsection{Communication-efficient FedAdam}

The integration of sparsification into FedAdam presents fundamentally distinct challenges due to its complex per-parameter learning rate adaptation\cite{DBLP:conf/aistats/ChenGSY21,DBLP:conf/icc/Li0WSLY20,DBLP:conf/icdm/XianHH22}. Unlike FedSGD, where model updates exhibit linear dependence on current gradients, the local model update rule of FedAdam yields intricate non-linear relationships with historical gradient information, involving first moment estimates that capture exponential moving averages of past gradients and the second moment estimates that track exponential moving averages of squared gradients, with these terms coupled through a non-linear division operation that creates adaptive per-parameter learning rates. This non-linear coupling introduces complex patterns of sparsification error propagation during the sparse FedAdam training process. Specifically, sparsification errors not only leading to biased updates of local model parameters and moment estimates in the current communication round but also propagate through the moment estimates across communication rounds, especially with the sparsification errors in both the first and second moment estimates compounding via the non-linear division operation to bias future model updates. Such cascading sparsification error propagation significantly complicates the theoretical analysis of sparse FedAdam, making it challenging to establish rigorous bounds for quantifying both the individual and coupled impacts of sparsification errors across model parameters and moment estimates on model performance, and to identify critical dimensions of local model parameters and moment estimates in sparse FedAdam.

Although extensive research \cite{
DBLP:journals/chinaf/WeiLMDSZCZ24,
sami2024secure,
DBLP:journals/compsec/LuLLGY23,
DBLP:conf/iclr/LinHM0D18,
DBLP:conf/nips/SahuDABCK21,
DBLP:conf/icml/XuH22,
DBLP:journals/twc/OhLWNC24,
DBLP:journals/tmc/JiangXXWLQQ24,
DBLP:journals/tpds/TangSLC23,
DBLP:journals/tvt/SunMH20,
DBLP:conf/icdcs/HanWL20,
DBLP:conf/globecom/ZhengDC23,
DBLP:conf/ecai/ShiT0ZC20,
DBLP:conf/iccspa/BeitollahiLL22,
DBLP:conf/ciss/JinDX23,
hu2023flexible,
chen2024latency} has explored sparse FL techniques, these advances have primarily focused on FedSGD. Investigations into sparse FedAdam remain nascent in the literature. In \cite{DBLP:conf/ijcai/HuGG21} and \cite{DBLP:conf/icml/WangLC22}, distributed devices perform local updates using SGD and apply Rand-$k$ or Top-$k$ sparsification to stochastic gradients before transmission to the centralized server, which then aggregates these sparse gradients to update global first and second moment estimates and determines adaptive per-parameter learning rates for global model updates. Building upon \cite{DBLP:conf/ijcai/HuGG21} and \cite{DBLP:conf/icml/WangLC22}, the authors in \cite{DBLP:journals/corr/abs-2201-02664} and \cite{DBLP:conf/icml/Li023} incorporated error compensation mechanisms to mitigate sparsification errors by maintaining error residuals for local model updates in subsequent communication rounds. Subsequently, \cite{DBLP:conf/icc/Li0WSLY20} proposed to perform local updates using Adam while only transmitting the sparsified local model updates to the server for aggregation, keeping first and second moment estimates exclusively local to reduce communication overhead. Further exploration emerged in \cite{DBLP:journals/tist/ChenSHL21} and \cite{DBLP:journals/corr/abs-2205-14473}, where authors proposed quantizing and uploading only the local model updates or parameters, while reducing model accuracy degradation through error compensation mechanisms. To further improve the model accuracy, \cite{DBLP:conf/icml/TangGARLLLZH21} introduced a two-phase quantized FedAdam algorithm that initially employs vanilla FedAdam in a warm-up phase, followed by a compression phase where the second moment estimates serve as a fixed precondition. \cite{DBLP:journals/iotj/MillsHM20} employs uniform and exponential quantization methods to separately quantize the local model updates and moment estimates in the proposed quantized FedAdam algorithm. 

However, the existing literature on sparse FedAdam\cite{DBLP:conf/aistats/ChenGSY21,DBLP:conf/icc/Li0WSLY20,DBLP:conf/icdm/XianHH22,DBLP:conf/ijcai/HuGG21,DBLP:conf/icml/WangLC22,DBLP:journals/corr/abs-2201-02664,DBLP:conf/icml/Li023,DBLP:journals/tist/ChenSHL21,DBLP:journals/corr/abs-2205-14473,DBLP:conf/icml/TangGARLLLZH21,DBLP:journals/iotj/MillsHM20} exhibits critical limitations. \cite{DBLP:conf/icml/WangLC22} and \cite{DBLP:journals/corr/abs-2201-02664} constrain per-parameter learning rate adaptation on the server side while enforcing fixed learning rates for local model updates on the client side, which significantly limits the effectiveness of Adam optimization. While the studies in \cite{DBLP:conf/icc/Li0WSLY20,DBLP:journals/tist/ChenSHL21,DBLP:journals/corr/abs-2205-14473} have advanced to implement client-side per-parameter learning rate adaptation, they solely focused on transmitting sparse local model updates to the centralized server for aggregation, neglecting the aggregation of local moment estimates into global moment estimates in each communication round. These oversights in \cite{DBLP:journals/corr/abs-2201-02664,DBLP:conf/icml/WangLC22,DBLP:conf/icc/Li0WSLY20,DBLP:journals/tist/ChenSHL21,DBLP:journals/corr/abs-2205-14473} can lead to suboptimal update of local model parameters and potentially divergent optimization trajectories across distributed devices, stemming from both the reliance on stale local moment estimates and the absence of global moment information. Going forward, although \cite{DBLP:conf/icml/TangGARLLLZH21} attempted to involve both server-side and client-side per-parameter learning rate adaptation through transmitting both local model parameters and moment estimates for global aggregation, it introduced a communication-intensive warm-up phase of vanilla FedAdam. Additionally, \cite{DBLP:journals/corr/abs-2201-02664,DBLP:conf/icc/Li0WSLY20,DBLP:journals/tist/ChenSHL21,DBLP:journals/corr/abs-2205-14473,DBLP:conf/icml/TangGARLLLZH21} restrict the number of local epochs to 1, which results in extremely frequent exchanges between distributed devices and the centralized server and amplifies both the compression error and the communication overhead in FedAdam training. Unlike FedSGD, incorporating error compensation mechanisms to reduce sparsification errors \cite{DBLP:journals/corr/abs-2201-02664, DBLP:conf/icc/Li0WSLY20,DBLP:conf/icml/Li023} remains limited in the context of FedAdam due to the inherent non-linearity of Adam optimization. Furthermore, due to the complex non-linear coupling between model updates and moment estimates in sparse FedAdam, the aforementioned works in \cite{DBLP:journals/corr/abs-2201-02664,DBLP:conf/icml/WangLC22,DBLP:conf/icc/Li0WSLY20,DBLP:conf/icml/Li023,DBLP:journals/tist/ChenSHL21,DBLP:journals/corr/abs-2205-14473,DBLP:conf/icml/TangGARLLLZH21} did not provide theoretical analysis for quantifying how sparsification errors propagate through and compound across both model parameters and moment estimates, fundamentally limiting the understanding of their individual and coupled impacts on model performance.

In this work, we propose a novel integration of Top-$k$ sparsification method with FedAdam, specifically addressing the challenge of reducing the threefold uplink communication overhead in sparse FedAdam compared to sparse FedSGD. Different from the aforementioned works on sparse FedAdam, we maintain both local per-parameter learning rate adaptation on the client side and global moment estimate aggregation on the server side, ensuring up-to-date moment information throughout the training process. In contrast to the conventional approach of separately sparsifying local model parameters and moment estimates, we propose to employ an SSM for each local updates, thus further reducing communication overhead. To mitigate performance degradation caused by sparsification error, we optimize the SSM by minimizing the deviation bound between the model trained by the proposed FedAdam-SSM and the target model trained by centralized Adam. 

To the best of our knowledge, our work represents the first attempt to 1) {\bf design a shared sparse mask} for the sparsification of the local updates of model parameters and moment estimates in sparse FedAdam; 2)  {\bf derive a deviation bound} for quantifying both the individual and coupled impacts of sparsification errors across model parameters and moment estimates on training performance in sparse FedAdam, thereby establishing a theoretical foundation for the proposed shared sparse mask design by proving that Top-$k$ sparsification of local model parameter updates optimally minimizes the deviation bound; and 3) {\bf provide theoretical convergence guarantees, revealing critical insights} into how local epochs, sparsification ratios, and data imbalance impact convergence rates and model accuracy in FedAdam-SSM.

\section{Preliminaries and Background}\label{section3}
In this section, we overview a few important concepts and definitions with regard to FedAdam and sparsification method.  
  
\subsection{FedAdam}
Consider an FL network of $N$ devices collaborating to train an ML model over their respective local datasets. Let $\mathcal{N}=\{1,2,...,N\}$ denote the index set of the devices. The goal of FL is to find a set of model parameters $\boldsymbol{w}^*\in\mathbb{R}^d$ that minimizes the global loss function $F(\boldsymbol{w})$ on all the local datasets, i.e.,\begin{alignat}{1} F(\boldsymbol{w})=\frac{\sum_{n=1}^N\vert\mathcal{D}_n\vert f(\boldsymbol{w},\mathcal{D}_n)}{\sum_{n=1}^N\vert\mathcal{D}_n\vert},\end{alignat}
where $f(\boldsymbol{w},\mathcal{D}_n)$ denotes the loss function on the local dataset $\mathcal{D}_n$. In standard FedSGD, distributed devices utilize SGD to minimize the local loss function $F_n(\boldsymbol{w})$ on its local dataset by updating the local model parameters over a total of $L$ local epochs. The update rule in the $l$-th local epoch is\begin{alignat}{1} \boldsymbol{w}_n^{l+1,t}=\boldsymbol{w}_n^{l,t}-\eta\nabla f(\boldsymbol{w}_n^{l,t}, \tilde{\mathcal{D}}_n), \end{alignat}where $\boldsymbol{w}_n^{l,t}$ denotes the local model parameters of the $n$-th device in the $l$-th local epoch and $t$-th communication round, $\eta>0$ is the learning rate, $\nabla f(\boldsymbol{w}_n^{l,t}, \tilde{\mathcal{D}}_n)$ is the mini-batch stochastic gradient of the local loss function. For each local epoch, $\tilde{\mathcal{D}}_n$ represents a fresh mini-batch that is resampled from the complete local dataset $\mathcal{D}_n$.  

To improve the accuracy of the ML model and accelerate the training process, Adam was proposed as a variant of SGD by taking into consideration moment and root mean square propagation\cite{DBLP:journals/tist/ChenSHL21,DBLP:journals/corr/abs-2205-14473,DBLP:conf/icml/TangGARLLLZH21}. Using Adam, the local model update rule in FL training can be written as\begin{alignat}{1}\label{update_rule_model}
	\boldsymbol{w}_n^{l,t}=\boldsymbol{w}_n^{l-1,t}-\eta\frac{\boldsymbol{m}_n^{l,t}}{\sqrt{\boldsymbol{v}_n^{l,t}+\epsilon}},
\end{alignat}where the first moment estimates $\boldsymbol{m}_n^{l,t}$ amplifies the learning rate for parameters with consistent gradient directions and attenuates it for those with fluctuating directions, the second moment estimates $\boldsymbol{m}_n^{l,t}$ scales per-parameter learning rates by incorporating the exponentially decaying average of past squared gradients in the denominator, and $\epsilon$ is a safety offset for division by the second moment estimates. Specifically, the first and second moment estimates are computed as\begin{alignat}{1}
	\boldsymbol{m}_n^{l,t}&=\beta_1\boldsymbol{m}_n^{l-1,t}+(1-\beta_1)\nabla f(\boldsymbol{w}_n^{l-1,t}, \tilde{\mathcal{D}}_n),\label{update_rule_m}\\
	\boldsymbol{v}_n^{l,t}&=\beta_2\boldsymbol{v}_n^{l-1,t}+(1-\beta_2)\left(\nabla f(\boldsymbol{w}_n^{l-1,t}, \tilde{\mathcal{D}}_n)\right)^2,\label{update_rule_v}
\end{alignat}where $\beta_1\in[0,1)$ and $\beta_2\in[0,1)$ control the exponential decay rates of the moving averages. Therefore, the procedure of FedAdam in each communication round is as follows:
\begin{enumerate}
	\item According to the update rule in (\ref{update_rule_model}), (\ref{update_rule_m}), and (\ref{update_rule_v}), each device updates the local moment estimates and model parameters (i.e., $\boldsymbol{m}_n^{l,t}$, $\boldsymbol{v}_n^{l,t}$, and $\boldsymbol{w}_n^{l,t}$) over a total of $L$ local epochs at the beginning of each round.
	\item Each device uploads the updated local moment estimates and model parameters (i.e., $\boldsymbol{m}_n^{L,t}$, $\boldsymbol{v}_n^{L,t}$, and $\boldsymbol{w}_n^{L,t}$) to the centralized server.
	\item Upon receiving all the local moment estimates and local model parameters, the centralized server updates the global moment estimates and model parameters according to the federated averaging (FedAvg) algorithm, i.e., $\boldsymbol{M}^{t+1}=\frac{\sum_{n=1}^N\vert\tilde{\mathcal{D}}_n\vert\boldsymbol{m}_n^{L,t}}{\sum_{n=1}^N\vert\tilde{\mathcal{D}}_n\vert}$, $\boldsymbol{V}^{t+1}=\frac{\sum_{n=1}^N\vert\tilde{\mathcal{D}}_n\vert\boldsymbol{v}_n^{L,t}}{\sum_{n=1}^N\vert\tilde{\mathcal{D}}_n\vert}$, and $\boldsymbol{W}^{t+1}=\frac{\sum_{n=1}^N\vert\tilde{\mathcal{D}}_n\vert\boldsymbol{w}_n^{L,t}}{\sum_{n=1}^N\vert\tilde{\mathcal{D}}_n\vert}$. 
	\item Each device downloads the updated global moment estimates and model parameters from the centralized server for the next communication round of FedAdam. 
\end{enumerate} 

\begin{figure*}[t!]
	\centering
	\includegraphics[width=5.6in,angle=0]{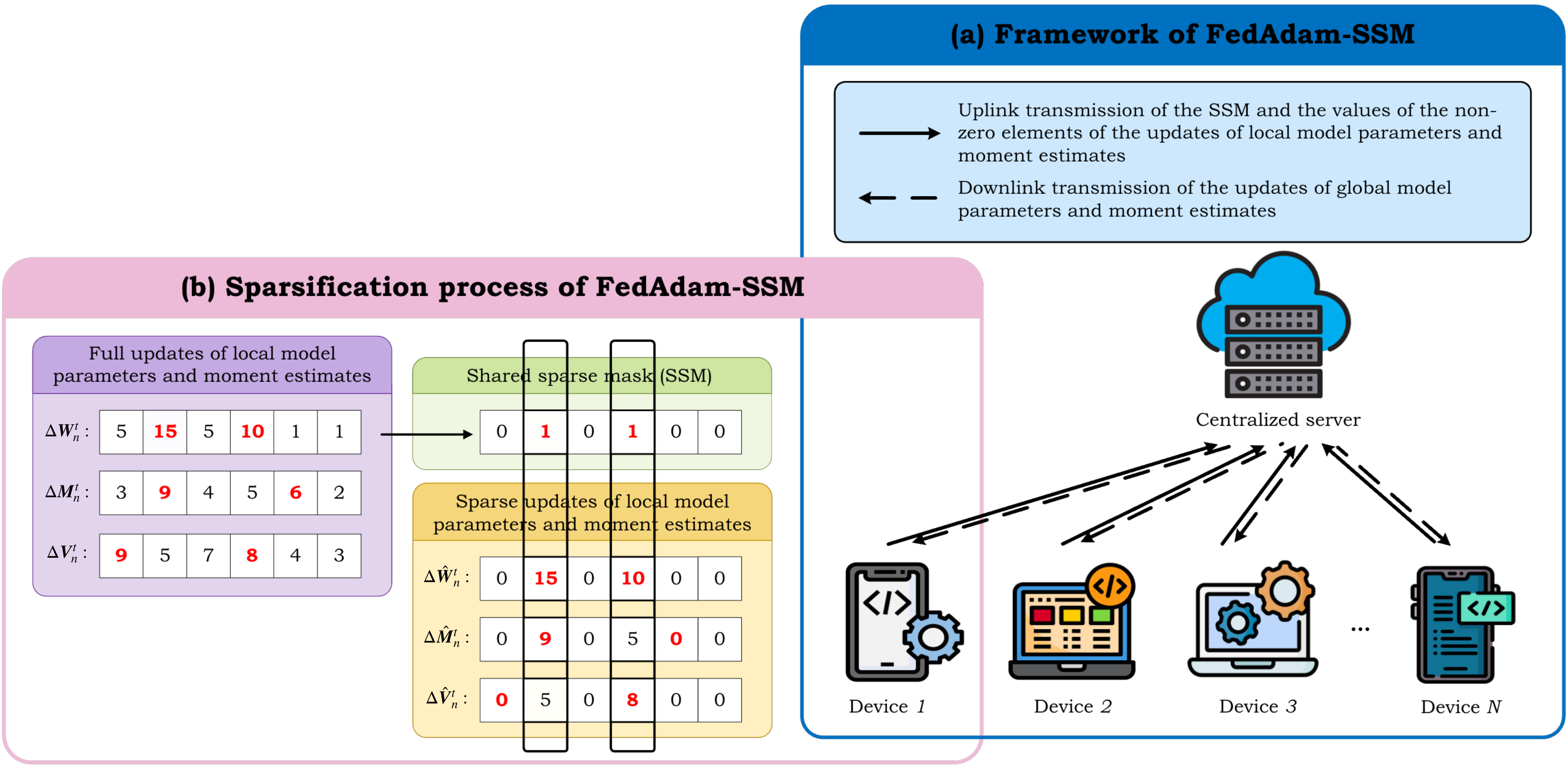}
	\caption{Illustration of the FedAdam-SSM algorithm.}
\end{figure*}

\subsection{Sparsification Method}

To reduce the communication overhead, sparsification methods have been extensively studied in FL to convert the local model parameters or accumulated gradients into a sparse representation before they are transmitted from distributed devices to the centralized server. Specifically, each device in FL induces a sparse mask on the $d$-dimensional tensor $\mathbb{R}^d$ to map each element of the tensor to either its original value or zero, and transmit the values of the non-zero elements and the sparse mask to the centralized server. The Top-$k$ sparsifier is defined as follows:
\begin{definition}
	(Top-$k$ sparsifier). For any positive integer $1\leq k\leq d$ and any tensor $\boldsymbol{x}\in\mathbb{R}^d$, the Top-$k$ sparsifier $\text{Top}_k$: $\mathbb{R}^d\rightarrow\mathbb{R}^d$ is defined as\begin{equation}
		\text{Top}_k(\boldsymbol{x})\coloneqq \boldsymbol{x} \odot\mathbbm{1}_{\text{Top}_k}(\boldsymbol{x}),
	\end{equation}where $\mathbbm{1}_{\text{Top}_k}(\boldsymbol{x})\in\{0,1\}^d$ represents the sparse mask of $\text{Top}_k$, and $\odot$ performs the element-wise product. The sparse mask of $\text{Top}_k$ is defined as\begin{equation}
		\left(\mathbbm{1}_{\text{Top}_k}(\boldsymbol{x})\right)_{\pi(j)}\coloneqq\begin{cases}
			1,&\mbox{if}~ j\leq k,\\
			0,&\mbox{otherwise},
		\end{cases}
	\end{equation}where $\pi$ is a permutation of $\left[d\right]$ such that $\vert\left(\boldsymbol{x}\right)_{\pi(j)}\vert\ge\vert\left(\boldsymbol{x}\right)_{\pi(j+1)}\vert$ for $j\in [1,d-1]$. The sparsification ratio is defined as the ratio of originally-valued elements over all elements, i.e., $\alpha = \frac{k}{d}$. 
\end{definition}

Clearly, the Top-$k$ sparsifier selects the top $k$ largest elements in terms of the absolute value. By selecting the $k$ largest elements of local model parameters and moment estimates, the Top-$k$ sparsifier can minimize the discrepancy between sparse and full updates for both global moment estimates and model parameters at a given sparsification ratio. This effectively reduces the potential loss of update information, thereby achieving superior effectiveness compared to other sparsifiers in the context of FedAdam.

\section{FedAdam-SSM}\label{section4}
In this section, we first present a straightforward sparse FedAdam algorithm called FedAdam-Top, which separately sparsifies the updates of local model parameters and moment estimates with the Top-$k$ sparsifier. Building upon FedAdam-Top, we propose FedAdam-SSM, which incorporates an SSM into the sparsification of the updates of local model parameters and moment estimates to further reduce the uplink communication overhead.

Using the Top-$k$ sparsifier, FedAdam-Top carries out in a different fashion from the vanilla FedAdam as follows:
\begin{enumerate}
	\item Let $\Delta\boldsymbol{M}_n^t=\boldsymbol{m}_n^{L,t}-\boldsymbol{M}^t$, $\Delta\boldsymbol{V}_n^t=\boldsymbol{v}_n^{L,t}-\boldsymbol{V}^t$, and $\Delta\boldsymbol{W}_n^t=\boldsymbol{w}_n^{L,t}-\boldsymbol{W}^t$ denote the updates of local moment estimates and model parameters. Upon completing the local model training, each device sparsifies the updates with the Top-$k$ sparsifier, i.e., $\Delta\hat{\boldsymbol{M}}_n^t=\Delta\boldsymbol{M}_n^t\odot\mathbbm{1}_{\text{Top}_k}\left(\Delta\boldsymbol{M}_n^t\right)$, $\Delta\hat{\boldsymbol{V}}_n^t=\Delta\boldsymbol{V}_n^t\odot\mathbbm{1}_{\text{Top}_k}\left(\Delta\boldsymbol{M}_n^t\right)$, and $\Delta\hat{\boldsymbol{W}}_n^t=\Delta\boldsymbol{W}_n^t\odot\mathbbm{1}_{\text{Top}_k}\left(\Delta\boldsymbol{M}_n^t\right)$. Then, each device uploads the Top-$k$ sparse masks (i.e., $\mathbbm{1}_{\text{Top}_k}\left(\Delta\boldsymbol{M}_n^t\right)$, $\mathbbm{1}_{\text{Top}_k}\left(\Delta\boldsymbol{V}_n^t\right)$, and $\mathbbm{1}_{\text{Top}_k}\left(\Delta\boldsymbol{W}_n^t\right)$) and the values of the non-zero elements of the sparse tensors (i.e., $\Delta\hat{\boldsymbol{M}}_n^t$, $\Delta\hat{\boldsymbol{V}}_n^t$, and $\Delta\hat{\boldsymbol{W}}_n^t$) to the centralized server.
	\item The centralized server reconstructs the sparse tensors $\Delta\hat{\boldsymbol{M}}_n^t$, $\Delta\hat{\boldsymbol{V}}_n^t$, and $\Delta\hat{\boldsymbol{W}}_n^t$ using the received sparse masks and values of the non-zero elements, and computes the updates of global moment estimates and model parameters as $\Delta\hat{\boldsymbol{M}}^t =\frac{\sum_{n=1}^N\vert\tilde{\mathcal{D}}_n\vert\Delta\hat{\boldsymbol{M}}_n^t}{\sum_{n=1}^N\vert\tilde{\mathcal{D}}_n\vert}$, $\Delta\hat{\boldsymbol{V}}^t =\frac{\sum_{n=1}^N\vert\tilde{\mathcal{D}}_n\vert\Delta\hat{\boldsymbol{V}}_n^t}{\sum_{n=1}^N\vert\tilde{\mathcal{D}}_n\vert}$, and $\Delta\hat{\boldsymbol{W}}^t =\frac{\sum_{n=1}^N\vert\tilde{\mathcal{D}}_n\vert\Delta\hat{\boldsymbol{W}}_n^t}{\sum_{n=1}^N\vert\tilde{\mathcal{D}}_n\vert}$. Afterwards, the centralized server transmits the updates of global moment estimates and model parameters $\Delta\hat{\boldsymbol{M}}^t$, $\Delta\hat{\boldsymbol{V}}^t$, and $\Delta\hat{\boldsymbol{W}}^t$ to each device.
	\item Upon receiving $\Delta\hat{\boldsymbol{M}}^t$, $\Delta\hat{\boldsymbol{V}}^t$, and $\Delta\hat{\boldsymbol{W}}^t$, each device updates the global moment estimates and model parameters as follows: $\boldsymbol{M}^{t+1} =\boldsymbol{M}^t+\Delta\hat{\boldsymbol{M}}^t$, $\boldsymbol{V}^{t+1} =\boldsymbol{V}^t+\Delta\hat{\boldsymbol{V}}^t$, and $\boldsymbol{W}^{t+1} =\boldsymbol{W}^t+\Delta\hat{\boldsymbol{W}}^t$.
\end{enumerate}We can see that by separately sparsifying the updates of local moment estimates and model parameters $\Delta\boldsymbol{M}_n^t$, $\Delta\boldsymbol{V}_n^t$, and $\Delta\boldsymbol{W}_n^t$ with the Top-$k$ sparsifier, distributed devices in FedAdam-Top transmit the values of the top $k$ largest elements of the tensors and the Top-$k$ sparse masks to the centralized server instead of the raw tensors. Let $q$ denote the floating-point precision. The total number of bits for uplink data transmission per communication round in FedAdam-Top is $3N(kq+d)$. Note that the total number of bits for uplink data transmission per communication round in the vanilla FedAdam is $3Ndq$. Therefore, the uplink data transmission volume per communication round can be reduced from $\mathcal{O}(3dq)$ in the vanilla FedAdam to $\mathcal{O}(3kq+3d)$ in FedAdam-Top. 

\begin{algorithm}[!t]
	\small
	\DontPrintSemicolon
	\SetKwBlock{DoParallel}{do in parallel}{end}
	\caption{FedAdam-SSM}\label{alg:Distributed Adam with Sparsification}
	Initialize: $\boldsymbol{W}^0$, $\boldsymbol{M}^0\gets 0$, $\boldsymbol{V}^0\gets 0$, $\Delta\hat{\boldsymbol{W}}^0\gets 0$, $\Delta\hat{\boldsymbol{M}}^0\gets 0$, $\Delta\hat{\boldsymbol{V}}^0\gets 0$;\\
	\For{\rm{each communication round} $t=1,...,T$}{
		\DoParallel{
			\For{\rm{each device} $n=1,2,...,N$}{
				Download $\Delta\hat{\boldsymbol{M}}^{t-1}$, $\Delta\hat{\boldsymbol{V}}^{t-1}$, $\Delta\hat{\boldsymbol{W}}^{t-1}$ from server;\\
				Update $\boldsymbol{M}^t\gets\boldsymbol{M}^{t-1}+\Delta\hat{\boldsymbol{M}}^{t-1}$, $\boldsymbol{V}^t\gets\boldsymbol{V}^{t-1}+\Delta\hat{\boldsymbol{V}}^{t-1}$, $\boldsymbol{W}^t\gets\boldsymbol{W}^{t-1}+\Delta\hat{\boldsymbol{W}}^{t-1}$;\\
				Set $\boldsymbol{w}_n^{0,t}\gets\boldsymbol{W}^t, \boldsymbol{m}_n^{0,t}\gets\boldsymbol{M}^t, \boldsymbol{v}_n^{0,t}\gets\boldsymbol{V}^t$;\\
				Update $\boldsymbol{m}_n^{l,t}$, $\boldsymbol{v}_n^{l,t}$, and $\boldsymbol{w}_n^{l,t}$ according to (\ref{update_rule_m}), (\ref{update_rule_v}), and (\ref{update_rule_model}) over $L$ epochs;\\
				Compute $\Delta\boldsymbol{M}_n^t\gets\boldsymbol{m}_n^{L,t}-\boldsymbol{M}^t$, $\Delta\boldsymbol{V}_n^t\gets\boldsymbol{v}_n^{L,t}-\boldsymbol{V}^t$, $\Delta\boldsymbol{W}_n^t\gets\boldsymbol{w}_n^{L,t}-\boldsymbol{W}^t$;\\
				Sparsify $\Delta\boldsymbol{M}_n^t$, $\Delta\boldsymbol{V}_n^t$, $\Delta\boldsymbol{W}_n^t$ as $\Delta\hat{\boldsymbol{M}}_n^t\gets\Delta\boldsymbol{M}_n^t\odot\mathbbm{1}_{\text{SSM}_n^t}$, $\Delta\hat{\boldsymbol{V}}_n^t\gets\Delta\boldsymbol{V}_n^t\odot\mathbbm{1}_{\text{SSM}_n^t}$, $\Delta\hat{\boldsymbol{W}}_n^t\gets\Delta\boldsymbol{W}_n^t\odot\mathbbm{1}_{\text{SSM}_n^t}$, and upload the SSM $\mathbbm{1}_{\text{SSM}_n^t}$ and the values of the non-zero elements of $\Delta\hat{\boldsymbol{M}}_n^t$, $\Delta\hat{\boldsymbol{V}}_n^t$, and $\Delta\hat{\boldsymbol{W}}_n^t$ to the server;\\}}
		The server reconstructs $\Delta\hat{\boldsymbol{M}}_n^t$, $\Delta\hat{\boldsymbol{V}}_n^t$, and $\Delta\hat{\boldsymbol{W}}_n^t$ with the received values and $\mathbbm{1}_{\text{SSM}_n^t}$, and updates $\Delta\hat{\boldsymbol{M}}^t$, $\Delta\hat{\boldsymbol{V}}^t$, and $\Delta\hat{\boldsymbol{W}}^t$ according to FedAvg;\\
	}
\end{algorithm}

To further reduce the communication overhead, we propose to improve FedAdam-Top by sparsifying the updates of local moment estimates and model parameters with an SSM. Denote the SSM by $\mathbbm{1}_{\text{SSM}_n^t}$. Notably, $\mathbbm{1}_{\text{SSM}_n^t}\in\{0,1\}^d$ is a binary tensor with exactly $k$ ones and $d-k$ zeros. As shown in Algorithm \ref{alg:Distributed Adam with Sparsification}, the proposed FedAdam-SSM algorithm sparsifies the updates $\Delta\boldsymbol{M}_n^t$, $\Delta\boldsymbol{V}_n^t$, and $\Delta\boldsymbol{W}_n^t$ as follows:\begin{alignat}{1}
	\Delta\hat{\boldsymbol{M}}_n^t =\Delta\boldsymbol{M}_n^t\odot\mathbbm{1}_{\text{SSM}_n^t},\\
	\Delta\hat{\boldsymbol{V}}_n^t =\Delta\boldsymbol{V}_n^t\odot\mathbbm{1}_{\text{SSM}_n^t},\\
	\Delta\hat{\boldsymbol{W}}_n^t =\Delta\boldsymbol{W}_n^t\odot\mathbbm{1}_{\text{SSM}_n^t}.
\end{alignat}As such, each device in FedAdam-SSM transmits a total of $4$ tensors to the centralized server, which includes $\mathbbm{1}_{\text{SSM}_n^t}$, and $3$ $k$-dimensional tensors containing the values of the non-zero elements of the sparse updates of local model parameters and moment estimates. FedAdam-SSM's uplink data transmission volume per communication round is $N\left(3kq+d\right)$ in bits. Therefore, the uplink data transmission volume per communication round can be further reduced from $\mathcal{O}(3kq+3d)$ in FedAdam-Top to $\mathcal{O}(3kq+d)$ in FedAdam-SSM.

\section{Design of A Shared Sparse Mask}\label{section:ssm}
In this section, we optimize the SSM to minimize learning performance degradation caused by sparsification errors. We begin by deriving an upper bound on the deviation between the local model trained by FedAdam-SSM and the target model obtained by centralized Adam, accounting for sparsification errors in both local model updates and moment estimates, as well as the imbalanced data distribution. Building upon this theoretical foundation, the main idea of optimizing the SSM is to minimize the deviation bound, enabling FedAdam-SSM to approximate the training performance of centralized Adam. 

\subsection{Learning Performance Analysis with Sparsification Error and Data Imbalance}

Before the theoretical analysis, we first make the following assumptions on the objective loss function.
\begin{assumption}[Lipschitz continuous gradients]\label{assumption1}
	The local loss function $f(\boldsymbol{w},\mathcal{D}_n)$ is differentiable, and its gradient $\nabla f(\boldsymbol{w},\mathcal{D}_n)$ is $\rho$-Lipschitz continuous, i.e., for any $n\in\mathcal{N}$ and $\boldsymbol{x}$, $\boldsymbol{y}\in\mathbb{R}^d$, $\Vert\nabla f(\boldsymbol{x},\mathcal{D}_n)$ $-\nabla f(\boldsymbol{y},\mathcal{D}_n)\Vert\leq \rho\left\Vert \boldsymbol{x}-\boldsymbol{y} \right\Vert$.
\end{assumption}
\begin{assumption}[Bounded gradients]\label{assumption3}
	For any $\boldsymbol{w}\in \mathbb{R}^d$, $\xi \in \cup\mathcal{D}_n$, and $j\in[d]$, the loss function $f(\boldsymbol{w},\xi)$ has bounded gradient, i.e., $ \left\vert\left[\nabla f(\boldsymbol{w}, \xi)\right]_j\right\vert \leq G$.
\end{assumption}
\begin{assumption}[Bounded gradient variances]\label{assumption2}
	The stochastic gradient $\nabla f(\boldsymbol{w},\xi)$ has bounded variance, i.e., for any $n\in\mathcal{N}$, $\xi\in\mathcal{D}_n$, and $\boldsymbol{w}\in \mathbb{R}^d$, $\mathbb{E}\left[\left\Vert\nabla f(\boldsymbol{w},\xi)-\nabla f(\boldsymbol{w},\mathcal{D}_n)\right\Vert^2\right]\leq\sigma_l^2$. We assume that the variance of the local gradient $\nabla f(\boldsymbol{w},\mathcal{D}_n)$ is bounded, i.e., $\left\Vert\nabla f(\boldsymbol{w},\mathcal{D}_n)\right.$ $\left.-\nabla f(\boldsymbol{w},\cup\mathcal{D}_n)\right\Vert^2\leq\sigma_g^2$.
\end{assumption}

Note that \textbf{Assumption} \ref{assumption1} on bounded smoothness and \textbf{Assumption} \ref{assumption3} on bounded gradients are widely used in the FL community for convergence analysis\cite{DBLP:journals/tist/ChenSHL21,DBLP:journals/corr/abs-2205-14473,DBLP:conf/icml/TangGARLLLZH21,DBLP:conf/icdm/XianHH22,DBLP:journals/iotj/LiLZD22}. In \textbf{Assumption} \ref{assumption2}, the bounded variances of local and global gradients characterize the sampling noise and data heterogeneity across distributed devices in FL, respectively. 

To facilitate our analysis, we next introduce several auxiliary notations. Let $\breve{\boldsymbol{M}^t}$, $\breve{\boldsymbol{V}^t}$, and $\breve{\boldsymbol{W}^t}$ denote the non-sparse global moment estimates and model parameters at the beginning of the $t$-th communication round, i.e., $\breve{\boldsymbol{M}^t}=\boldsymbol{M}^{t-1}+\frac{\sum_{n=1}^N\vert\tilde{\mathcal{D}}_n\vert\Delta\boldsymbol{M}_n^{t-1}}{\sum_{n=1}^N\vert\tilde{\mathcal{D}}_n\vert}$, $\breve{\boldsymbol{V}^t}=\boldsymbol{V}^{t-1}+\frac{\sum_{n=1}^N\vert\tilde{\mathcal{D}}_n\vert\Delta\boldsymbol{V}_n^{t-1}}{\sum_{n=1}^N\vert\tilde{\mathcal{D}}_n\vert}$, and $\breve{\boldsymbol{W}^t}=\boldsymbol{W}^{t-1}+\frac{\sum_{n=1}^N\vert\tilde{\mathcal{D}}_n\vert\Delta\boldsymbol{W}_n^{t-1}}{\sum_{n=1}^N\vert\tilde{\mathcal{D}}_n\vert}$. Let $\check{\boldsymbol{m}}^{l,t}$, $\check{\boldsymbol{v}}^{l,t}$, and $\check{\boldsymbol{w}}^{l,t}$ denote the auxiliary moment estimates and model parameters that follow the update rule of centralized Adam. To be specific, $\check{\boldsymbol{m}}^{l,t}$, $\check{\boldsymbol{v}}^{l,t}$, and $\check{\boldsymbol{w}}^{l,t}$ start from the non-sparse global moment estimates and model parameters $\breve{\boldsymbol{M}^t}$, $\breve{\boldsymbol{V}^t}$, and $\breve{\boldsymbol{W}^t}$, and update with the full global gradient descent $\nabla f(\check{\boldsymbol{w}}^{l,t}, \cup\mathcal{D}_n)$ as follows:\begin{alignat}{1}&\;\check{\boldsymbol{m}}^{l+1,t}=\beta_1\check{\boldsymbol{m}}^{l,t}+\left(1-\beta_1\right)\nabla f(\check{\boldsymbol{w}}^{l,t}, \cup\mathcal{D}_n),\\& \check{\boldsymbol{v}}^{l+1,t}=\beta_2\check{\boldsymbol{v}}^{l,t}+\left(1-\beta_2\right)\left(\nabla f(\check{\boldsymbol{w}}^{l,t},\cup\mathcal{D}_n)\right)^2,\\&\quad\quad\quad \check{\boldsymbol{w}}^{l+1,t}=\check{\boldsymbol{w}}^{l,t}-\eta\frac{\check{\boldsymbol{m}}^{l+1,t}}{\sqrt{\check{\boldsymbol{v}}^{l+1,t}+\epsilon}}.\end{alignat}

For ease of exposition, we introduce the following shorthand notations: $D_n=\vert\mathcal{D}_n\vert$, $\tilde{D}_n=\vert\tilde{\mathcal{D}}_n\vert$, $\tilde{F}_n(\boldsymbol{w})=f(\boldsymbol{w},\tilde{\mathcal{D}}_n)$, $F_n(\boldsymbol{w})=f(\boldsymbol{w},\mathcal{D}_n)$, and $F(\boldsymbol{w})=f(\boldsymbol{w},\cup\mathcal{D}_n)$.

\begin{theorem}\label{theorem_1}
	Suppose \textbf{Assumptions} \ref{assumption1}, \ref{assumption3}, and \ref{assumption2} hold. For any $n\in\mathcal{N}$, $l\in\mathcal{L}$, and $t\in\mathcal{T}$, the deviation between the local model trained by FedAdam-SSM and the target model trained by centralized Adam can be bounded as follows:\begin{alignat}{1}
		&\left\Vert\boldsymbol{w}_n^{l,t}-\check{\boldsymbol{w}}^{l,t}\right\Vert\nonumber\\&\leq\Gamma\left\Vert\boldsymbol{W}^{t}-\breve{\boldsymbol{W}^t}\right\Vert+\Lambda\left\Vert\boldsymbol{M}^{t}-\breve{\boldsymbol{M}^t}\right\Vert+\Theta\left\Vert\boldsymbol{V}^{t}-\breve{\boldsymbol{V}^t}\right\Vert+\Phi\nonumber\\
		&\leq\sum_{n=1}^N\!\frac{\tilde{D}_n}{\sum_{n=1}^N\!\tilde{D}_n}\bigg(\Gamma\Big\Vert\Big(1-\mathbbm{1}_{\text{SSM}_n^{t-1}}\Big)\!\odot\!\Delta\boldsymbol{W}_n^{t-1}\Big\Vert+\Lambda\left\Vert\left(1-\mathbbm{1}_{\text{SSM}_n^{t-1}}\right)\right.\nonumber\\&\left.\odot \Delta\boldsymbol{M}_n^{t-1}\right\Vert+\Theta\left\Vert\left(1-\mathbbm{1}_{\text{SSM}_n^{t-1}}\right)\odot\Delta\boldsymbol{V}_n^{t-1}\right\Vert\bigg)+\Phi\label{weighted_sum},
	\end{alignat}where\begin{alignat}{1}\label{equ11:theorem_1}
		&\Gamma=\frac{1}{\sqrt{\psi^2+4\phi}}\left(\left(\frac{\psi-\sqrt{\psi^2+4\phi}}{2}\right)^l\!\right.\Bigg(\phi+\frac{\sqrt{\psi^2+4\phi}-\psi}{2}-\beta_1\left(1\right.\nonumber\\&\left.-\beta_2\right)\frac{dG^2\eta\rho}{\epsilon\sqrt{\epsilon}}\Bigg)+\Bigg(\frac{\sqrt{\psi^2+4\phi}+\psi}{2}-\phi+\left.\frac{dG^2\eta\rho}{\epsilon\sqrt{\epsilon}}\beta_1\left(1-\beta_2\right) \Bigg)\right.\nonumber\\&\left.\times\left(\frac{\psi+\sqrt{\psi^2+4\phi}}{2}\right)^l\right),\!\!\!\!\end{alignat}\begin{alignat}{1}\label{equ12:theorem_1}
		\Lambda\!=\!\frac{\eta\beta_1}{\sqrt{\epsilon}\sqrt{\psi^2+4\phi}}\left(\left(\frac{\psi\!+\!\sqrt{\psi^2\!+\!4\phi}}{2}\right)^l\!-\!\left(\frac{\psi\!-\!\sqrt{\psi^2\!+\!4\phi}}{2}\right)^l\right),\end{alignat}\begin{alignat}{1}\label{equ13:theorem_1}
		\Theta\!=\!\frac{\sqrt{d}G\eta\beta_2}{2\epsilon\sqrt{\epsilon}\sqrt{\psi^2\!+\!4\phi}}\left(\left(\frac{\psi\!+\!\sqrt{\psi^2\!+\!4\phi}}{2}\right)^l\left(\frac{\psi\!-\!\sqrt{\psi^2\!+\!4\phi}}{2}\right)^l\right),\end{alignat}and\begin{alignat}{1}\label{equ14:theorem_1}
		&\Phi=\!\frac{\frac{\sigma_l}{\sqrt{\tilde{D}_n}}\!+\!\sigma_g}{\sqrt{\psi^2\!+\!4\phi}}\!\left(\!\frac{\eta}{\sqrt{\epsilon}}\left(1-\beta_1\right)\!+\!\frac{dG^2\eta}{\epsilon\sqrt{\epsilon}}\left(1-\beta_2\right) \!\right)\!\left(\Bigg(\frac{\psi\!+\!\sqrt{\psi^2\!+\!4\phi}}{2}\Bigg)^l\right.\nonumber\\&-\left.\!\left(\frac{\psi\!-\!\sqrt{\psi^2\!+\!4\phi}}{2}\!\right)^l\right)\!+\!\frac{\chi}{1-\!\psi-\!\phi}\!\left(\!\frac{1}{\sqrt{\psi^2\!+\!4\phi}}\left(\Bigg(1-\frac{\psi\!+\!\sqrt{\psi^2\!+\!4\phi}}{2}\Bigg)\!\right.\right.\nonumber\\&\times\!\left(\!\frac{\psi\!-\!\sqrt{\psi^2\!+\!4\phi}}{2}\!\right)^l\!\!\!-\left.\left.\!\!\!\Bigg(\!1\!-\!\frac{\psi\!-\!\sqrt{\psi^2\!+\!4\phi}}{2}\!\Bigg)\!\left(\!\frac{\psi\!+\!\sqrt{\psi^2\!+\!4\phi}}{2}\!\right)^l\right)\!+\!1\!\right).\!\!\!\!
	\end{alignat}Note that $\phi$, $\psi$ and $\chi$ are given as\begin{alignat}{1}\label{phi}
		\phi=\frac{\beta_1}{\sqrt{\beta_2}},\end{alignat}\begin{alignat}{1}&\label{psi}
		\psi=1+\frac{\beta_1}{\sqrt{\beta_2}}+\frac{\eta\rho\left(1-\beta_1\right)}{\sqrt{\epsilon}}\left(1+\frac{\left(1-\beta_2\right)dG^2}{\epsilon}\right),\!\!\end{alignat}\begin{alignat}{1}\label{chi}
		&\chi=dG\eta\left(\frac{2\beta_1\left(1-\sqrt{\beta_2}\right)}{\epsilon\sqrt{\epsilon\beta_2}}\left(G^2+\epsilon\right)+\frac{\left(1-\beta_1\right)\beta_2}{\epsilon\sqrt{\epsilon}}G^2\right)\nonumber\\&+\frac{\left(1-\beta_1\right)\eta\left(\frac{\sigma_l}{\sqrt{\tilde{D}_n}}+\sigma_g\right)}{\sqrt{\epsilon}}\left(1+\frac{\left(1-\beta_2\right)dG^2}{\epsilon}\right).
	\end{alignat}
\end{theorem}
\begin{IEEEproof}
	See the proof in the supplementary material.
\end{IEEEproof}

\begin{remark}
	\textbf{Theorem} \ref{theorem_1} demonstrates that the upper bound on $\left\Vert\boldsymbol{w}_n^{l,t}-\check{\boldsymbol{w}}^{l,t}\right\Vert$ is dominated by the term $\Phi$ and the weighted sum of $\left\Vert\boldsymbol{W}^{t}-\breve{\boldsymbol{W}^t}\right\Vert$, $\left\Vert\boldsymbol{M}^{t}-\breve{\boldsymbol{M}^t}\right\Vert$ and $\left\Vert\boldsymbol{V}^{t}-\breve{\boldsymbol{V}^t}\right\Vert$ as follows:
	\begin{enumerate}
	\item The term $\Phi$ is determined by the variances of local and global gradient (i.e., $\frac{\sigma_l}{\sqrt{\tilde{D}_n}}$ and $\sigma_g$). Decreases in local and global gradient variances can contribute to a smaller deviation between the model parameters trained by FedAdam-SSM and centralized Adam, which indicates an improved learning performance in FedAdam-SSM training. This is consistent with the fact that FL performs better on IID datasets than non-IID datasets.
	\item The weighted sum term reveals how the sparsification of local model parameter and moment estimate updates affects the performance of FedAdam-SSM. Specifically, reducing the sparsification errors of model parameters and moment estimates, i.e., $\left\Vert\boldsymbol{W}^{t}-\breve{\boldsymbol{W}^t}\right\Vert$, $\left\Vert\boldsymbol{M}^{t}-\breve{\boldsymbol{M}^t}\right\Vert$, and $\left\Vert\boldsymbol{V}^{t}-\breve{\boldsymbol{V}^t}\right\Vert$, can lead to a reduced gap between $\boldsymbol{w}_n^{l,t}$ and $\check{\boldsymbol{w}}^{l,t}$, thereby improving the learning performance of FedAdam-SSM. In the case where sparsification errors vanish entirely, i.e., $\left\Vert\boldsymbol{W}^{t}-\breve{\boldsymbol{W}^t}\right\Vert=0$, $\left\Vert\boldsymbol{M}^{t}-\breve{\boldsymbol{M}^t}\right\Vert=0$, and $\left\Vert\boldsymbol{V}^{t}-\breve{\boldsymbol{V}^t}\right\Vert=0$, the deviation between FedAdam-SSM and centralized Adam is reduced to the deviation between vanilla FedAdam and centralized Adam, i.e.,
		\begin{alignat}{1} \left\Vert\boldsymbol{w}_n^{l,t}-\check{\boldsymbol{w}}^{l,t}\right\Vert\leq\Phi,
		\end{alignat}relying solely on data heterogeneity across FL devices. 
\end{enumerate}
\end{remark}

\begin{remark}
	\textbf{Theorem} \ref{theorem_1} also shows that for the same sparsification ratio, FedAdam-Top can achieve the lowest weighted sum of the sparsification errors by separately sparsifying the local updates $\Delta\boldsymbol{W}_n^t$, $\Delta\boldsymbol{M}_n^t$, and $\Delta\boldsymbol{V}_n^t$ with Top-$k$ sparsification. This indicates that the deviation between FedAdam-Top and centralized Adam serves as a lower bound on the deviation between sparse FedAdam and centralized Adam.
\end{remark}

\subsection{Optimal Shared Sparse Mask} 

From \textbf{Theorem} \ref{theorem_1}, the main idea of minimizing the deviation between FedAdam-SSM and centralized Adam is to minimize the weighted sum of the sparsification errors, i.e.,\begin{alignat}{1}\label{equ:weighted_sum}
	&\Gamma\left\Vert\Delta\boldsymbol{W}_n^t\odot\left(1-\mathbbm{1}_{\text{SSM}_n^t}\right)\right\Vert+\Lambda\Big\Vert\Delta\boldsymbol{M}_n^t\odot\left(1-\mathbbm{1}_{\text{SSM}_n^t}\right)\Big\Vert\nonumber\\&+\Theta\left\Vert\Delta\boldsymbol{V}_n^t\odot\left(1-\mathbbm{1}_{\text{SSM}_n^t}\right)\right\Vert.\!\!\!\!
\end{alignat}To this end, we proceed to find the optimal SSM by comparing the magnitude of $\Gamma$, $\Lambda$ and $\Theta$ as follows.\begin{proposition}\label{proposition_1}
	If the exponential decay rate $\beta_2$ satisfies\begin{alignat}{1} 
		\beta_2<1-\frac{1}{1+2G\rho\sqrt{d}},\label{condition} \end{alignat}then\begin{alignat}{1}
		\Gamma>\Theta>\Lambda.
	\end{alignat} 
\end{proposition}
\begin{IEEEproof}
	See the proof in the supplementary material.
\end{IEEEproof}
\begin{remark}\label{remark_1}
	Notably, $\beta_2\in[0,1)$ is the exponential decay rate of the second moment estimates, $\rho>0$ is the Lipschitz constant, and $G>0$ is the upper bound on the stochastic gradient $\left\vert\left[\nabla f(\boldsymbol{w}, \xi)\right]_j\right\vert$. Given that the number of model parameters $d>0$ is typically extremely large, $1-\frac{1}{1+2G\rho\sqrt{d}}$ approaches $1$. Therefore, the inequality in (\ref{condition}) can be readily satisfied since $\beta_2\in[0,1)$ is often set to $0.999$ in the Adam optimizer.
\end{remark}

Theoretical and experimental analysis of model parameters and moment estimates in Adam\cite{DBLP:journals/iotj/MillsHM20, DBLP:conf/icann/BockW19, DBLP:conf/icml/LiuSLHHC21} demonstrate that $\Delta\boldsymbol{W}_n^t$ exhibits significantly larger magnitudes than $\Delta\boldsymbol{M}_n^t$ and $\Delta\boldsymbol{V}_n^t$. Together with \textbf{Proposition} \ref{proposition_1}, it can be derived that the term $\Gamma\left\Vert\Delta\boldsymbol{W}_n^t\odot\left(1-\mathbbm{1}_{\text{SSM}_n^t}\right)\right\Vert$ is the dominant contributor to the deviation between FedAdam-SSM and centralized Adam. Therefore, the minimization of the weighted sum of the sparsification errors in (\ref{equ:weighted_sum}) can be equivalently transformed into the minimization of $\left\Vert\Delta\boldsymbol{W}_n^t\odot\left(1-\mathbbm{1}_{\text{SSM}_n^t}\right)\right\Vert$. As such, an optimal SSM can be determined by the Top-$k$ sparsification of the updates of local model parameters, i.e., \begin{alignat}{1}\label{design}\mathbbm{1}_{\text{SSM}_n^t}=\mathbbm{1}_{\text{Top}_k}\left(\Delta\boldsymbol{W}_n^t\right).
\end{alignat}With this optimal SSM, the proposed FedAdam-SSM can effectively mitigate the learning performance degradation caused by sparsification error.

\section{Convergence Analysis}\label{section5}
In this section, we analyze the convergence of the proposed FedAdam-SSM in both convex and non-convex settings.

We first present the convergence of FedAdam-SSM for general non-convex loss functions as follows.
\begin{theorem}[Non-convexity]\label{theorem_2}
	Under \textbf{Assumptions} \ref{assumption1}, \ref{assumption3}, and \ref{assumption2}, the following convergence bound holds for FedAdam-SSM after $T$ communication rounds:
	\begin{alignat}{1}\label{equ:theorem1}&\frac{1}{T}\!\sum_{t=0}^{T-1}\!\left\Vert\nabla F(\boldsymbol{W}^t)\right\Vert^2\!\leq \!\frac{2}{\eta T}\left(F\left(\boldsymbol{W}^0\right)\!-\!F\left(\boldsymbol{W}^{T}\right)\right)\!+\!2\left(\left(\eta\rho\!+\!2\right)\!(1\!-\!\alpha)\right.\nonumber\\&\left.+\eta\rho-1\right)\frac{\eta G^2dL^2}{\epsilon}\!+\!6 G^2d\Bigg(\!\left(L-\frac{\beta_2\left(1-(\beta_2)^L\right)}{1-\beta_2}\right)\frac{G^4dL}{4\epsilon^3}+\frac{L^2}{\epsilon}\nonumber\\&+\frac{4\beta_1\left(1-(\beta_1)^L\right)}{\epsilon\left(1-\beta_1\right)^2}+1+\frac{\rho^2L^2}{3\epsilon}\Bigg)+6\!\sum_{n=1}^N\!\frac{\tilde{D}_n\left(\frac{\sigma_l}{\sqrt{\tilde{D}_n}}+\sigma_g\right)^2}{\sum_{n=1}^N\tilde{D}_n}.\!\!\!\!\!\!
	\end{alignat}
\end{theorem}\begin{IEEEproof}
	See the proof in the supplementary material.
\end{IEEEproof}\begin{remark}The convergence bound of FedAdam-SSM given in \textbf{Theorem} \ref{theorem_2} includes a vanishing part as given by\begin{alignat}{1}\label{wdsfv}6\sum_{n=1}^N\frac{\tilde{D}_n\left(\frac{\sigma_l}{\sqrt{\tilde{D}_n}}+\sigma_g\right)^2}{\sum_{n=1}^N\tilde{D}_n},\end{alignat}that shows how data imbalance degrades FedAdam-SSM's training performance, which is determined by local and global variances, i.e., $\sigma_l$ captures the stochastic sampling noise, and $\sigma_g$ characterizes the imbalanced data distributions across FL participants, respectively. Specifically, imbalanced data distributions cause the global model to produce biased gradient estimates on local datasets, thereby resulting in a degraded convergence rate and model accuracy. This coincides with our intuitive understanding: a lower sparsification ratio can contribute to a reduced sparsification error, and thereby an improved model performance, while a non-IID data distribution can result in a global model that produces shifted gradients on local datasets, leading to degraded model performance.\end{remark}
\begin{remark}
From \textbf{Theorem} \ref{theorem_2}, the term in (\ref{wdsfv}) is decoupled from the sparsification ratio $\alpha$, revealing that FedAdam-SSM's sensitivity to data imbalance is governed by the intrinsic nature of federated optimization rather than the sparsification mechanism. This theoretical insight suggests that FedAdam-SSM exhibits sensitivity to imbalanced data distributions that is comparable to vanilla FedAdam, with both algorithms demonstrating comparable model performance degradation as data imbalance intensifies, thereby reflecting this fundamental challenge inherent to FL.
\end{remark}

\begin{proposition}\label{proposition_6}
	Under \textbf{Assumptions} \ref{assumption1}, \ref{assumption3}, and \ref{assumption2}, and with learning rate $\eta=\mathcal{O}\left(\frac{1}{L^2\sqrt{T}}\right)$, the following convergence bound holds for FedAdam-SSM:\begin{alignat}{1}
		&\frac{1}{T}\sum_{t=0}^{T-1}\left\Vert\nabla F(\boldsymbol{W}^t)\right\Vert^2\leq \mathcal{O}\left(\frac{\left(F\left(\boldsymbol{W}^0\right)-F\left(\boldsymbol{W}^{T}\right)\right)L^2}{\sqrt{T}}\right)\nonumber\\&+\mathcal{O}\left(\frac{ (1-\alpha) G^2d}{\epsilon\sqrt{T}}\right)+\mathcal{O}\left(\frac{(1-\alpha)\rho G^2d}{\epsilon  L^2T}\right).
	\end{alignat} 
\end{proposition}\begin{remark}\textbf{Proposition} \ref{proposition_6} reveals that for sufficiently large communication round $T$, the dominant term in the asymptotic convergence rate of FedAdam-SSM achieves a linear speedup of $\mathcal{O}\left(\frac{1}{\sqrt{T}}\right)$. This implies that to reach any target error $\delta>0$, FedAdam-SSM needs $\mathcal{O}\left(\frac{1}{\delta^2}\right)$ communication rounds on non-convex loss functions.\end{remark} 
\begin{remark}\textbf{Proposition} \ref{proposition_6} reveals that when \begin{alignat}{1}L<\left(\frac{(1-\alpha)\rho G^2d}{\epsilon \left(F\left(\boldsymbol{W}^0\right)-F\left(\boldsymbol{W}^{T}\right)\right)\sqrt{T}}\right)^\frac{1}{4},\end{alignat}the term $\mathcal{O}\left(\frac{(1-\alpha)\rho G^2d}{\epsilon  L^2T}\right)$ dominates the convergence bound of FedAdam-SSM, which indicates that the convergence rate can be improved by increasing the local epoch $L$. This aligns with our intuitive understanding that a large local epoch $L$ can lead to a local minimizer and thereby accelerates the overall FedAdam-SSM training process. However, there is a point where increasing the local epoch $L$ too much can degrade the model performance. When $L>\left(\frac{(1-\alpha)\rho G^2d}{\epsilon \left(F\left(\boldsymbol{W}^0\right)-F\left(\boldsymbol{W}^{T}\right)\right)\sqrt{T}}\right)^\frac{1}{4}$, the dominant term becomes $\mathcal{O}\left(\frac{\left(F\left(\boldsymbol{W}^0\right)-F\left(\boldsymbol{W}^{T}\right)\right)L^2}{\sqrt{T}}\right)$. In this regime, a large local epoch $L$ leads to a strong local model drift, which degrades the training performance of FedAdam-SSM. This analysis provides theoretical guidance for selecting an appropriate value of $L$ that balances local optimization against local model drift, thereby speeding up the convergence of FedAdam-SSM.\end{remark}

\begin{remark}\label{fsdghjklkjyhtg}Now consider vanilla FedAdam as a special case of FedAdam-SSM with the sparsification ratio $\alpha = 1$, yielding that $\left(\frac{(1-\alpha)\rho G^2d}{\epsilon \left(F\left(\boldsymbol{W}^0\right)-F\left(\boldsymbol{W}^{T}\right)\right)\sqrt{T}}\right)^\frac{1}{4}=0$. This implies for any $L > 0$, the convergence rate of vanilla FedAdam is dominated by the term $\mathcal{O}\left(\frac{\left(F\left(\boldsymbol{W}^0\right)-F\left(\boldsymbol{W}^{T}\right)\right)L^2}{\sqrt{T}}\right)$, indicating monotonic degradation in model performance as the number of local epochs increases. In this case, vanilla FedAdam precludes the initial regime observed in FedAdam-SSM, in which an increased number of local epochs $L$ enhances training performance. This aligns with the intuitive observation that vanilla FedAdam reduces to centralized Adam when the number of local epochs is set to 1, yielding the optimal training performance.
	
This key distinction in how the number of local epochs $L$ impacts training performance between FedAdam-SSM and vanilla FedAdam stems from FedAdam-SSM introducing a shared sparse mask to sparsify local model parameters and moment estimates, which implicitly regularizes the local model updates by attenuating the accumulation of inconsistent gradient estimates across FL participants, thereby effectively reducing local model drift. Consequently, FedAdam-SSM exhibits greater robustness to larger values of local epochs $L$ and enables more efficient local training.
\end{remark}

In the following, we present the convergence of FedAdam-SSM under the Polyak-$\L$ojasiewicz (P$\L$) condition.
\begin{assumption}[P$\L$ condition]\label{assumption5}
	The loss function $F(\boldsymbol{w})$ satisfies a P$\L$-$\mu$ inequality, i.e., $\left\Vert\nabla F(\boldsymbol{w})\right\Vert^2$ $\ge 2\mu\left(F(\boldsymbol{w})-F(\boldsymbol{w}^*)\right)$, $\forall \boldsymbol{w}\in\mathbb{R}^d$.
\end{assumption}

\begin{theorem}[P$\L$ condition]\label{theorem_3}
	Under \textbf{Assumptions} \ref{assumption1}, \ref{assumption3}, \ref{assumption2}, and \ref{assumption5}, the following convergence bound holds for FedAdam-SSM after $T$ communication rounds:\begin{alignat}{1}\label{42}&F\left(\boldsymbol{W}^{T}\right)-F(\boldsymbol{w}^*)\leq\left(1-\eta\mu\right)^T\left(F(\boldsymbol{W}^0)-F(\boldsymbol{w}^*)\right)+\frac{\eta G^2dL^2}{\mu\epsilon}\nonumber\\&((\eta\rho\!+\!2)(1\!-\!\alpha)\!+\!\eta\rho\!-\!1)\!+\!\frac{3 G^2d}{\mu}\left(\frac{4\beta_1\left(1\!-\!(\beta_1)^L\right)}{\epsilon\left(1\!-\!\beta_1\right)^2}\!+\!\frac{L^2}{\epsilon}\!+\!\frac{\rho^2L^2}{3\epsilon}\right.\nonumber\\&\left.+1\!+\!\frac{G^4dL}{4\epsilon^3}\Bigg(L\!-\!\frac{\beta_2\left(1\!-\!(\beta_2)^L\right)}{1\!-\!\beta_2}\Bigg)\right)+\frac{3}{\mu}\!\sum_{n=1}^N\!\frac{\tilde{D}_n\!\left(\!\frac{\sigma_l}{\sqrt{\tilde{D}_n}}\!+\!\sigma_g\!\right)^2}{\sum_{n=1}^N\tilde{D}_n}.\!\!\!\!\!
	\end{alignat}
\end{theorem}\begin{IEEEproof}
	See the proof in the supplementary material.
\end{IEEEproof}\begin{remark}From \textbf{Theorem} \ref{theorem_3}, it can be inferred that, under the P$\L$ condition, a small learning rate can slow down the convergence speed, while a large learning rate can compromise the model performance. Specifically, a small learning rate $\eta$ can result in a slow decay of the term $\left(1-\eta\mu\right)^T\left(F(\boldsymbol{W}^0)-F(\boldsymbol{w}^*)\right)$, thereby yielding a deceleration of the training process. On the other hand, using a large learning rate $\eta$ can contribute to an increased term $\frac{\left((\eta\rho+2)(1-\alpha)+\eta\rho-1\right)\eta G^2dL^2}{\mu\epsilon}$, thereby leading to degraded model accuracy. This is consistent with our intuitive understanding that a small $\eta$ can cause the optimizer to converge slowly, while a large $\eta$ can cause drastic updates which can lead to divergent behaviour.\end{remark}

\begin{proposition}\label{proposition_7}
	Under \textbf{Assumptions} \ref{assumption1}, \ref{assumption3}, \ref{assumption2}, and \ref{assumption5}, and with learning rate $\eta=\mathcal{O}\left(\frac{\ln T}{L^2T}\right)$, the following convergence bound holds for FedAdam-SSM:\begin{alignat}{1}\label{43}&F\left(\boldsymbol{W}^T\right)-F(\boldsymbol{w}^*)\leq\widetilde{\mathcal{O}}\left(\frac{F(\boldsymbol{W}^0)-F(\boldsymbol{w}^*)}{T}\right)\nonumber\\&+\widetilde{\mathcal{O}}\left(\left((2-\alpha)\right)\frac{\rho G^2d}{\mu\epsilon L^2T^2}\right)+\widetilde{\mathcal{O}}\left((1-\alpha)\frac{G^2d}{\mu\epsilon T}\right).
	\end{alignat}
\end{proposition}\begin{IEEEproof}
	See the proof in the supplemental material.
\end{IEEEproof}
\begin{remark}From \textbf{Proposition} \ref{proposition_7}, we can observe that under the P$\L$ condition, the convergence rate of the FedAdam-SSM algorithm can be improved to $\mathcal{O}\left(\frac{1}{T}\right)$ by setting $\eta=\mathcal{O}\left(\frac{\ln T}{L^2T}\right)$. That is, under the P$\L$ condition, FedAdam-SSM only requires $\mathcal{O}\left(\frac{1}{\delta}\right)$ communication rounds to reach an error $\delta>0$. \end{remark}

\section{Experimental Results}\label{section6}
\subsection{Experimental Settings} 

To evaluate the performance of the proposed FedAdam-SSM, we train a CNN on the Fashion-MNIST dataset, VGG-11 on the CIFAR-10 dataset, and ResNet-18 on the SVHN dataset with IID and non-IID settings to demonstrate the training efficiency.

\textbf{Datasets and models.} 1) Fashion-MNIST consists of a training set of $60000$ $28\times28$ grayscale images belonging to $10$ different classes, and a test set of $10000$ images. 2) CIFAR-10 consists of $60000$ $32\times32$ RGB images in $10$ classes (from $0$ to $9$), with $5000$ training images and $1000$ test images per class. 3) SVHN contains over $60000$ $32\times32$ RGB images in $10$ classes (from $0$ to $9$), which is cropped from real-world pictures of house number plates. For the Fashion-MNIST dataset, we adopt a CNN model with two $5\times5$ convolutional layers (each followed by ReLU activation and a $2\times2$ max pooling layer), two fully connected layers, and a final softmax output layer. For the CIFAR-10 dataset, we adopt a VGG-11 model that consists of eight $3\times3$ convolutional layers, three fully connected layers, and a final softmax output layer. For the SVHN dataset, we adopt a ResNet-18 model that consists of a $2\times2$ convolutional layer, two pooling layers, eight residual units (each with two $3\times3$ convolutional layers), a fully connected layer, and a final softmax output layer.

\textbf{Data distribution.} Following the previous works \cite{DBLP:conf/icml/YurochkinAGGHK19,DBLP:conf/iclr/WangYSPK20}, we implement the Non-IID data distribution using the Dirichlet distribution with concentration parameter $\theta=0.1$. Note that the concentration parameter serves as a control for the level of data imbalance, with lower values indicating higher degrees of Non-IID data distribution.

\textbf{Baselines.} We evaluate FedAdam-SSM using the following baseline algorithms: 1) Existing FL methods including vanilla Federated Adam and SGD (FedAdam, FedSGD) and their sparse variants (FedAdam-Top, 1-bit Adam\cite{DBLP:conf/icml/TangGARLLLZH21}, Efficient Adam\cite{DBLP:journals/corr/abs-2205-14473}, Sparse FedSGD). 2) Ablation studies including three FedAdam-SSM variants($\text{FedAdam-SSM}_V$, $\text{FedAdam-SSM}_M$ and Fairness-top\cite{DBLP:conf/icdcs/HanWL20}). Note that FedAdam represents a special case of FedAdam-SSM and FedAdam-Top when the sparsification ratio is set to $1$. The three ablation variants employ SSMs to sparsify the local moment estimates and model parameters, reducing uplink communication overhead. Specifically, Fairness-top determines its SSM by applying Top-$k$ sparsification to the union of local moment estimate and model parameter updates. $\text{FedAdam-SSM}_M$ and $\text{FedAdam-SSM}_V$ determine the SSM by applying Top-$k$ sparsification to the local updates of first moment estimates and second moment estimates, respectively.

\textbf{Implementation.} In the following experiments, we set the number of distributed devices $N=20$. To ensure a fair and robust comparison, we conducted a grid search to determine the hyperparameters for the proposed FedAdam-SSM and baseline algorithms. The final hyperparameter configurations are as follows. For FedAdam, sparse and quantized FedAdam algorithms, we set the hyperparameters of the Adam optimizer to $\beta_1=0.9$, $\beta_2=0.999$, and $\epsilon = 10^{-6}$. For FedAdam, FedSGD, and their sparse counterparts, we set the local epoch $L=30$ and sparsification ratio $\alpha = 0.05$. We set the learning rate $\eta = 0.001$ for FedAdam and sparse FedAdam algorithms. For 1-bit Adam, we set the number of warm epochs to 1, $\eta = 0.0001$ for CNN and ResNet, and $\eta = 0.001$ for VGG-11. For Efficient Adam, we set the quantization level to 2-bit for CNN, and 6-bit for ResNet and VGG-11, $\eta = 0.001$ for CNN and VGG-11, and $\eta = 0.0001$ for ResNet. For FedSGD and Sparse FedSGD, we set $\eta = 0.001$ for CNN and ResNet, and $\eta = 0.01$ for VGG-11. As previously stated, FedAdam-SSM uploads the SSM to the centralized server to represent the position of each non-zero element in the sparse tensors. Alternatively, we could upload the indices of these non-zero elements to represent their positions instead of uploading the SSM. This method necessitates $\log_2(d)$ bits to inform the centralized server of the index of each non-zero element within the $d$-dimensional sparse tensor\cite{DBLP:conf/isit/OzfaturaOG21,DBLP:journals/tgcn/LinLCGH23}. In our experiment, we employ both methods and select the one with the lower communication overhead. Therefore, the total number of bits for uplink data transmission per communication round in FedAdam-SSM and FedAdam-Top is $\min\{N(3kq+d), Nk\left(3q+\log_2(d)\right)\}$ and $\min\{3N(kq+d), 3Nk\left(q+\log_2(d)\right)\}$, respectively.

\subsection{Results and Discussion}

\subsubsection{Comparison of $\Delta\boldsymbol{W}_n^t$, $\Delta\boldsymbol{M}_n^t$ and $\Delta\boldsymbol{V}_n^t$}
\begin{figure}[!t]
	\centering
	\begin{subfigure}[!t]{0.4\textwidth}
		\includegraphics[width=\textwidth]{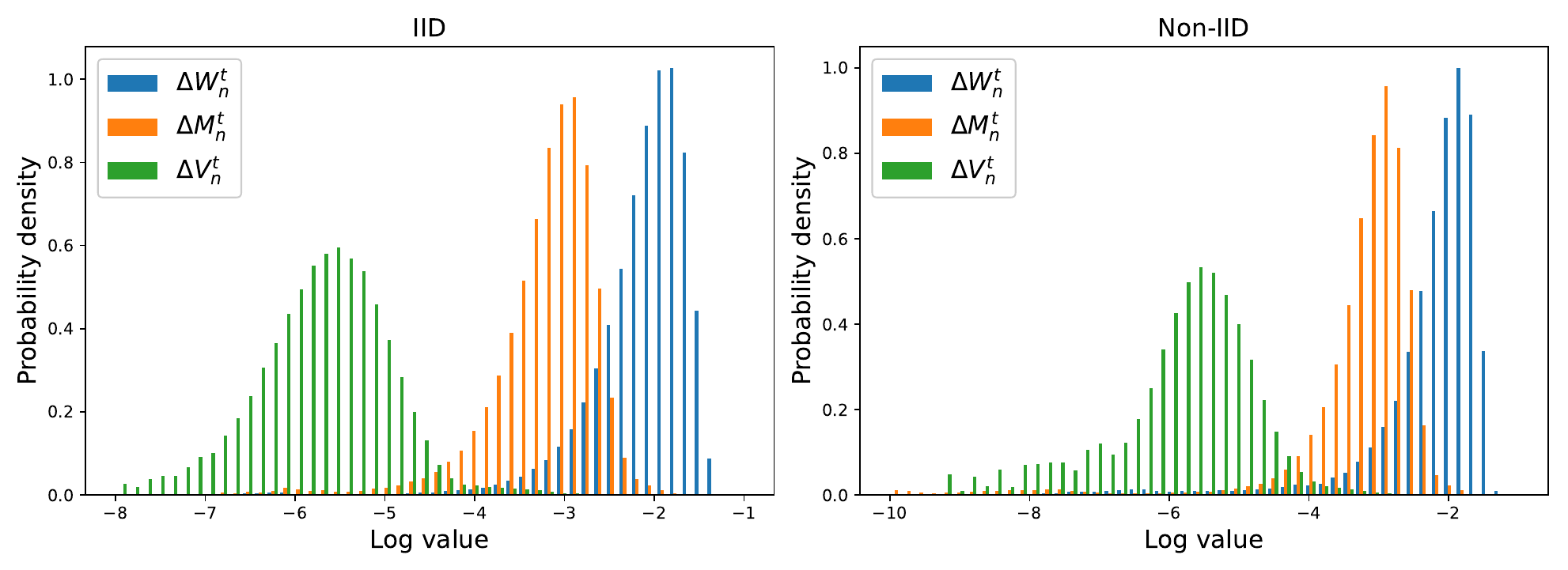}\vspace{-4pt}
		\caption{CNN on Fashion-MNIST}
		\label{fig1:1}
	\end{subfigure}
	\begin{subfigure}[!t]{0.4\textwidth}
		\includegraphics[width=\textwidth]{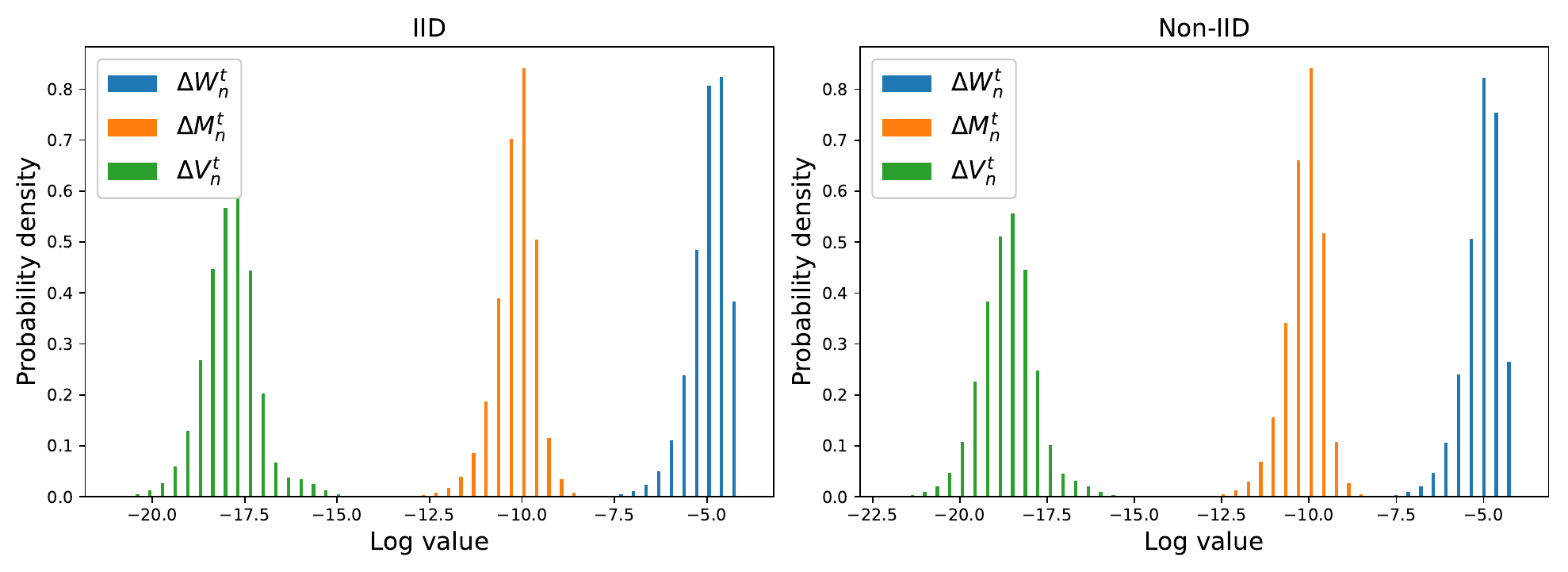}\vspace{-4pt}
		\caption{ResNet-18 on SVHN}
		\label{fig1:2}
	\end{subfigure}	
	\begin{subfigure}[!t]{0.4\textwidth}
		\includegraphics[width=\textwidth]{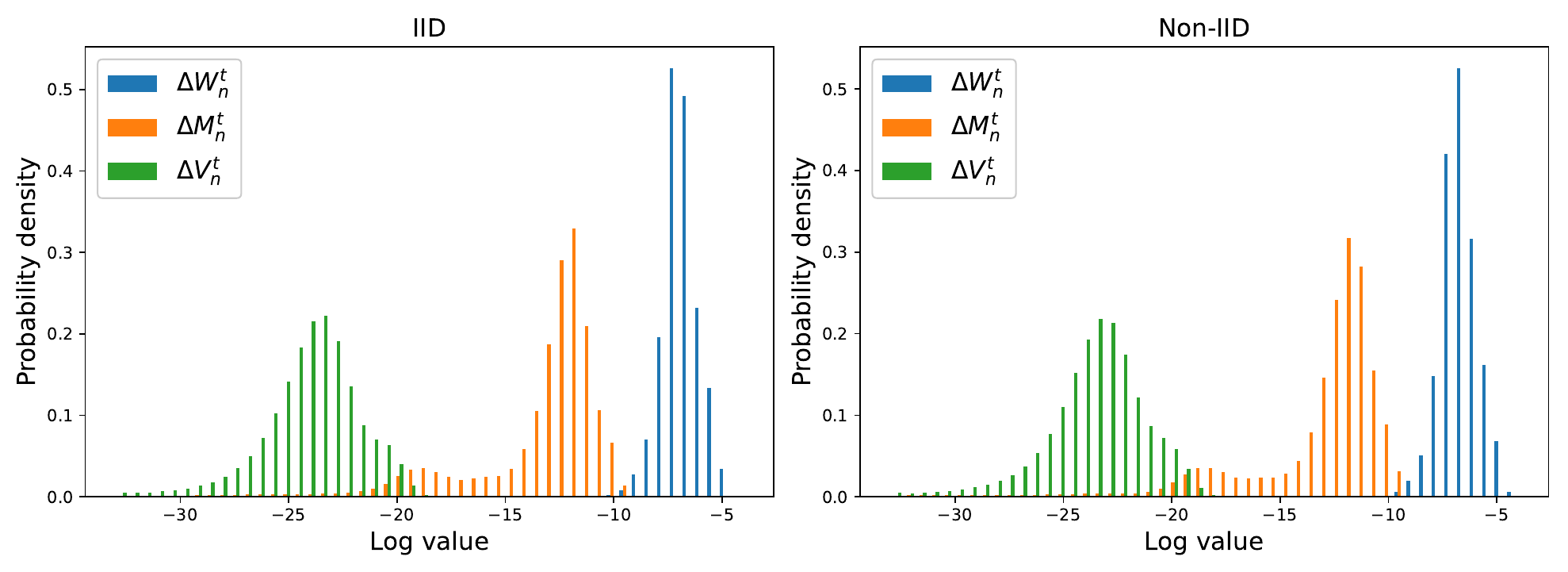}\vspace{-4pt}
		\caption{VGG-11 on CIFAR-10}
		\label{fig1:3}
	\end{subfigure}
	\caption{Probability density of the log value of $\Delta\boldsymbol{W}_n^t$, $\Delta\boldsymbol{M}_n^t$ and $\Delta\boldsymbol{V}_n^t$ on different models and datasets.}
	\label{fig1}
\end{figure}

To demonstrate the magnitudes of model parameter and moment estimate updates, Fig.~\ref{fig1} compares the probability density of their logarithmic values on different datasets and models. From Fig.~\ref{fig1}, we can observe that the logarithms of $\Delta\boldsymbol{W}_n^t$, $\Delta\boldsymbol{M}_n^t$ and $\Delta\boldsymbol{V}_n^t$ appear to approximately follow normal distributions with different means and variances, and the logarithmic values indicate that  $\Delta\boldsymbol{W}_n^t>\Delta\boldsymbol{M}_n^t>\Delta\boldsymbol{V}_n^t$. Specifically, for CNN on Fashion-MNIST, Fig.~\ref{fig1} (a) shows that the logarithms of $\Delta\boldsymbol{W}_n^t$, $\Delta\boldsymbol{M}_n^t$ and $\Delta\boldsymbol{V}_n^t$ exhibit approximate normal distributions centered around $-2$, $-3$ and $-6$, respectively. For ResNet-18 on SVHN, the logarithms of $\Delta\boldsymbol{W}_n^t$, $\Delta\boldsymbol{M}_n^t$ and $\Delta\boldsymbol{V}_n^t$ span ranges of $-7.5$ to $-4$, $-12.5$ to $-8$, and $-20$ to $-15$, respectively. For VGG-11 on CIFAR-10, the logarithmic values span ranges of $-10$ to $-5$, $-25$ to $-10$, and $-35$ to $-20$, respectively. The experimental results show that model parameter updates exhibit significantly larger magnitudes than moment estimate updates, corroborating prior theoretical and empirical analyses of Adam\cite{DBLP:journals/iotj/MillsHM20, DBLP:conf/icann/BockW19, DBLP:conf/icml/LiuSLHHC21}. This finding supports our theoretical analysis and the proposed SSM design of focusing on sparsifying model parameter updates.

\subsubsection{Performance Comparison versus Communication Overhead with Baseline Algorithms}
\begin{figure*}[!t]
	\centering
	\begin{subfigure}[!t]{0.3\textwidth}
		\includegraphics[width=\textwidth]{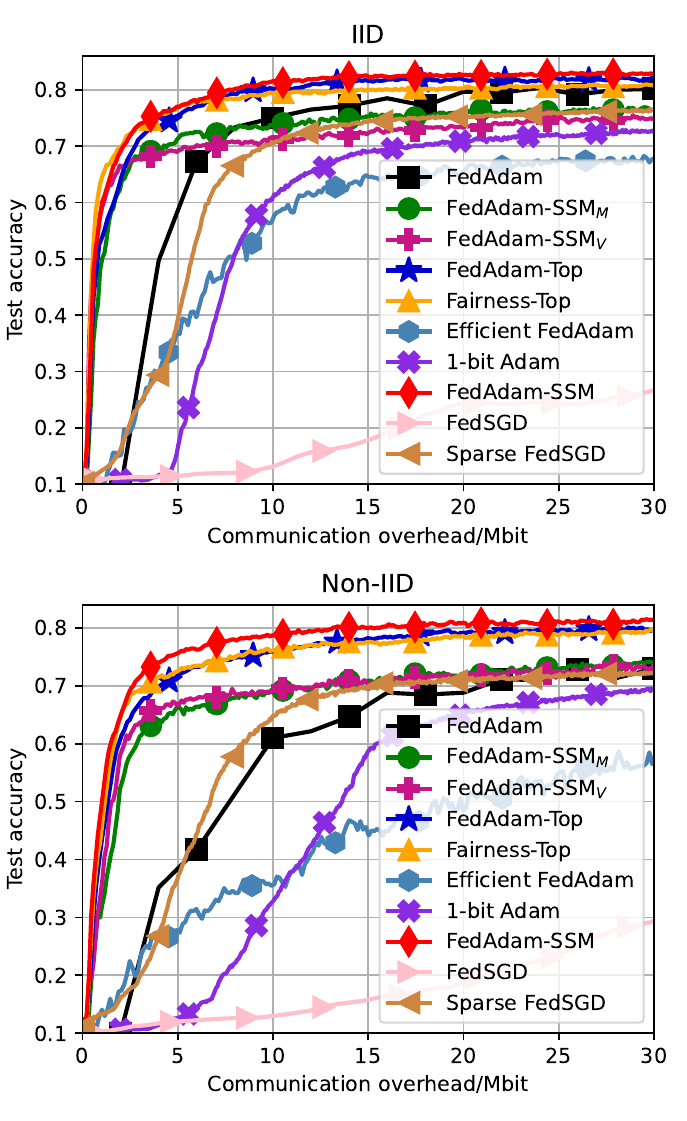}
		\caption{CNN on Fashion-MNIST}
		\label{fig2:1}
	\end{subfigure}
	\begin{subfigure}[!t]{0.3\textwidth}
		\includegraphics[width=\textwidth]{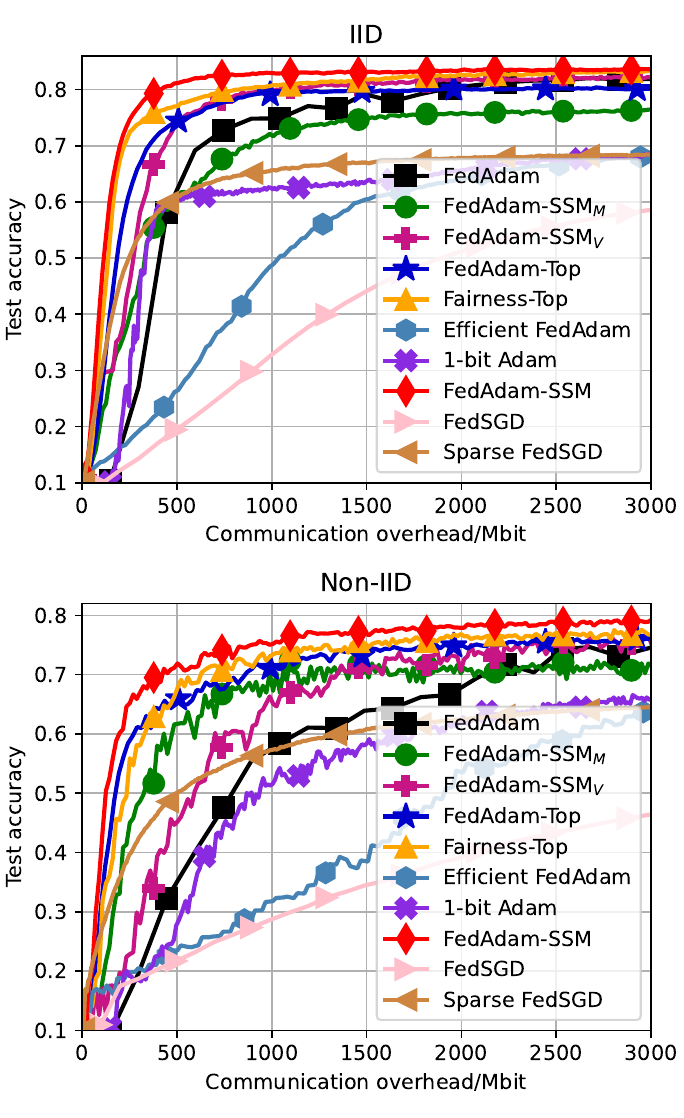}
		\caption{ResNet-18 on SVHN}
		\label{fig2:2}
	\end{subfigure}
	\begin{subfigure}[!t]{0.3\textwidth}
		\includegraphics[width=\textwidth]{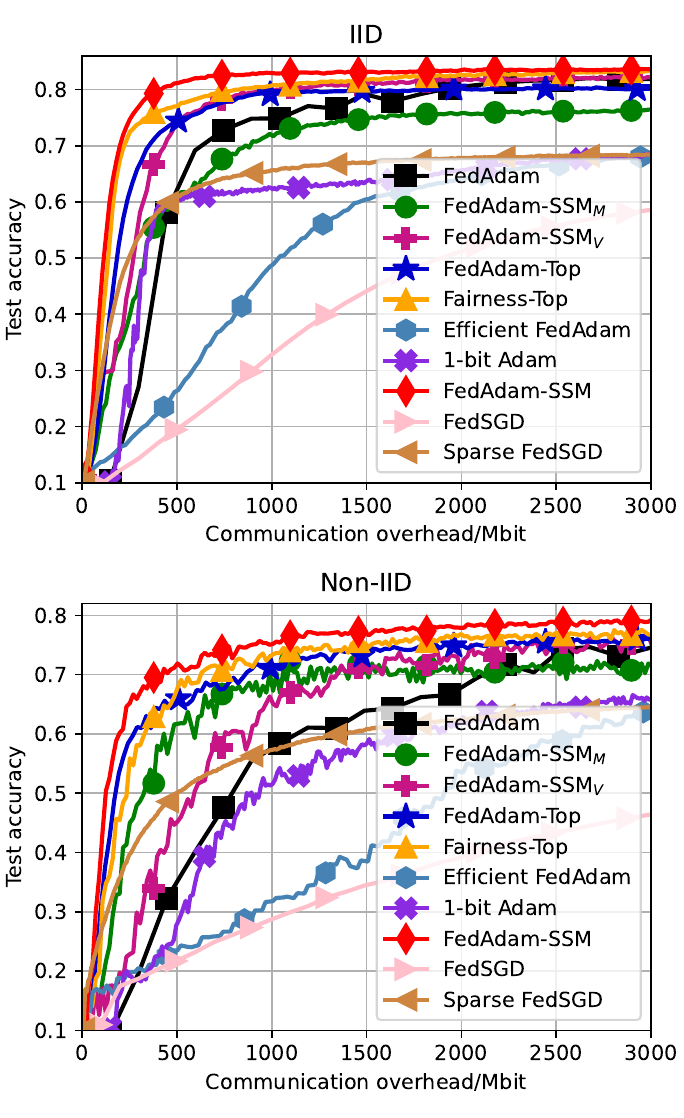}
		\caption{VGG-11 on CIFAR-10}
		\label{fig2:3}
	\end{subfigure}
	\caption{Comparison of model accuracy versus communication overhead between FedAdam-SSM and baseline algorithms.}
	\label{fig2}
\end{figure*}\begin{table*}[!t]
	\caption{Convergence rate of FedAdam-SSM and baseline algorithms. “Acc.” represents the target test accuracy. “Comm.” represents the minimum communication overhead required to achieve the target test accuracy. $\infty$ means it is impossible to achieve the target test accuracy during the training process.\label{table1}}
	\centering
	\renewcommand{\arraystretch}{1.15}
	\begin{tabular}{!{\vrule width 0pt}cccccccc!{\vrule width 0pt}}
		\noalign{\hrule height 0.5pt}
		~&~&\multicolumn{2}{c}{CNN on Fashion-MNIST}&\multicolumn{2}{c}{ResNet on SVHN}&\multicolumn{2}{c}{VGG-11 on CIFAR-10}\\\noalign{\hrule height 0.5pt}
		Setting&Algorithm&Acc. (\%)&Comm. (Mbit)&Acc. (\%)&Comm. (Mbit)&Acc. (\%)&Comm. (Mbit)\\\noalign{\hrule height 0.5pt}
		\multirow{8}*{IID}&\textbf{FedAdam-SSM}&\multirow{8}*{80.4}&\textbf{8.15}&\multirow{8}*{80.2}&\textbf{591.0}&\multirow{8}*{62.4}&\textbf{2073.0}\\
		~&FedAdam-Top&~&11.3 (1.39$\times$)&~&1959.1 (3.31$\times$)&~&11605.0 (5.59$\times$)\\	
		~&Fairness-Top&~&17.1 (2.09$\times$)&~&1623.0 (2.74$\times$)&~&4327.0 (2.09$\times$)\\	
		~&$\text{FedAdam-SSM}_M$&~&$\infty$&~&$\infty$&~&11651.0 (5.62$\times$)\\	
		~&$\text{FedAdam-SSM}_V$&~&$\infty$&~&1870.7 (3.16$\times$)&~&$\infty$\\	
		~&FedAdam&~&24.0 (2.94$\times$)&~&2641.8 (4.47$\times$)&~&18559.0 (8.95$\times$)\\	
		~&1-bit Adam&~&$\infty$&~&$\infty$&~&$\infty$\\	
		~&Efficient Adam&~&$\infty$&~&$\infty$&~&$\infty$\\\noalign{\hrule height 0.5pt}
		\multirow{8}*{Non-IID}&\textbf{FedAdam-SSM}&\multirow{8}*{79.8}&\textbf{13.0}&\multirow{8}*{74.4}&\textbf{737.3}&\multirow{8}*{56.2}&\textbf{8161.0}\\
		~&FedAdam-Top&~&24.5 (1.88$\times$)&~&1521.1 (2.06$\times$)&~&11172.0 (1.37$\times$)\\	
		~&Fairness-Top&~&31.5 (2.42$\times$)&~&1059.3 (1.44$\times$)&~&8966.0 (1.10$\times$)\\	
		~&$\text{FedAdam-SSM}_M$&~&$\infty$&~&$\infty$&~&17951.0 (2.19$\times$)\\	
		~&$\text{FedAdam-SSM}_V$&~&$\infty$&~&2042.1 (2.77$\times$)&~&$\infty$\\	
		~&FedAdam&~&70.0 (5.38$\times$)&~&2556.0 (3.47$\times$)&~&27400.0 (3.36$\times$)\\	
		~&1-bit Adam&~&$\infty$&~&$\infty$&~&$\infty$\\	
		~&Efficient Adam&~&$\infty$&~&$\infty$&~&$\infty$\\\noalign{\hrule height 0.5pt}
	\end{tabular}
\end{table*}

Fig.~\ref{fig2} compares test accuracy versus communication overhead between FedAdam-SSM and baseline algorithms. For CNN on Fashion-MNIST, the improvement in test accuracy compared to FedAdam-Top, Fairness-Top, $\text{FedAdam-SSM}_M$, $\text{FedAdam-SSM}_V$, FedAdam, 1-bit Adam and Efficient Adam is approximately 1.3\%, 2.9\%, 8.2\%, 10.9\%, 3.7\%, 14.4\%, 21.8\% in the IID setting, and 1.8\%, 2.5\%, 9.7\%, 10.7\%, 11.7\%, 17.3\%, 39.3\% in the Non-IID setting. For ResNet on SVHN, the improvement in test accuracy is approximately 4.5\%, 1.1\%, 9.6\%, 1.9\%, 2.3\%, 22\%, 23.1\% in the IID setting, and 4.3\%, 2.9\%, 10.7\%, 5.1\%, 6.4\%, 20.2\%, 24.9\% in the Non-IID setting. For VGG-11 on CIFAR-10, the improvement in test accuracy is approximately 4.1\%, 1.5\%, 3.8\%, 10.7\%, 4.8\%, 15.2\%, 14.4\% in the IID setting, and 2.7\%, 5.9\%, 6.1\%, 18.4\%, 5.9\%, 36.5\%, 39.6\% in the Non-IID setting. Table \ref{table1} compares the convergence rate of FedAdam-SSM and baseline algorithms. 

From Fig.~\ref{fig2} and Table \ref{table1}, we can observe that in terms of both IID and Non-IID data distribution settings, FedAdam-SSM outperforms baseline algorithms both on model accuracy and convergence rate. The observations are listed as follows. Firstly, FedAdam-SSM outperforms its counterparts among the sparse FedAdam algorithms both on test accuracy and convergence rate, including $\text{FedAdam-SSM}_M$ and $\text{FedAdam-SSM}_V$. Secondly, Fig.~\ref{fig2} shows that FedAdam-SSM significantly outperforms quantized FedAdam baselines, i.e., 1-bit Adam and Efficient Adam, in terms of test accuracy and convergence rate. Thirdly, it can be observed that FedAdam-SSM prevails slightly over FedAdam-Top and Fairness-Top in terms of test accuracy and convergence rate. Furthermore, FedAdam-SSM reduces the computational complexity by over 66.6\% compared to both FedAdam-Top and Fairness-Top. Specifically, FedAdam-Top employs the Top-$k$ sparsifier to separately sparsify the local updates of model parameters and moment estimates, while FedAdam-SSM employs an SSM to sparsify the local updates of model parameters and moment estimates. The computational complexity orders of FedAdam-SSM, FedAdam-Top, and Fairness-Top are $\mathcal{O}\left(d\log(k)\right)$, $\mathcal{O}\left(3d\log(k)\right)$, and $\mathcal{O}\left(9dk\right)$, respectively.

In addition, we can observe that given the same communication overhead, FedAdam-SSM significantly outperforms FedSGD and Sparse FedSGD in terms of both test accuracy and convergence rate. Note that FedSGD and sparse FedSGD use the SGD optimizer with a fixed learning rate for local model updates, whereas FedAdam and FedAdam-SSM employ Adam which dynamically tunes adaptive per-parameter learning rates using first and second moment estimates. On the one hand, FedSGD and Sparse FedSGD only require the transmission of local model parameter updates to the centralized server for aggregation. In contrast, FedAdam and FedAdam-SSM necessitate the transmission of not only local model parameter updates but also local moment estimate updates, leading to higher communication overhead per communication round. On the other hand, despite this increased communication overhead, Adam’s adaptive per-parameter learning rate optimization enables it to significantly outperform SGD in terms of test accuracy versus communication overhead. This finding highlights the critical role of Adam’s adaptive learning rate optimization in enhancing the efficiency of FL.

\subsubsection{Performance Comparison versus Communication Round with Baseline Algorithms}
\begin{figure}[!t]
	\centering
	\begin{subfigure}[!t]{0.45\textwidth}
		\includegraphics[width=\textwidth]{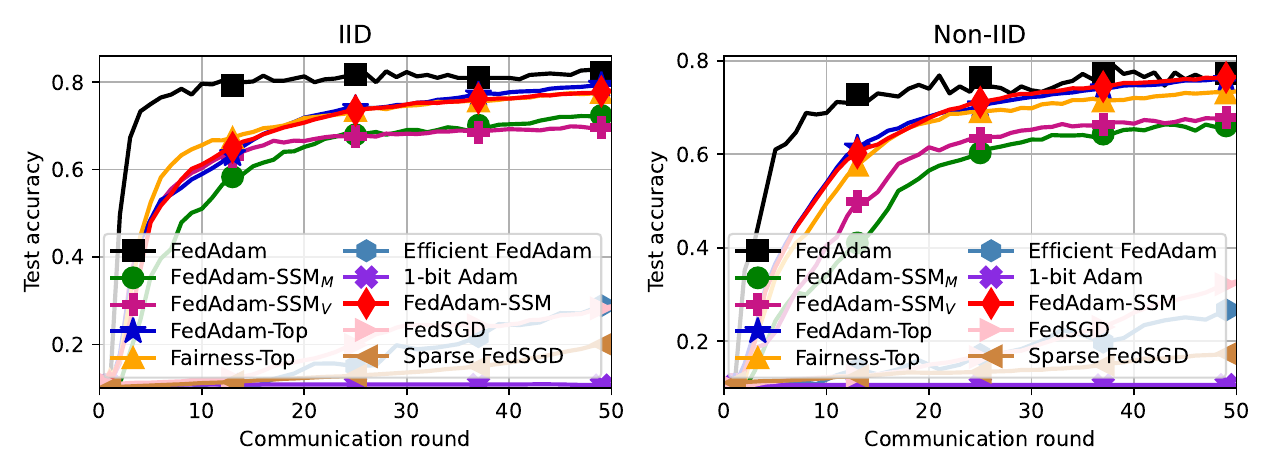}\vspace{-5pt}
		\caption{CNN on Fashion-MNIST}
		\label{fig33:1}
	\end{subfigure}\\\vspace{-1pt}
	\begin{subfigure}[!t]{0.45\textwidth}
		\includegraphics[width=\textwidth]{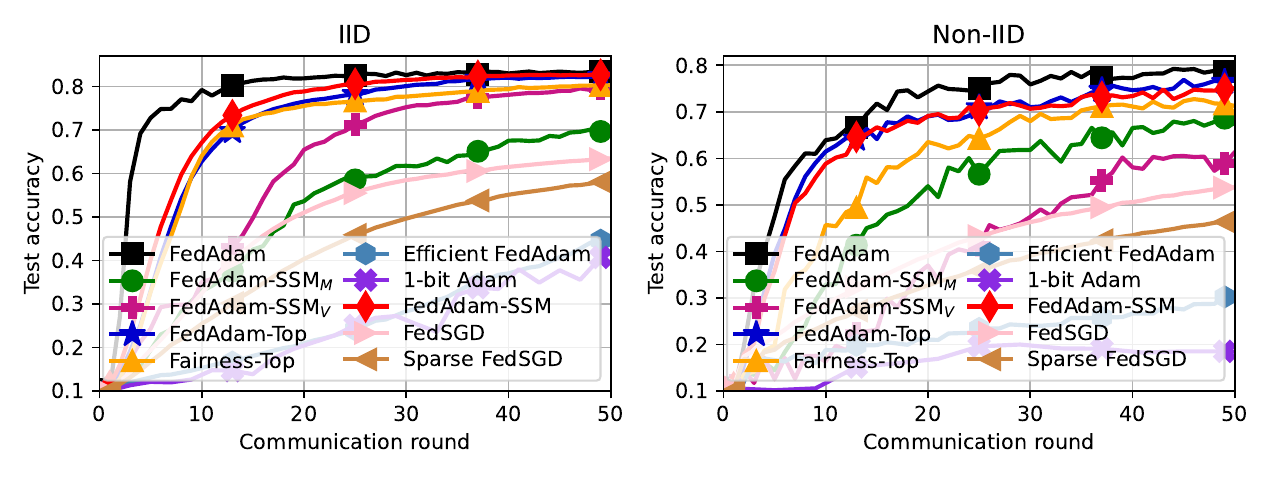}\vspace{-5pt}
		\caption{ResNet-18 on SVHN}
		\label{fig33:2}
	\end{subfigure}\\\vspace{-1pt}
	\begin{subfigure}[!t]{0.45\textwidth}
		\includegraphics[width=\textwidth]{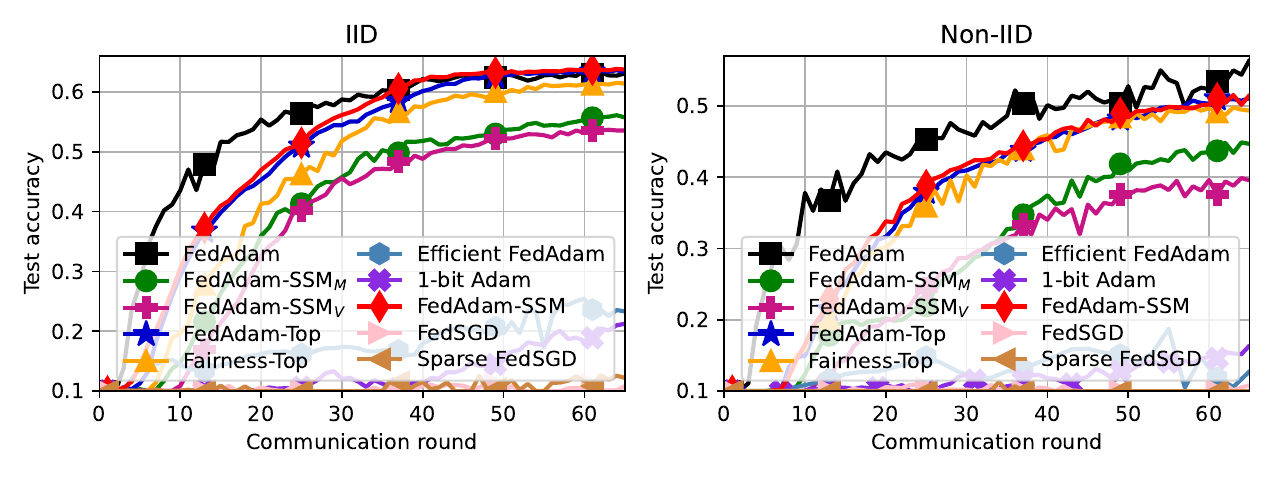}\vspace{-5pt}
		\caption{VGG-11 on CIFAR-10}
		\label{fig33:3}
	\end{subfigure}
	\caption{Comparison of model accuracy versus communication round between FedAdam-SSM and baseline algorithms.}
	\label{fig33}
\end{figure}Fig.~\ref{fig33} compares the model accuracy versus communication round between FedAdam-SSM and baselines. Firstly, we can observe that FedAdam demonstrates superior model performance in terms of both test accuracy and convergence rate compared to its sparse and quantized counterparts, including FedAdam-SSM, $\text{FedAdam-SSM}_M$, $\text{FedAdam-SSM}_V$, FedAdam-Top, Fairness-top, 1-bit Adam, and Efficient Adam. This observation aligns with theoretical expectations, as FedAdam preserves full update information for both local model parameters and moment estimates, avoiding compression errors introduced by sparsification or quantization techniques.

Secondly, FedAdam-SSM outperforms $\text{FedAdam-SSM}_M$, $\text{FedAdam-SSM}_V$, and Fairness-Top in both test accuracy and convergence rate. Note that FedAdam-SSM, $\text{FedAdam-SSM}_M$, $\text{FedAdam-SSM}_V$, and Fairness-Top employ different SSMs to sparsify the updates of local model parameters and moment estimates, maintaining identical uplink communication overhead of $\mathcal{O}(3kq + d)$ per communication round. The superior training performance of FedAdam-SSM can be attributed to its shared sparse mask (SSM) design. Specifically, with the same sparsification ratio, our proposed SSM can effectively reduce sparsification error compared to $\text{FedAdam-SSM}_M$, $\text{FedAdam-SSM}_V$, and Fairness-Top. This observation aligns well with \textbf{Theorem} \ref{theorem_1}, which establishes that the proposed SSM minimizes the deviation between centralized Adam and sparse FedAdam.

Thirdly, it can be observed that FedAdam-Top slightly outperforms FedAdam-SSM in terms of test accuracy and convergence rate. This advantage stems from that FedAdam-Top employs Top-$k$ sparsifier to separately sparsify the local updates of model parameters and moment estimates, while FedAdam-SSM employs an SSM to sparsify the local updates of model parameters and moment estimates. Although FedAdam-SSM could theoretically incur higher sparsification errors and thus reduced training performance per communication round compared to FedAdam-Top, our experimental results indicate minimal degradation in both test accuracy and convergence rate.

For quantized FedAdam baselines, FedAdam-SSM significantly outperforms 1-bit Adam and Efficient Adam in terms of both test accuracy and convergence rate. This performance gap primarily arises from that 1-bit Adam and Efficient Adam restrict the number of local epochs to 1, resulting in a slow convergence rate per communication round. Additionally, the extremely frequent exchanges between distributed devices and the centralized server can amplify compression errors in the training process, resulting in inferior test accuracy.

In addition, Fig.~\ref{fig33} demonstrates that FedAdam-SSM significantly outperforms sparse FedSGD in both test accuracy and convergence rate. This superior performance validates the effectiveness of the adaptive learning rate mechanism with our proposed SSM design. Despite potential omission of some significant elements in the local updates of model parameters and moment estimates due to sparsification, FedAdam-SSM successfully retains a significant portion of these critical elements in their sparse updates. This preserved update information is then used to calculate the adaptive per-parameter learning rate for the updates of local model parameters in subsequent local training epochs, playing a crucial role in preserving Adam’s capacity to adaptively tune per-parameter learning rates.
\begin{figure}[!t]
	\centering
	\begin{subfigure}[!t]{0.45\textwidth}
		\includegraphics[width=\textwidth]{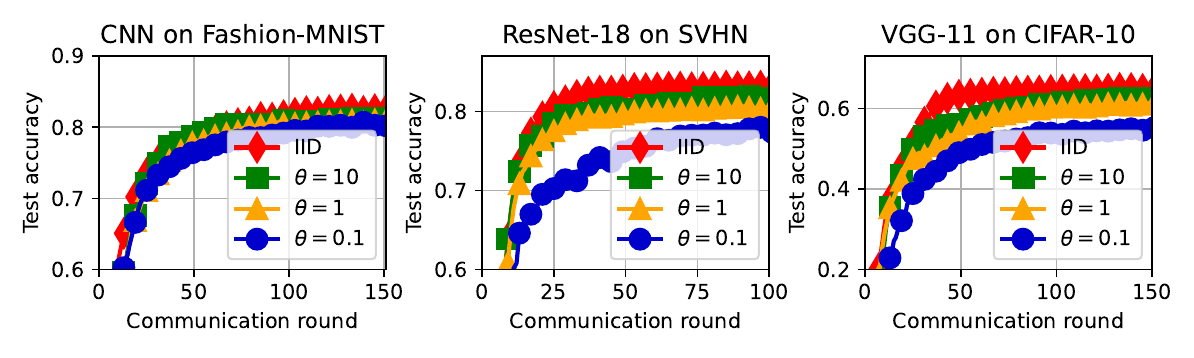}\vspace{-5pt}
		\caption{FedAdam-SSM}
	\label{fig44:1}
	\end{subfigure}\\
	\begin{subfigure}[!t]{0.45\textwidth}
		\includegraphics[width=\textwidth]{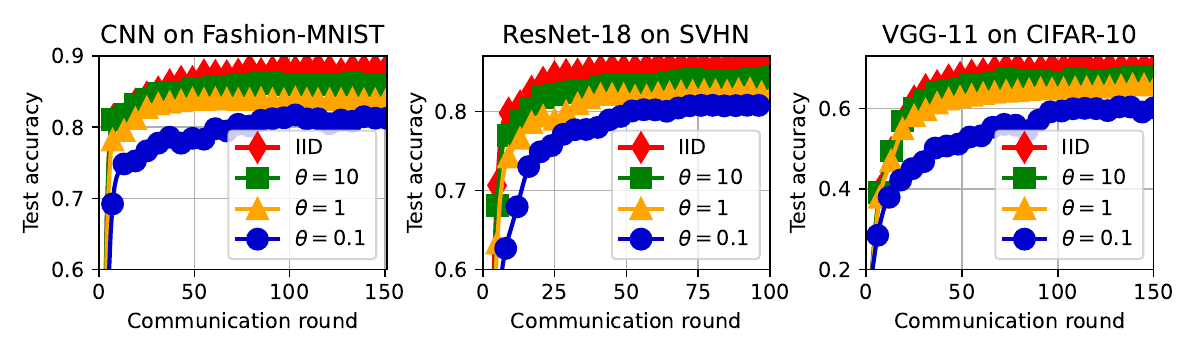}\vspace{-5pt}
		\caption{FedAdam}
	\label{fig44:2}
	\end{subfigure}
	\caption{Model accuracy of FedAdam-SSM and FedAdam for different data distribution on different models and datasets.}
	\label{fig44}
\end{figure}

\subsubsection{Sensitivity to Hyperparameters}
\begin{figure}[!t]
	\centering
	\begin{subfigure}[!t]{0.45\textwidth}
		\includegraphics[width=\textwidth]{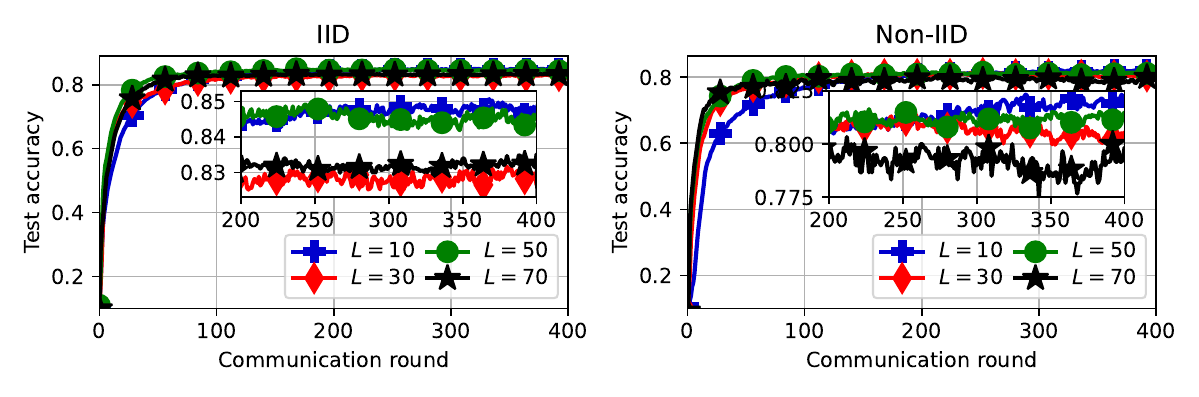}\vspace{-5pt}
		\caption{CNN on Fashion-MNIST}
		\label{fig3:1}
	\end{subfigure}\\\vspace{-1pt}
	\begin{subfigure}[!t]{0.45\textwidth}
		\includegraphics[width=\textwidth]{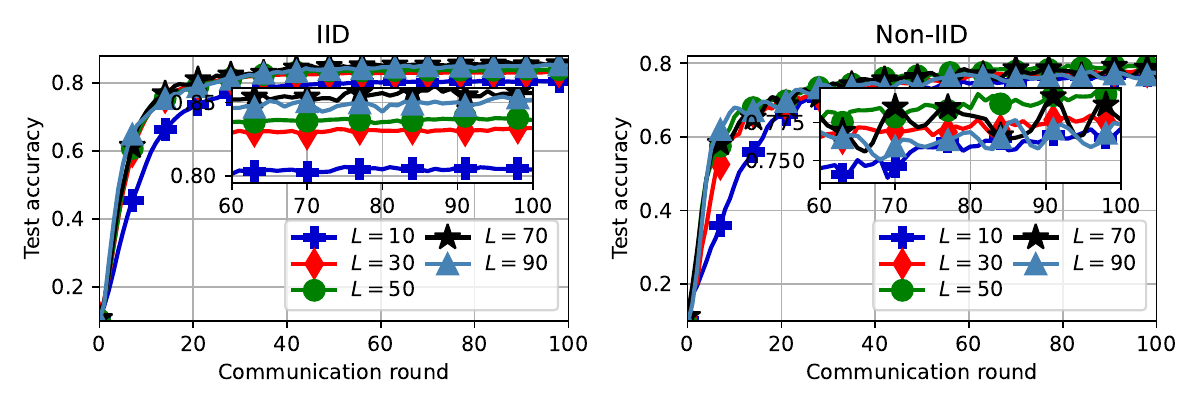}\vspace{-5pt}
		\caption{ResNet-18 on SVHN}
		\label{fig3:3}
	\end{subfigure}\\\vspace{-1pt}
	\begin{subfigure}[!t]{0.45\textwidth}
		\includegraphics[width=\textwidth]{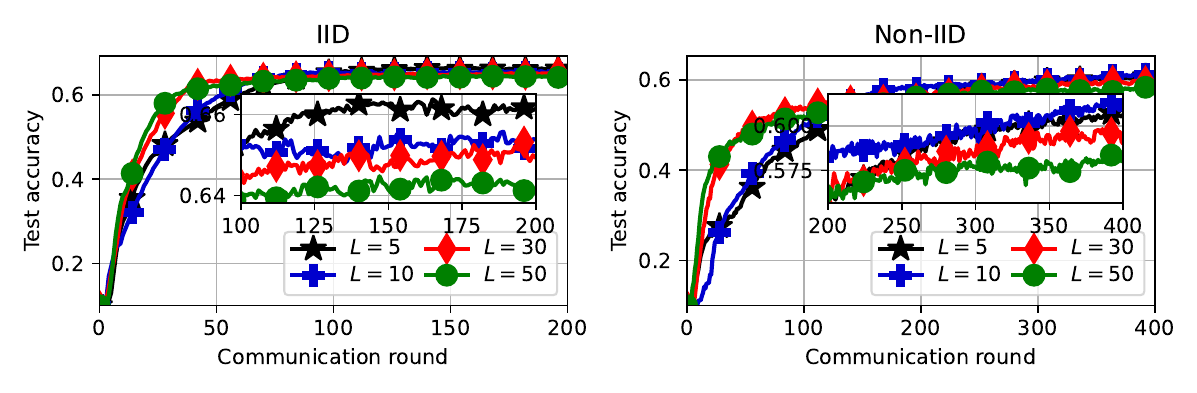}\vspace{-5pt}
		\caption{VGG-11 on CIFAR-10}
		\label{fig3:2}
	\end{subfigure}
	\caption{Model accuracy of FedAdam-SSM for different local epoch $L$ on different models and datasets.}
	\label{fig3}
\end{figure}
\begin{figure}[!t]
	\centering
	\begin{subfigure}[!t]{0.45\textwidth}
		\includegraphics[width=\textwidth]{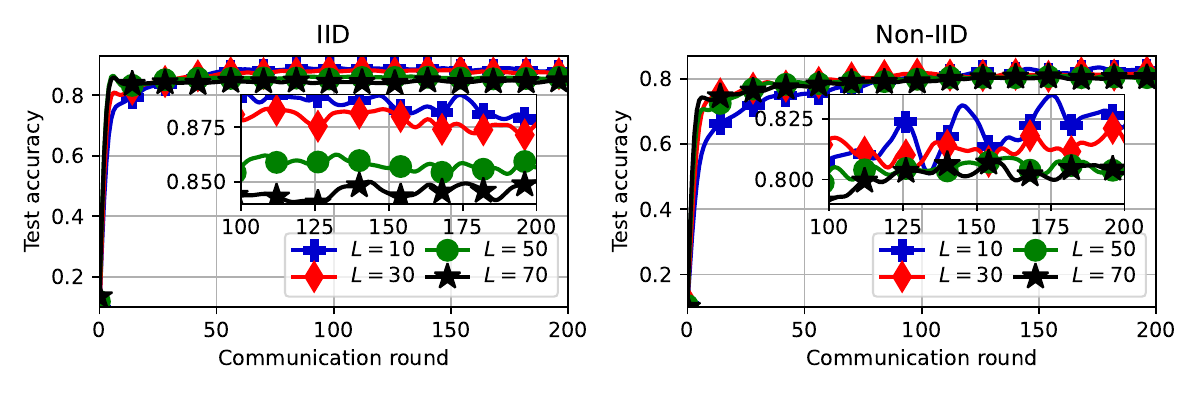}\vspace{-5pt}
		\caption{CNN on Fashion-MNIST}
		\label{fig35:1}
	\end{subfigure}\\\vspace{-1pt}
	\begin{subfigure}[!t]{0.45\textwidth}
		\includegraphics[width=\textwidth]{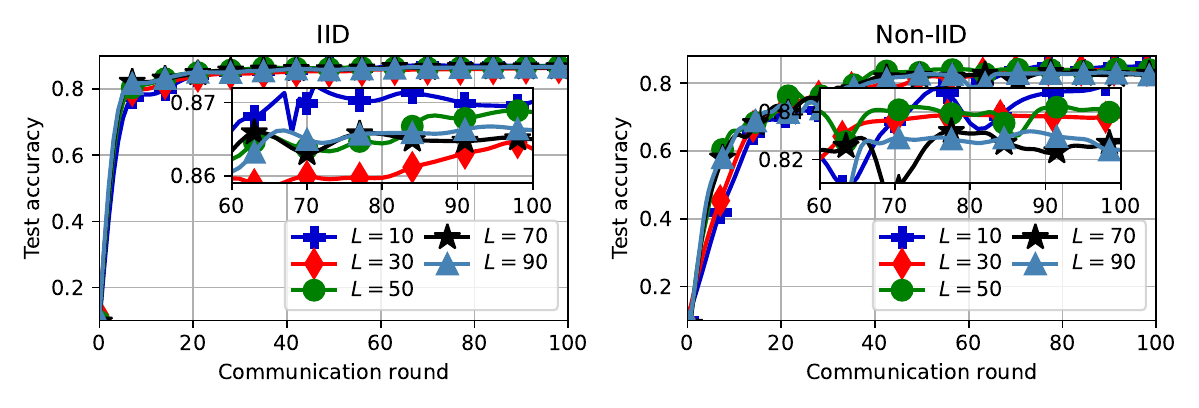}\vspace{-5pt}
		\caption{ResNet-18 on SVHN}
		\label{fig35:3}
	\end{subfigure}\\\vspace{-1pt}
	\begin{subfigure}[!t]{0.45\textwidth}
		\includegraphics[width=\textwidth]{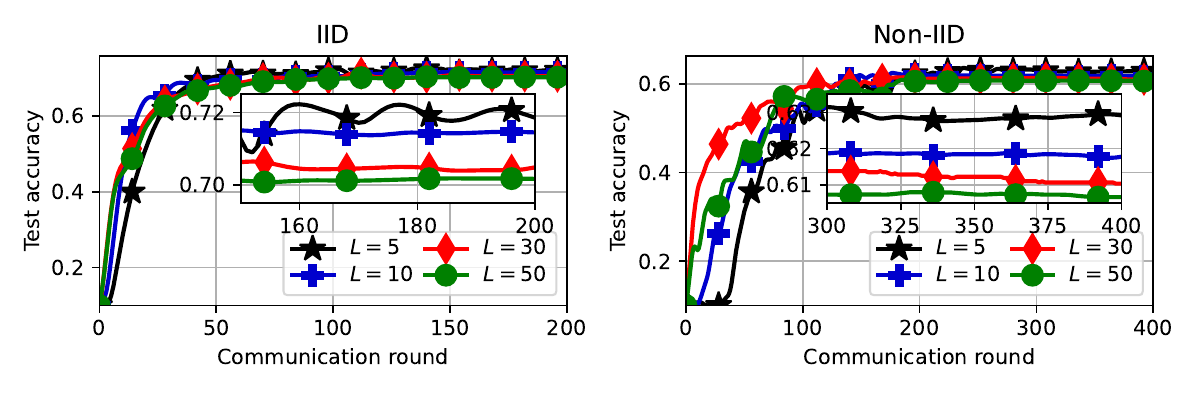}\vspace{-5pt}
		\caption{VGG-11 on CIFAR-10}
		\label{fig35:2}
	\end{subfigure}
	\caption{Model accuracy of FedAdam for different local epoch $L$ on different models and datasets.}
	\label{fig35}
\end{figure}\begin{figure}[!t]
\centering\vspace{-11pt}
\begin{subfigure}[!t]{0.45\textwidth}
	\includegraphics[width=\textwidth]{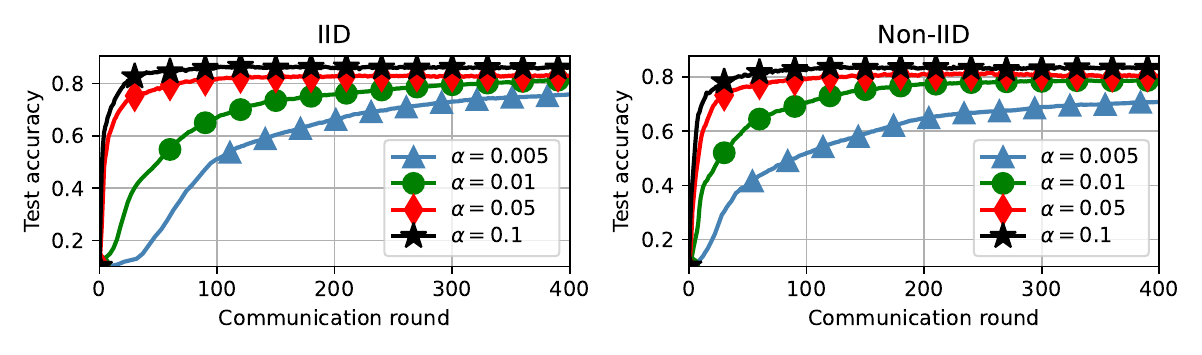}\vspace{-5pt}
	\caption{CNN on Fashion-MNIST}
	\label{fig5:1}
\end{subfigure}\\\vspace{-1pt}
\begin{subfigure}[!t]{0.45\textwidth}
	\includegraphics[width=\textwidth]{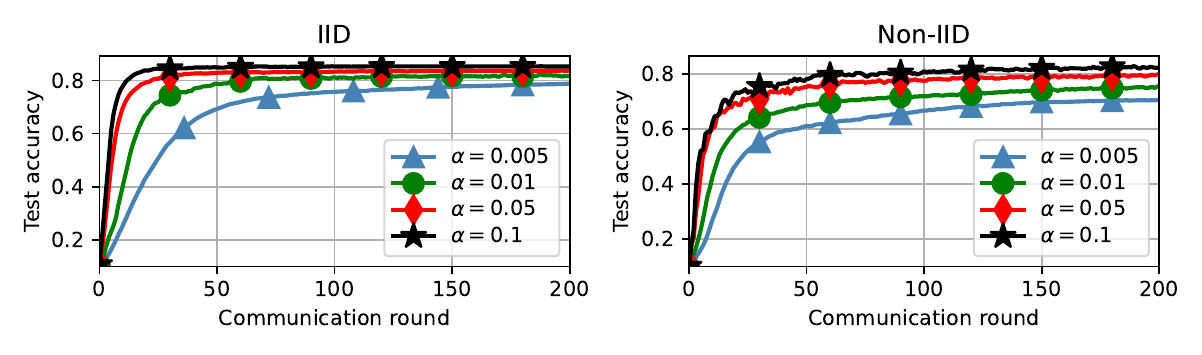}\vspace{-5pt}
	\caption{ResNet-18 on SVHN}
	\label{fig5:3}
\end{subfigure}\\\vspace{-1pt}
\begin{subfigure}[!t]{0.45\textwidth}
	\includegraphics[width=\textwidth]{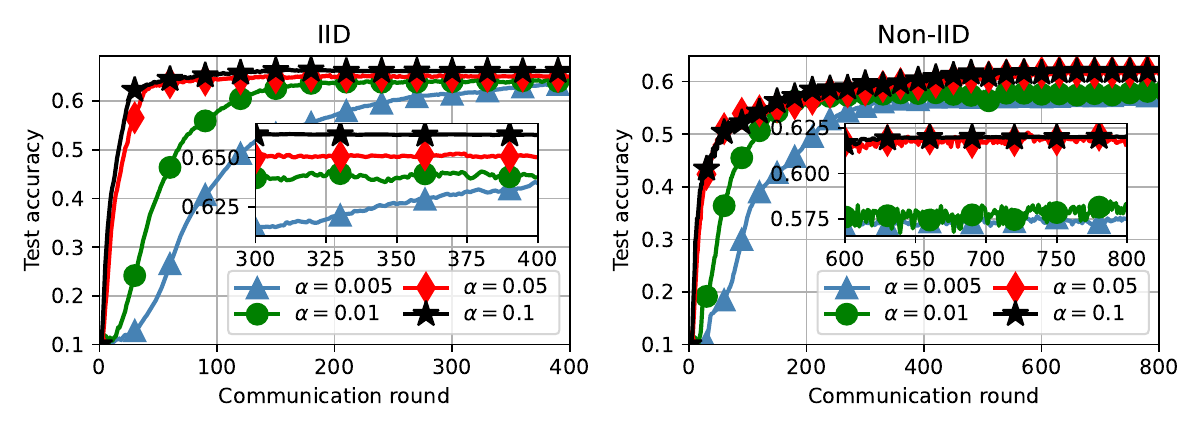}\vspace{-5pt}
	\caption{VGG-11 on CIFAR-10}
	\label{fig5:2}
\end{subfigure}
\caption{Model accuracy of FedAdam-SSM for different learning rate $\eta$ on different models and datasets.}
\label{fig5}
\end{figure}

\textbf{Data imbalance} characterizes the heterogeneity of both the data distribution across distributed devices and the feature distributions within individual devices. Fig.~\ref{fig44}(a) demonstrates that FedAdam-SSM exhibits superior test accuracy and faster convergence rate as the degree of data imbalance decreases. This observation is consistent with our theoretical analysis presented in \textbf{Theorem} \ref{theorem_2}, which establishes that reduced local and global variances $\sigma_l$ and $\sigma_g$ can lead to improved training performance. This can be attributed to the fact that imbalanced data distributions can cause shifted gradients on local datasets, resulting in increased local and global variances, and thereby degraded model accuracy. 

Fig.~\ref{fig44}(b) demonstrates that FedAdam-SSM is less sensitive to accuracy degradation as the degree of non-IID data distribution increases compared to FedAdam. On one hand, this observation empirically validates our theoretical analysis that the proposed shared sparse mask design introduces no further vulnerability to data imbalance beyond this fundamental challenge of FL. On the other hand, the reduced sensitivity arises from sparsification focusing parameter updates on most critical coordinates, thus helps reduce local model drift caused by data heterogeneity.

The \textbf{local epoch} measures the number of iterations of local gradient descents. Fig.~\ref{fig3} plots the test accuracy of FedAdam-SSM for different values of local epoch $L$. These results reveal that for FedAdam-SSM, the convergence rate exhibits a decreasing trend as $L$ increases, while the test accuracy shows an initial increasing trend followed by a subsequent decrease as $L$ increases. This is due to the fact that increasing the local epoch $L$ can lead to a local minimizer and thereby improve the convergence rate. On the other hand, increasing $L$ too much can lead to a strong local model drift, which compromises the model accuracy and slows down the convergence. This observeation highlights a trade-off in selecting the appropriate value of $L$ to balance the convergence rate and model accuracy, which aligns with \textbf{Proposition} \ref{proposition_6}. For comparison, Fig.~\ref{fig35} shows that vanilla FedAdam exhibits increasing convergence rate and decreasing model accuracy as $L$ increases, which is consistent with our analysis in \textbf{Remark} \ref{fsdghjklkjyhtg}. Additionally, FedAdam-SSM exhibits greater robustness to local epochs compared to FedAdam. Its robustness arises from sparsification focusing parameter updates on critical coordinates, thus reducing local model drift caused by local training. 

\textbf{Learning rate} determines the step size in each gradient descent iteration while moving towards the minimum of the loss function. Fig.~\ref{fig5} plots the test accuracy of FedAdam-SSM for different learning rate $\eta$ on different models and datasets. We can observe that on the one hand, increasing the learning rate $\eta$ can improve the test accuracy and speed up the convergence. On the other hand, a too large $\eta$ compromises the model accuracy and consequently degrades the convergence rate. This is due to the fact that a small $\eta$ can cause the optimizer to converge slowly or get stuck in plateaus or undesirable local minima, while a large $\eta$ can cause drastic updates which can lead to divergent behaviour. The above discussion highlights a guidance for selecting an appropriate value of $\eta$ to optimize the training performance, which is consistent with \textbf{Theorem} \ref{theorem_3}. 

\begin{figure}[!t]
	\centering
	\begin{subfigure}[!t]{0.45\textwidth}
		\includegraphics[width=\textwidth]{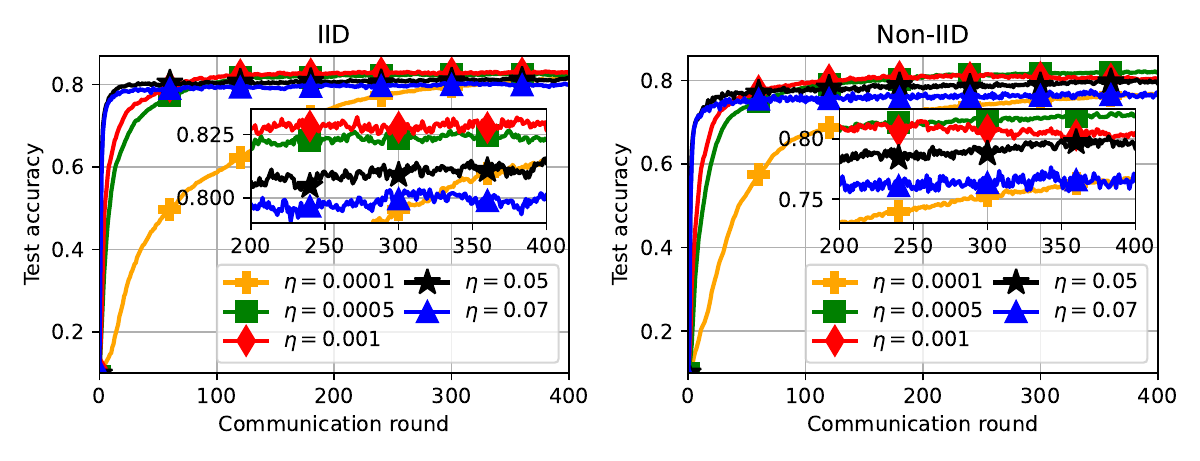}\vspace{-5pt}
		\caption{CNN on Fashion-MNIST}
		\label{fig4:1}
	\end{subfigure}\\\vspace{-1pt}
	\begin{subfigure}[!t]{0.45\textwidth}
		\includegraphics[width=\textwidth]{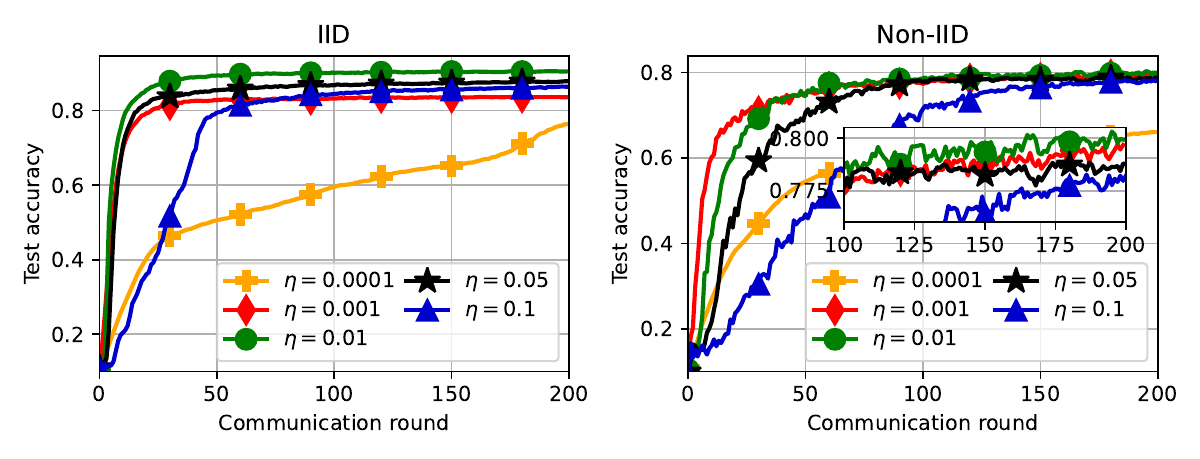}\vspace{-5pt}
		\caption{ResNet-18 on SVHN}
		\label{fig4:3}
	\end{subfigure}\\\vspace{-1pt}
	\begin{subfigure}[!t]{0.45\textwidth}
		\includegraphics[width=\textwidth]{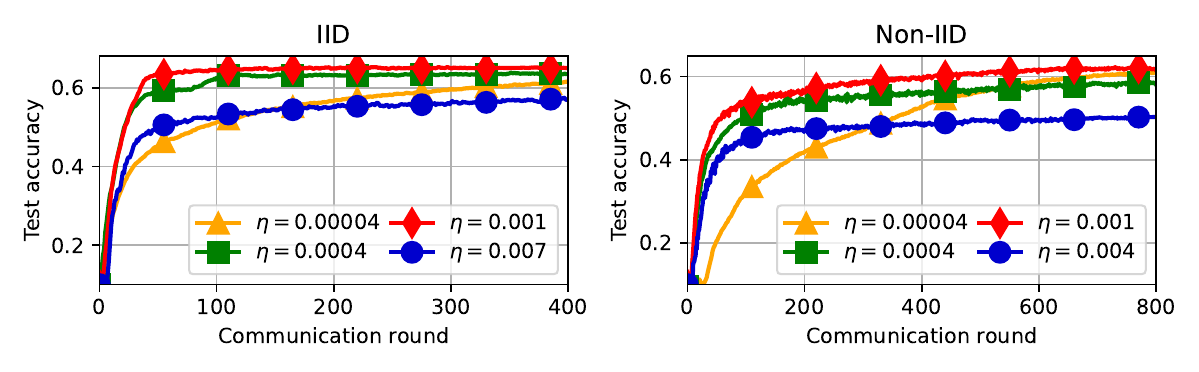}\vspace{-5pt}
		\caption{VGG-11 on CIFAR-10}
		\label{fig4:2}
	\end{subfigure}
	\caption{Model accuracy of FedAdam-SSM for different sparsification ratio $\alpha$ on different models and datasets.}
	\label{fig4}
\end{figure}

\textbf{Sparsification ratio} measures the sparsification error in the training process. Fig.~\ref{fig4} plots the test accuracy of FedAdam-SSM for different sparsification ratio $\alpha$ on different models and datasets. We can see that increasing the learning rate $\eta$ can improve the test accuracy and speed up the convergence. This aligns with the findings of \textbf{Theorem} \ref{theorem_2}, demonstrating that a decreased sparsification ratio can lead to a diminished sparsification error and thereby enhanced training performance.

\section{Conclusion}\label{section7}
In this paper, we have proposed a novel sparse FedAdam algorithm called FedAdam-SSM, which incorporates an SSM into the sparsification of the updates of local model parameters and moment estimates to reduce the uplink communication overhead. We have provided an upper bound on the deviation between the local model trained by FedAdam-SSM and the target model trained by centralized Adam, which is related to the sparsification error and imbalanced data distribution. Based on this deviation bound, we have optimized the SSM to mitigate the learning performance degradation caused by the sparsification error. We have provided convergence bounds for the proposed FedAdam-SSM with both non-convex and convex objective function settings. We have investigated the impact of local epoch, learning rate and sparsification ratio on the convergence rate of FedAdam-SSM, and provided guidances for selecting appropriate values of local epoch, learning rate and sparsification ratio to improve the training performance. Extensive experiments on Fashion-MNIST, SVHN and CIFAR-10 datasets have verified the theoretical analysis and showed that FedAdam-SSM outperforms baselines in terms of both convergence rate and test accuracy. This work represents the first attempt to design the SSM for the sparsification of local model updates and moment estimates in sparse FedAdam and to provide a corresponding theoretical analysis.

Several interesting directions immediately follow from this work. First, integrating adaptive aggregation methods into FedAdam-SSM presents significant potential for further addressing data imbalance challenges inherent in FL. Second, conducting large-scale evaluations across diverse model architectures and datasets would further substantiate and generalize our theoretical framework, especially beyond the current focus on Transformer architectures.

\ifCLASSOPTIONcaptionsoff
\newpage
\fi
\bibliographystyle{IEEEtran}
\bibliography{references}

\begin{thebibliography}{10}
\providecommand{\url}[1]{#1}
\csname url@samestyle\endcsname
\providecommand{\newblock}{\relax}
\providecommand{\bibinfo}[2]{#2}
\providecommand{\BIBentrySTDinterwordspacing}{\spaceskip=0pt\relax}
\providecommand{\BIBentryALTinterwordstretchfactor}{4}
\providecommand{\BIBentryALTinterwordspacing}{\spaceskip=\fontdimen2\font plus
\BIBentryALTinterwordstretchfactor\fontdimen3\font minus
  \fontdimen4\font\relax}
\providecommand{\BIBforeignlanguage}[2]{{%
\expandafter\ifx\csname l@#1\endcsname\relax
\typeout{** WARNING: IEEEtran.bst: No hyphenation pattern has been}%
\typeout{** loaded for the language `#1'. Using the pattern for}%
\typeout{** the default language instead.}%
\else
\language=\csname l@#1\endcsname
\fi
#2}}
\providecommand{\BIBdecl}{\relax}
\BIBdecl

\bibitem{DBLP:journals/iotj/ChangLXCT21}
Z.~Chang, S.~Liu, X.~Xiong, Z.~Cai, and G.~Tu, ``A survey of recent advances in
  edge-computing-powered artificial intelligence of things,'' \emph{{IEEE}
  Internet Things J.}, vol.~8, no.~18, pp. 13\,849--13\,875, 2021.

\bibitem{DBLP:journals/corr/abs-2202-03402}
T.~D.~T. Nguyen and M.~T. Thai, ``Preserving privacy and security in federated
  learning,'' \emph{{IEEE/ACM} Trans. Netw.}, vol.~32, no.~1, pp. 833--843,
  2024.

\bibitem{DBLP:journals/tpds/LiSWDMSHP22}
J.~Li, Y.~Shao, K.~Wei, M.~Ding, C.~Ma, L.~Shi, Z.~Han, and H.~V. Poor,
  ``Blockchain assisted decentralized federated learning {(BLADE-FL):}
  performance analysis and resource allocation,'' \emph{{IEEE} Trans. Parallel
  Distributed Syst.}, vol.~33, no.~10, pp. 2401--2415, 2022.

\bibitem{DBLP:journals/titb/AliNTK23}
M.~Ali, F.~Naeem, M.~Tariq, and G.~Kaddoum, ``Federated learning for privacy
  preservation in smart healthcare systems: {A} comprehensive survey,''
  \emph{{IEEE} J. Biomed. Health Informatics}, vol.~27, no.~2, pp. 778--789,
  2023.

\bibitem{DBLP:journals/jsac/DengLMWSDC23}
X.~Deng, J.~Li, C.~Ma, K.~Wei, L.~Shi, M.~Ding, and W.~Chen, ``Low-latency
  federated learning with {DNN} partition in distributed industrial iot
  networks,'' \emph{{IEEE} J. Sel. Areas Commun.}, vol.~41, no.~3, pp.
  755--775, 2023.

\bibitem{DBLP:journals/iotj/BianAXLLCWDG22}
J.~Bian, A.~A. Arafat, H.~Xiong, J.~Li, L.~Li, H.~Chen, J.~Wang, D.~Dou, and
  Z.~Guo, ``Machine learning in real-time internet of things ({IoT}) systems:
  {A} survey,'' \emph{{IEEE} Internet Things J.}, vol.~9, no.~11, pp.
  8364--8386, 2022.

\bibitem{DBLP:journals/tkde/WangFHHW22}
M.~Wang, W.~Fu, X.~He, S.~Hao, and X.~Wu, ``A survey on large-scale machine
  learning,'' \emph{{IEEE} Trans. Knowl. Data Eng.}, vol.~34, no.~6, pp.
  2574--2594, 2022.

\bibitem{DBLP:journals/corr/abs-2111-00856}
X.~He, F.~Xue, X.~Ren, and Y.~You, ``Large-scale deep learning optimizations:
  {A} comprehensive survey,'' \emph{CoRR}, vol. abs/2111.00856, 2021.

\bibitem{DBLP:journals/jmlr/TranRMF22}
B.~Tran, S.~Rossi, D.~Milios, and M.~Filippone, ``All you need is a good
  functional prior for bayesian deep learning,'' \emph{J. Mach. Learn. Res.},
  vol.~23, no.~74, pp. 3210--3265, 2022.

\bibitem{DBLP:journals/chinaf/WeiLMDSZCZ24}
K.~Wei, J.~Li, C.~Ma, M.~Ding, F.~Shu, H.~Zhao, W.~Chen, and H.~Zhu, ``Gradient
  sparsification for efficient wireless federated learning with differential
  privacy,'' \emph{Sci. China Inf. Sci.}, vol.~67, no.~4, p. 142303, 2024.

\bibitem{sami2024secure}
H.~U. Sami and B.~G{\"u}ler, ``Secure gradient aggregation with sparsification
  for resource-limited federated learning,'' \emph{{IEEE} Trans. Commun.},
  early access, 2024.

\bibitem{DBLP:journals/compsec/LuLLGY23}
S.~Lu, R.~Li, W.~Liu, C.~Guan, and X.~Yang, ``Top-\emph{k} sparsification with
  secure aggregation for privacy-preserving federated learning,'' \emph{Comput.
  Secur.}, vol. 124, p. 102993, 2023.

\bibitem{DBLP:conf/iclr/LinHM0D18}
Y.~Lin, S.~Han, H.~Mao, Y.~Wang, and B.~Dally, ``Deep gradient compression:
  Reducing the communication bandwidth for distributed training,'' in \emph{6th
  International Conference on Learning Representations, {ICLR} 2018, Vancouver,
  BC, Canada, April 30 - May 3, 2018, Conference Track Proceedings}.

\bibitem{DBLP:conf/nips/SahuDABCK21}
A.~N. Sahu, A.~Dutta, A.~M. Abdelmoniem, T.~Banerjee, M.~Canini, and P.~Kalnis,
  ``Rethinking gradient sparsification as total error minimization,'' in
  \emph{Advances in Neural Information Processing Systems 34: Annual Conference
  on Neural Information Processing Systems 2021, NeurIPS 2021, December 6-14,
  2021, virtual}, pp. 8133--8146.

\bibitem{DBLP:conf/icml/XuH22}
A.~Xu and H.~Huang, ``Detached error feedback for distributed {SGD} with random
  sparsification,'' in \emph{International Conference on Machine Learning,
  {ICML} 2022, 17-23 July 2022, Baltimore, Maryland, {USA}}, vol. 162, pp.
  24\,550--24\,575.

\bibitem{DBLP:journals/twc/OhLWNC24}
J.~Oh, D.~Lee, D.~Won, W.~Noh, and S.~Cho, ``Communication-efficient federated
  learning over-the-air with sparse one-bit quantization,'' \emph{{IEEE} Trans.
  Wirel. Commun.}, vol.~23, no.~10, pp. 15\,673--15\,689, 2024.

\bibitem{DBLP:conf/icdcs/HanWL20}
P.~Han, S.~Wang, and K.~K. Leung, ``Adaptive gradient sparsification for
  efficient federated learning: An online learning approach,'' in \emph{40th
  {IEEE} International Conference on Distributed Computing Systems, {ICDCS}
  2020, Singapore, November 29 - December 1, 2020}, pp. 300--310.

\bibitem{DBLP:journals/tmc/JiangXXWLQQ24}
Z.~Jiang, Y.~Xu, H.~Xu, Z.~Wang, J.~Liu, C.~Qian, and C.~Qiao, ``Computation
  and communication efficient federated learning with adaptive model pruning,''
  \emph{{IEEE} Trans. Mob. Comput.}, vol.~23, no.~3, pp. 2003--2021, 2024.

\bibitem{DBLP:journals/tpds/TangSLC23}
Z.~Tang, S.~Shi, B.~Li, and X.~Chu, ``{GossipFL}: {A} decentralized federated
  learning framework with sparsified and adaptive communication,'' \emph{{IEEE}
  Trans. Parallel Distributed Syst.}, vol.~34, no.~3, pp. 909--922, 2023.

\bibitem{DBLP:journals/tvt/SunMH20}
H.~Sun, X.~Ma, and R.~Q. Hu, ``Adaptive federated learning with gradient
  compression in uplink {NOMA},'' \emph{{IEEE} Trans. Veh. Technol.}, vol.~69,
  no.~12, pp. 16\,325--16\,329, 2020.

\bibitem{DBLP:conf/globecom/ZhengDC23}
S.~Zheng, Y.~Dong, and X.~Chen, ``Efficient model compression via global
  sparsification for over-the-air federated learning,'' in \emph{{IEEE}
  Globecom Workshops 2023, Kuala Lumpur, Malaysia, December 4-8, 2023}, pp.
  920--925.

\bibitem{DBLP:conf/ecai/ShiT0ZC20}
S.~Shi, Z.~Tang, Q.~Wang, K.~Zhao, and X.~Chu, ``Layer-wise adaptive gradient
  sparsification for distributed deep learning with convergence guarantees,''
  in \emph{{ECAI} 2020 - 24th European Conference on Artificial Intelligence,
  Santiago de Compostela, Spain, August 29 - September 8, 2020}, vol. 325, pp.
  1467--1474.

\bibitem{DBLP:conf/iccspa/BeitollahiLL22}
M.~Beitollahi, M.~Liu, and N.~Lu, ``{DSFL:} dynamic sparsification for
  federated learning,'' in \emph{5th International Conference on
  Communications, Signal Processing, and their Applications, {ICCSPA} 2022,
  Cairo, Egypt, December 27-29, 2022}, pp. 1--6.

\bibitem{DBLP:conf/ciss/JinDX23}
R.~Jin, P.~Dai, and K.~Xiong, ``Communication-efficient federated learning with
  channel-aware sparsification over wireless networks,'' in \emph{57th Annual
  Conference on Information Sciences and Systems, {CISS} 2023, Baltimore, MD,
  USA, March 22-24, 2023}, pp. 1--6.

\bibitem{hu2023flexible}
Y.~Hu, T.~Liu, C.~Yang, Y.~Huang, and S.~Suo, ``A flexible model compression
  and resource allocation scheme for federated learning,'' \emph{{IEEE} Trans.
  mach. learn. commun. netw}, vol.~1, pp. 168--184, 2023.

\bibitem{chen2024latency}
X.~Chen, A.~Wang, X.~Deng, and J.~Gui, ``Latency-efficient wireless federated
  learning with spasification and quantization for heterogeneous devices,''
  \emph{{IEEE} Internet Things J.}, early access, 2024.

\bibitem{DBLP:journals/tpds/ZhouYL22}
Y.~Zhou, Q.~Ye, and J.~Lv, ``Communication-efficient federated learning with
  compensated overlap-{FedAvg},'' \emph{{IEEE} Trans. Parallel Distributed
  Syst.}, vol.~33, no.~1, pp. 192--205, 2022.

\bibitem{DBLP:conf/icpp/DinhTNBZZ20}
C.~T. Dinh, N.~H. Tran, T.~D. Nguyen, W.~Bao, A.~Y. Zomaya, and B.~B. Zhou,
  ``Federated learning with proximal stochastic variance reduced gradient
  algorithms,'' in \emph{{ICPP} 2020: 49th International Conference on Parallel
  Processing, Edmonton, AB, Canada, August 17-20, 2020}, pp. 48:1--48:11.

\bibitem{DBLP:journals/tsp/WangFLZ23}
B.~Wang, J.~Fang, H.~Li, and B.~Zeng, ``Communication-efficient federated
  learning: {A} variance-reduced stochastic approach with adaptive
  sparsification,'' \emph{{IEEE} Trans. Signal Process.}, vol.~71, pp.
  3562--3576, 2023.

\bibitem{DBLP:conf/aistats/ChenGSY21}
T.~Chen, Z.~Guo, Y.~Sun, and W.~Yin, ``{CADA:} communication-adaptive
  distributed adam,'' in \emph{The 24th International Conference on Artificial
  Intelligence and Statistics, {AISTATS} 2021, April 13-15, 2021, Virtual
  Event}, vol. 130, pp. 613--621.

\bibitem{DBLP:conf/icc/Li0WSLY20}
S.~Li, Q.~Qi, J.~Wang, H.~Sun, Y.~Li, and F.~R. Yu, ``{GGS:} general gradient
  sparsification for federated learning in edge computing\({}^{\mbox{*}}\),''
  in \emph{2020 {IEEE} International Conference on Communications, {ICC} 2020,
  Dublin, Ireland, June 7-11, 2020}, pp. 1--7.

\bibitem{DBLP:conf/icdm/XianHH22}
W.~Xian, F.~Huang, and H.~Huang, ``Communication-efficient adam-type algorithms
  for distributed data mining,'' in \emph{{IEEE} International Conference on
  Data Mining, {ICDM} 2022, Orlando, FL, USA, November 28 - Dec. 1, 2022},
  X.~Zhu, S.~Ranka, M.~T. Thai, T.~Washio, and X.~Wu, Eds., pp. 1245--1250.

\bibitem{DBLP:conf/ijcai/HuGG21}
R.~Hu, Y.~Gong, and Y.~Guo, ``Federated learning with sparsification-amplified
  privacy and adaptive optimization,'' in \emph{Proceedings of the Thirtieth
  International Joint Conference on Artificial Intelligence, {IJCAI} 2021,
  Virtual Event / Montreal, Canada, 19-27 August 2021}, pp. 1463--1469.

\bibitem{DBLP:conf/icml/WangLC22}
Y.~Wang, L.~Lin, and J.~Chen, ``Communication-efficient adaptive federated
  learning,'' in \emph{International Conference on Machine Learning, {ICML}
  2022, 17-23 July 2022, Baltimore, Maryland, {USA}}, vol. 162, pp.
  22\,802--22\,838.

\bibitem{DBLP:journals/corr/abs-2201-02664}
N.~Mitchell, J.~Ball{\'{e}}, Z.~Charles, and J.~Kone{\v{c}}n{\'y}, ``Optimizing
  the communication-accuracy trade-off in federated learning with
  rate-distortion theory,'' \emph{CoRR}, vol. abs/2201.02664, 2022.

\bibitem{DBLP:conf/icml/Li023}
X.~Li and P.~Li, ``Analysis of error feedback in federated non-convex
  optimization with biased compression: Fast convergence and partial
  participation,'' in \emph{International Conference on Machine Learning,
  {ICML} 2023, 23-29 July 2023, Honolulu, Hawaii, {USA}}, vol. 202, pp.
  19\,638--19\,688.

\bibitem{DBLP:journals/tist/ChenSHL21}
C.~Chen, L.~Shen, H.~Huang, and W.~Liu, ``Quantized {Adam} with error
  feedback,'' \emph{{ACM} Trans. Intell. Syst. Technol.}, vol.~12, no.~56, pp.
  1--26, 2021.

\bibitem{DBLP:journals/corr/abs-2205-14473}
C.~Chen, L.~Shen, W.~Liu, and Z.~Luo, ``Efficient-adam: Communication-efficient
  distributed adam,'' \emph{{IEEE} Trans. Signal Process.}, vol.~71, pp.
  3257--3266, 2023.

\bibitem{DBLP:conf/icml/TangGARLLLZH21}
H.~Tang, S.~Gan, A.~A. Awan, S.~Rajbhandari, C.~Li, X.~Lian, J.~Liu, C.~Zhang,
  and Y.~He, ``1-bit {Adam}: Communication efficient large-scale training with
  {Adam}'s convergence speed,'' in \emph{Proceedings of the 38th International
  Conference on Machine Learning, {ICML} 2021, 18-24 July 2021, Virtual Event},
  vol. 139, pp. 10\,118--10\,129.

\bibitem{DBLP:journals/iotj/MillsHM20}
J.~Mills, J.~Hu, and G.~Min, ``Communication-efficient federated learning for
  wireless edge intelligence in iot,'' \emph{{IEEE} Internet Things J.},
  vol.~7, no.~7, pp. 5986--5994, 2020.

\bibitem{DBLP:journals/iotj/LiLZD22}
Y.~Li, W.~Li, B.~Zhang, and J.~Du, ``Federated {Adam}-type algorithm for
  distributed optimization with lazy strategy,'' \emph{{IEEE} Internet Things
  J.}, vol.~9, no.~20, pp. 20\,519--20\,531, 2022.

\bibitem{DBLP:conf/icann/BockW19}
S.~Bock and M.~G. Wei{\ss}, ``Non-convergence and limit cycles in the {Adam}
  optimizer,'' in \emph{Artificial Neural Networks and Machine Learning -
  {ICANN} 2019: Deep Learning - 28th International Conference on Artificial
  Neural Networks, Munich, Germany, September 17-19, 2019, Proceedings, Part
  {II}}, vol. 11728, pp. 232--243.

\bibitem{DBLP:conf/icml/LiuSLHHC21}
Z.~Liu, Z.~Shen, S.~Li, K.~Helwegen, D.~Huang, and K.~Cheng, ``How do {Adam}
  and training strategies help {BNNs} optimization,'' in \emph{Proceedings of
  the 38th International Conference on Machine Learning, {ICML} 2021, 18-24
  July 2021, Virtual Event}, vol. 139, pp. 6936--6946.

\bibitem{DBLP:conf/icml/YurochkinAGGHK19}
M.~Yurochkin, M.~Agarwal, S.~Ghosh, K.~H. Greenewald, T.~N. Hoang, and
  Y.~Khazaeni, ``Bayesian nonparametric federated learning of neural
  networks,'' in \emph{Proceedings of the 36th International Conference on
  Machine Learning, {ICML} 2019, 9-15 June 2019, Long Beach, California,
  {USA}}, vol.~97, pp. 7252--7261.

\bibitem{DBLP:conf/iclr/WangYSPK20}
H.~Wang, M.~Yurochkin, Y.~Sun, D.~S. Papailiopoulos, and Y.~Khazaeni,
  ``Federated learning with matched averaging,'' in \emph{8th International
  Conference on Learning Representations, {ICLR} 2020, Addis Ababa, Ethiopia,
  April 26-30, 2020}.

\bibitem{DBLP:conf/isit/OzfaturaOG21}
E.~Ozfatura, K.~Ozfatura, and D.~G{\"{u}}nd{\"{u}}z, ``Time-correlated
  sparsification for communication-efficient federated learning,'' in
  \emph{{IEEE} International Symposium on Information Theory, {ISIT} 2021,
  Melbourne, Australia, July 12-20, 2021}, pp. 461--466.

\bibitem{DBLP:journals/tgcn/LinLCGH23}
X.~Lin, Y.~Liu, F.~Chen, X.~Ge, and Y.~Huang, ``Joint gradient sparsification
  and device scheduling for federated learning,'' \emph{{IEEE} Trans. Green
  Commun. Netw.}, vol.~7, no.~3, pp. 1407--1419, 2023.

\end{thebibliography}

\begin{IEEEbiography}[{\includegraphics[width=1in,height=1.25in,clip,keepaspectratio]{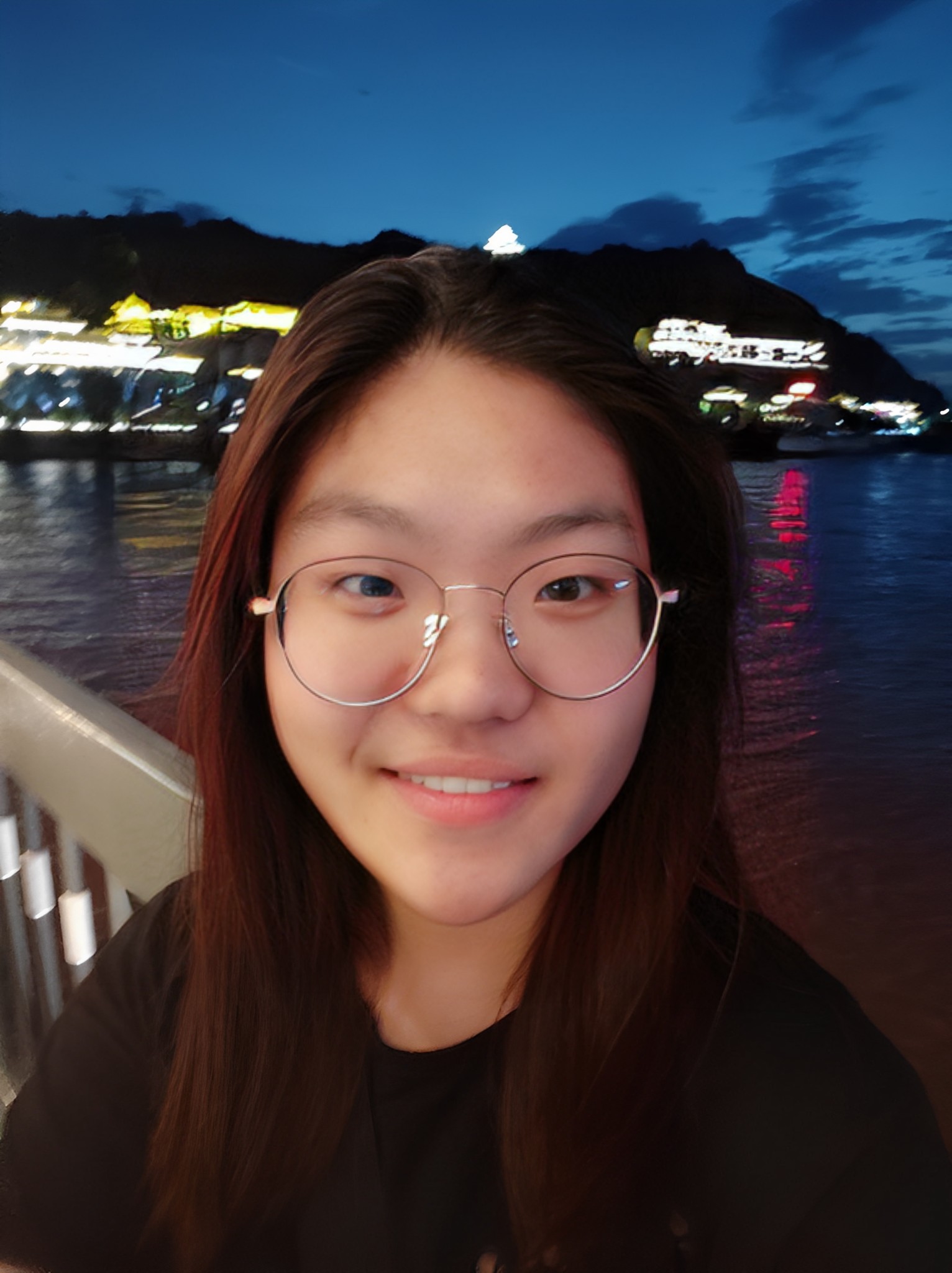}}]
	{Xiumei Deng} is currently a postdoctoral fellow at the Singapore University of Technology and Design. Before that, she received the Ph.D. degree in Information and Communications Engineering from Nanjing University of Science and Technology, China, in 2024, and the B.E. degree in Electronic Information Engineering from Nanjing University of Science and Technology, China, in 2018. Her research focuses on algorithm design and optimization for edge intelligence, with interests in edge generative AI, on-device large language models, federated learning, network resource optimization, and blockchain.
\end{IEEEbiography}

\begin{IEEEbiography}[{\includegraphics[width=1in,height=1.25in,clip,keepaspectratio]{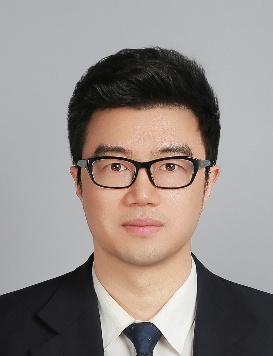}}]{Jun Li} (M’09-SM’16-F’25) received Ph.D. degree in Electronic Engineering from Shanghai Jiao Tong University, Shanghai, P. R. China in 2009. From January 2009 to June 2009, he worked in the Department of Research and Innovation, Alcatel Lucent Shanghai Bell as a Research Scientist. From June 2009 to April 2012, he was a Postdoctoral Fellow at the School of Electrical Engineering and Telecommunications, the University of New South Wales, Australia. From April 2012 to June 2015, he was a Research Fellow at the School of Electrical Engineering, the University of Sydney, Australia. From June 2015 to June 2024, he was a Professor at the School of Electronic and Optical Engineering, Nanjing University of Science and Technology, Nanjing, China. He is now a Professor at the School of Information Science and Engineering, Southeast University, Nanjing, China. He was a visiting professor at Princeton University from 2018 to 2019. His research interests include distributed intelligence, multiple agent reinforcement learning, and their applications in ultra-dense wireless networks, mobile edge computing, network privacy and security, and industrial Internet of Things. He has co-authored more than 300 papers in IEEE journals and conferences. He was serving as an editor of IEEE Transactions on Wireless Communication and TPC member for several flagship IEEE conferences.
\end{IEEEbiography}

\begin{IEEEbiography}[{\includegraphics[width=1in,height=1.25in,clip,keepaspectratio]{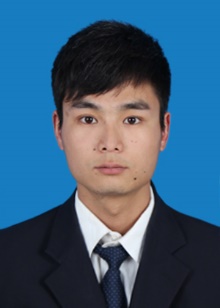}}]
	{Kang Wei} received his Ph.D. degree from Nanjing University of Science and Technology, Nanjing, China, in 2023. Before that, he received a B.S. degree in information engineering from Xidian University, Xian, China, in 2014. He is currently an associate professor at Southeast University. He has won the 2022 IEEE Signal Processing Society Best Paper Award and the 2022 Wiley China Open Science Author of the Year. He mainly focuses on privacy protection and optimization techniques for edge intelligence, including federated learning, differential privacy, and network resource allocation.
\end{IEEEbiography}

\begin{IEEEbiography}[{\includegraphics[width=1in,height=1.25in,clip,keepaspectratio]{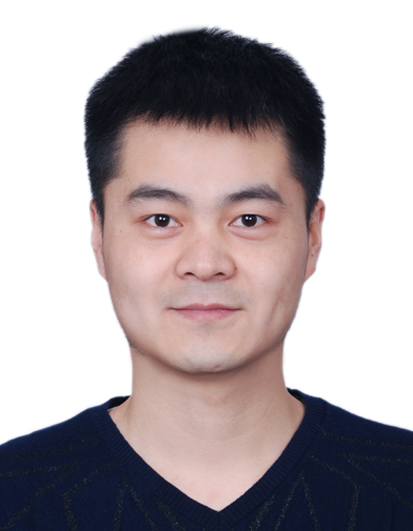}}]
	{Long Shi} (Senior Member, IEEE) received the Ph.D. degree in Electrical Engineering from the University of New South Wales, Sydney, Australia, in 2012. From 2013 to 2016, he was a Postdoctoral Fellow at the Institute of Network Coding, Chinese University of Hong Kong, China. From 2014 to 2017, he was a Lecturer at Nanjing University of Aeronautics and Astronautics, Nanjing, China. From 2017 to 2020, he was a Research Fellow at the Singapore University of Technology and Design. Now he is a Professor at the School of Electronic and Optical Engineering, Nanjing University of Science and Technology, Nanjing, China. His research interests include wireless communications, decentralized security, and edge intelligence. He is serving as an editor of IEEE Transactions on Cognitive Communications and Networking.
\end{IEEEbiography}

\begin{IEEEbiography}[{\includegraphics[width=1in,height=1.25in,clip,keepaspectratio]{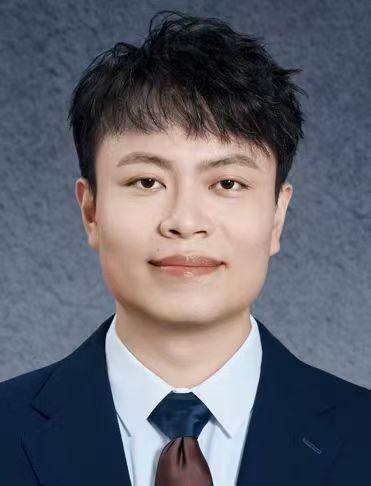}}]
	{Zehui Xiong} is currently a Full Professor with the School of Electronics, Electrical Engineering and Computer Science, Queen's University Belfast, United Kingdom. Prior to that, he was with Singapore University of Technology and Design, and Nanyang Technological University (NTU). He received his Ph.D. degree from NTU and was a visiting scholar with Princeton University and University of Waterloo. Recognized as a Clarivate Highly Cited Researcher, he has published over 250 peer-reviewed research papers in leading journals, with numerous Best Paper Awards from international flagship conferences. Featured in Forbes Asia 30U30, he serves as the Editor for many leading journals and Chair for numerous international conferences. His honors include the IEEE Asia Pacific Outstanding Young Researcher Award, IEEE VTS Early Career Award, IEEE Early Career Award for Excellence in Scalable Computing, IEEE Technical Committee on Blockchain and Distributed Ledger Technologies Early Career Award, IEEE Internet Technical Committee Early Achievement Award, IEEE TCSVC Rising Star Award, IEEE TCI Rising Star Award, IEEE TCCLD Rising Star Award, IEEE ComSoc Outstanding Paper Award, IEEE Best Land Transport Paper Award, IEEE Asia Pacific Outstanding Paper Award, IEEE CSIM Technical Committee Best Journal Paper Award, IEEE SPCC Technical Committee Best Paper Award, and IEEE Big Data Best Influential Conference Paper Award.
\end{IEEEbiography}

\begin{IEEEbiography}[{\includegraphics[width=1in,height=1.25in,clip,keepaspectratio]{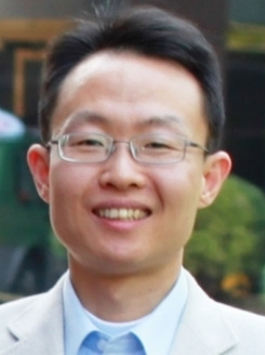}}]
	{Ming Ding} (IEEE M’12-SM’17) received the B.S. (with first-class Hons.) and M.S. degrees in electronics engineering from Shanghai Jiao Tong University (SJTU), China, and the Doctor of Philosophy (Ph.D.) degree in signal and information processing from SJTU, in 2004, 2007, and 2011, respectively. From April 2007 to September 2014, he worked at Sharp Laboratories of China as a Researcher/Senior Researcher/Principal Researcher. Currently, he is the Group Leader of the Privacy Technology Group at CSIRO’s Data61 in Sydney, NSW, Australia. Also, he is an Adjunct Professor at Swinburne University of Technology and University of Technology Sydney, Australia. His research interests include data privacy and security, machine learning and AI, and information technology. He has co-authored more than 300 papers in IEEE/ACM journals and conferences, all in recognized venues, and around 20 3GPP standardization contributions, as well as two books, i.e., “Multi-point Cooperative Communication Systems: Theory and Applications” (Springer, 2013) and “Fundamentals of Ultra-Dense Wireless Networks” (Cambridge University Press, 2022). Also, he holds 21 US patents and has co-invented another 100+ patents on 4G/5G technologies. Currently, he is an editor of IEEE Communications Surveys and Tutorials and IEEE Transactions on Network Science and Engineering. Besides, he has served as a guest editor/co-chair/co-tutor/TPC member for multiple IEEE top-tier journals/conferences and received several awards for his research work and professional services, including the prestigious IEEE Signal Processing Society Best Paper Award in 2022 and Highly Cited Researcher recognized by Clarivate Analytics in 2024.
\end{IEEEbiography}

\begin{IEEEbiography}[{\includegraphics[width=1in,height=1.25in,clip,keepaspectratio]{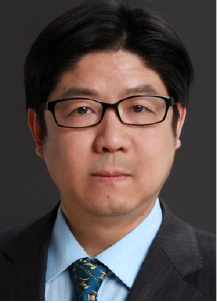}}]
	{Wen Chen} (M’03–SM’11) received BS and MS from Wuhan University, China in 1990 and 1993 respectively, and PhD from University of Electro-communications, Japan in 1999. He is now a tenured Professor with the Department of Electronic Engineering, Shanghai Jiao Tong University, China. He is a fellow of Chinese Institute of Electronics and the distinguished lecturers of IEEE Communications Society and IEEE Vehicular Technology Society. He also received Shanghai Natural Science Award in 2022. He is the Shanghai Chapter Chair of IEEE Vehicular Technology Society, a vice president of Shanghai Institute of Electronics, Editors of IEEE Transactions on Wireless Communications, IEEE Transactions on Communications, IEEE Access and IEEE Open Journal of Vehicular Technology. His research interests include multiple access, wireless AI and RIS communications. He has published more than 200 papers in IEEE journals with citations more than 11,000 in Google scholar. 
\end{IEEEbiography}

\begin{IEEEbiography}[{\includegraphics[width=1in,height=1.25in,clip,keepaspectratio]{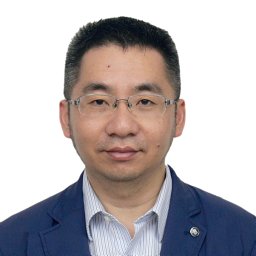}}]
	{Shi Jin} received the B.S. degree in communications engineering from the Guilin University of Electronic Technology, Guilin, China, in 1996, the M.S. degree from the Nanjing University of Posts and Telecommunications, Nanjing, China, in 2003, and the Ph.D. degree in information and communications engineering from Southeast University, Nanjing, in 2007. From June 2007 to October 2009, he was a Research Fellow with the Adastral Park Research Campus, University College London, London, U.K. He is currently with the Faculty of the National Mobile Communications Research Laboratory, Southeast University. His research interests include space time wireless communications, random matrix theory, and information theory. He and his coauthors have been awarded the 2011 IEEE Communications Society Stephen O. Rice Prize Paper Award in the field of communication theory and the 2010 Young Author Best Paper Award by the IEEE Signal Processing Society. He serves as an Associate Editor for the IEEE Transactions on Communications, IEEE  Transactions on Wireless Communications, the IEEE Communications Letters, and IET Communications.
\end{IEEEbiography}

\begin{IEEEbiography}[{\includegraphics[width=1in,height=1.25in,clip,keepaspectratio]{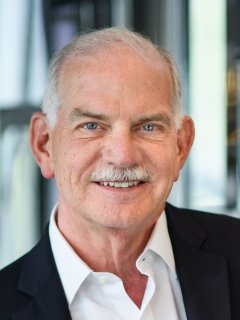}}]
	{H. Vincent Poor} (S’72, M’77, SM’82, F’87) received the Ph.D. degree in EECS from Princeton University in 1977.  From 1977 until 1990, he was on the faculty of the University of Illinois at Urbana-Champaign. Since 1990 he has been on the faculty at Princeton, where he is currently the Michael Henry Strater University Professor. During 2006 to 2016, he served as the dean of Princeton’s School of Engineering and Applied Science, and he has also held visiting appointments at several other universities, including most recently at Berkeley and Caltech. His research interests are in the areas of information theory, machine learning and network science, and their applications in wireless networks, energy systems and related fields. Among his publications in these areas is the book \textit{Machine Learning and Wireless Communications}.  (Cambridge University Press, 2022). Dr. Poor is a member of the National Academy of Engineering and the National Academy of Sciences and is a foreign member of the Royal Society and other national and international academies. He received the IEEE Alexander Graham Bell Medal in 2017.
\end{IEEEbiography}

\clearpage
\twocolumn[
\begin{center}
	{\LARGE Supplementary Material: Towards Communication-efficient Federated Learning via Sparse and Aligned Adaptive Optimization \par}
	\vspace{1em}
	
	{\large Xiumei~Deng,~Jun~Li,~Kang~Wei,~Long~Shi,~Zehui~Xiong,~Ming~Ding,~Wen~Chen,\\~Shi~Jin,~and~H.~Vincent~Poor \par}
	\vspace{1.5em}
\end{center}
]

\setcounter{equation}{32}
\section{Proof of Theorem 1}

Before presenting the main proof of \textbf{Theorem} 1, we first establish \textbf{Definition} 1 to formalise the $k$-contraction property of the Top-$k$ sparsifier, \textbf{Lemma} 1 to bound the global moment estimates, and \textbf{Lemma} 2 to characterize the gradient deviation between FedAdam-SSM and centralized Adam.\begin{definition}
	($k$-contraction property). For a positive integer $1\leq k\leq d$ and any vector $\boldsymbol{x}\in\mathbb{R}^d$, a $k$-contraction operator $\text{Comp}$: $\mathbb{R}^d\rightarrow\mathbb{R}^d$ satisfies the contraction property:\begin{small}\begin{equation}\label{k-contraction}
			\mathbb{E}\left[\left\Vert\boldsymbol{x}-\text{Comp}(\boldsymbol{x})\right\Vert^2\right]\leq \left(1-\frac{k}{d}\right)\left\Vert\boldsymbol{x}\right\Vert^2.
	\end{equation} \end{small}
\end{definition}

\begin{lemma}\label{lemma:upper bound on M&V}
	Suppose \textbf{Assumption} 2 holds. After running \textbf{Algorithm} 2 for $T$ communication rounds, for any $t\in\mathcal{T}$, and $j\in[d]$, we have
	\begin{small}\begin{alignat}{1}\label{equ1:lemma1}
			\left\vert\left[\boldsymbol{M}^t\right]_j\right\vert\leq G\left(1-(\beta_1)^{Lt}\right),
	\end{alignat}\end{small}\begin{small}\begin{alignat}{1}\label{equ2:lemma1}
			\left\vert\left[\boldsymbol{V}^t\right]_j\right\vert\leq G^2\left(1-(\beta_2)^{Lt}\right).
	\end{alignat}\end{small}
\end{lemma}
\begin{IEEEproof}We derive the upper bound on $\left\vert\left[\boldsymbol{M}^t\right]_j\right\vert$ in (\ref{equ1:lemma1}) by induction. First, the upper bound on $\left\vert\left[\boldsymbol{M}^t\right]_j\right\vert$ holds at $t=0$ since $\boldsymbol{M}^0=0$. Suppose that it holds at the $t$-th communication round. From the update rule in (4), we then obtain that
	\begin{small}\begin{alignat}{1}\label{equ3:lemma1}
			\left\vert\left[\boldsymbol{m}_n^{L,t}\right]_j\right\vert&\leq(\beta_1)^L\left\vert\left[\boldsymbol{M}^t\right]_j\right\vert+(1-\beta_1)\sum_{l=1}^L(\beta_1)^{L-l}\left\vert\left[\nabla \tilde{F}_n\left(\boldsymbol{w}_n^{l,t}\right)\right]_j\right\vert\nonumber\\&\leq G\left(1-(\beta_1)^{L(t+1)}\right),
	\end{alignat}\end{small}where the second inequality holds by \textbf{Assumption} 2 imposing a bound the stochastic gradients. From FedAvg, we derive that\begin{small}\begin{alignat}{1}
			&\left\vert\left[\boldsymbol{M}^{t+1}\right]_j\right\vert=\left\vert\left[\boldsymbol{M}^t\right]_j+\frac{\sum_{n=1}^N\tilde{D}_n\left[\text{SSM}_k\left(\boldsymbol{m}_n^{L,t}-\boldsymbol{M}^t\right)\right]_j}{\sum_{n=1}^N\tilde{D}_n}\right\vert\leq\max\Bigg\{\nonumber\\&\left.\frac{1}{\sum_{n=1}^N\tilde{D}_n}\sum_{n=1}^N\tilde{D}_n\left\vert\left[\boldsymbol{m}_n^{L,t}\right]_j\right\vert,\left[\boldsymbol{M}^t\right]_j\right\}\leq G\left(1-(\beta_1)^{L(t+1)}\right).
	\end{alignat}\end{small}Therefore, the upper bound on $\left\vert\left[\boldsymbol{M}^t\right]_j\right\vert$ holds for any $t\in\mathcal{T}$ by induction. Similarly, the bound in (\ref{equ2:lemma1}) is derived via induction. This completes the proof of \textbf{Lemma} 1.\end{IEEEproof}

From \textbf{Lemma} 1 and the update rule in (4) and (5), we have\begin{small}\begin{alignat}{1}\label{equ3:theorem1}
		0\leq \left\vert\left[\boldsymbol{m}_n^{l,t}\right]_j\right\vert\leq G,\quad0\leq \left\vert\left[\check{\boldsymbol{m}}^{l,t}\right]_j\right\vert\leq G,
\end{alignat}\end{small}\begin{small}\begin{alignat}{1}\label{equ4:theorem1}
		0\leq \left\vert\left[\boldsymbol{v}_n^{l,t}\right]_j\right\vert\leq G^2,\quad0\leq \left\vert\left[\check{\boldsymbol{v}}^{l,t}\right]_j\right\vert\leq G^2.
\end{alignat}\end{small}

\begin{lemma}\label{lemma2:theorem1}
	Suppose \textbf{Assumptions} 1 and 3 hold. For any $n\in\mathcal{N}$, $l\in\mathcal{L}$, and $t\in\mathcal{T}$, we have
	\begin{small}\begin{alignat}{1}\label{equ5:theorem1}
			\left\Vert\nabla \tilde{F}_n(\boldsymbol{w}_n^{l,t})-\nabla F(\check{\boldsymbol{w}}^{l,t})\right\Vert\leq\frac{\sigma_l}{\sqrt{\tilde{D}_n}}+\sigma_g+\rho\left\Vert\boldsymbol{w}_n^{l,t}-\check{\boldsymbol{w}}^{l,t}\right\Vert.
	\end{alignat}\end{small}
\end{lemma}
\begin{IEEEproof}
	Based on \textbf{Assumptions} 1 and 3, it holds that\begin{small}\begin{alignat}{1}
			&\left\Vert\nabla \tilde{F}_n(\boldsymbol{w}_n^{l,t})-\nabla F(\check{\boldsymbol{w}}^{l,t})\right\Vert\leq\left\Vert\nabla \tilde{F}_n(\boldsymbol{w}_n^{l,t})-\nabla F_n(\boldsymbol{w}_n^{l,t})\right\Vert\nonumber\\&+\left\Vert\nabla F_n(\boldsymbol{w}_n^{l,t})-\nabla F_n(\check{\boldsymbol{w}}^{l,t})\right\Vert+\left\Vert\nabla F_n(\check{\boldsymbol{w}}^{l,t})-\nabla F(\check{\boldsymbol{w}}^{l,t})\right\Vert\nonumber\\&\leq\frac{\sigma_l}{\sqrt{\tilde{D}_n}}+\sigma_g+\rho\left\Vert\boldsymbol{w}_n^{l,t}-\check{\boldsymbol{w}}^{l,t}\right\Vert,
	\end{alignat}\end{small}where the first inequality follows from Minkowski's inequality, and the second inequality holds by \textbf{Assumptions} 1 and 3. 
\end{IEEEproof}

Next, based on \textbf{Lemmas} \ref{lemma:upper bound on M&V} and \ref{lemma2:theorem1}, we obtain \textbf{Lemmas} \ref{lemma4} and \ref{lemma5:theorem1} that characterize the deviation between the model parameters of FedAdam-SSM and centralized Adam as follows.
\begin{lemma}\label{lemma4}
	Given \textbf{Assumptions} 1, 2, and 3, it follows that
	\begin{small}\begin{alignat}{1}\label{equ:lemma4}
			\left\Vert\boldsymbol{w}_n^{l+2,t}-\check{\boldsymbol{w}}^{l+2,t}\right\Vert\!\leq\! \psi\left\Vert\boldsymbol{w}_n^{l+1,t}-\check{\boldsymbol{w}}^{l+1,t}\right\Vert+\phi\left\Vert\boldsymbol{w}_n^{l,t}-\check{\boldsymbol{w}}^{l,t}\right\Vert+\chi.
	\end{alignat}\end{small}
\end{lemma}
\begin{IEEEproof}
	From the update rule in (4) and (5), we derive that\begin{small}\begin{alignat}{1}\label{equ6:theorem1}
			&\Bigg\Vert\frac{\boldsymbol{m}_n^{l+1,t}}{\sqrt{\boldsymbol{v}_n^{l+1,t}+\epsilon}}-\frac{\check{\boldsymbol{m}}^{l+1,t}}{\sqrt{\check{\boldsymbol{v}}^{l+1,t}+\epsilon}}\Bigg\Vert\leq\frac{\beta_1}{\sqrt{\beta_2}}\Bigg\Vert\frac{\boldsymbol{m}_n^{l,t}}{\sqrt{\boldsymbol{v}_n^{l,t}+\epsilon}}-\frac{\check{\boldsymbol{m}}^{l,t}}{\sqrt{\check{\boldsymbol{v}}^{l,t}+\epsilon}}\Bigg\Vert\nonumber\\&+\Bigg\Vert\frac{\beta_1\boldsymbol{m}_n^{l,t}}{\sqrt{\beta_2\boldsymbol{v}_n^{l,t}+\beta_2\epsilon}\sqrt{\beta_2\boldsymbol{v}_n^{l,t}+\left(1-\beta_2\right)\left(\nabla \tilde{F}_n(\boldsymbol{w}_n^{l,t})\right)^2+\epsilon}}\Bigg\Vert\nonumber\\&\times\frac{\left\Vert\beta_2\boldsymbol{v}_n^{l,t}+\left(1-\beta_2\right)\left(\nabla \tilde{F}_n(\boldsymbol{w}_n^{l,t})\right)^2+\epsilon-\left(\beta_2\boldsymbol{v}_n^{l,t}+\beta_2\epsilon\right)\right\Vert}{\left\Vert\sqrt{\beta_2\boldsymbol{v}_n^{l,t}\!+\left(1-\beta_2\right)\left(\nabla \tilde{F}_n(\boldsymbol{w}_n^{l,t})\right)^2+\epsilon}+\sqrt{\beta_2\boldsymbol{v}_n^{l,t}\!+\beta_2\epsilon}\right\Vert}\nonumber\\&+\Bigg\Vert\frac{\beta_1\check{\boldsymbol{m}}^{l,t}}{\sqrt{\beta_2\check{\boldsymbol{v}}^{l,t}+\beta_2\epsilon}\sqrt{\beta_2\check{\boldsymbol{v}}^{l,t}+\left(1-\beta_2\right)\left(\nabla F(\check{\boldsymbol{w}}^{l,t})\right)^2+\epsilon}}\Bigg\Vert\nonumber\\&\times\frac{\left\Vert\beta_2\check{\boldsymbol{v}}^{l,t}+\left(1-\beta_2\right)\left(\nabla F(\check{\boldsymbol{w}}^{l,t})\right)^2+\epsilon-\left(\beta_2\check{\boldsymbol{v}}^{l,t}+\beta_2\epsilon\right)\right\Vert}{\left\Vert\sqrt{\beta_2\check{\boldsymbol{v}}^{l,t}+\left(1-\beta_2\right)\left(\nabla F(\check{\boldsymbol{w}}^{l,t})\right)^2+\epsilon}+\sqrt{\beta_2\check{\boldsymbol{v}}^{l,t}+\beta_2\epsilon}\right\Vert}\nonumber\\&+\Bigg\Vert\frac{\left(1-\beta_1\right)\left(\nabla \tilde{F}_n(\boldsymbol{w}_n^{l,t})-\nabla F(\check{\boldsymbol{w}}^{l,t})\right)}{\sqrt{\beta_2\boldsymbol{v}_n^{l,t}+\left(1-\beta_2\right)\left(\nabla \tilde{F}_n(\boldsymbol{w}_n^{l,t})\right)^2+\epsilon}}\Bigg\Vert\nonumber\\&+\Bigg\Vert\frac{\left(1-\beta_1\right)\nabla F(\check{\boldsymbol{w}}^{l,t})}{\sqrt{\beta_2\boldsymbol{v}_n^{l,t}+\left(1-\beta_2\right)\left(\nabla \tilde{F}_n(\boldsymbol{w}_n^{l,t})\right)^2+\epsilon}}\nonumber\\&-\frac{\left(1-\beta_1\right)\nabla F(\check{\boldsymbol{w}}^{l,t})}{\sqrt{\beta_2\check{\boldsymbol{v}}^{l,t}+\left(1-\beta_2\right)\left(\nabla F(\check{\boldsymbol{w}}^{l,t})\right)^2+\epsilon}}\Bigg\Vert\nonumber\\&+\left\Vert\left(1\!-\!\beta_1\right)\nabla F(\check{\boldsymbol{w}}^{l,t})\left(\beta_2\boldsymbol{v}_n^{l,t}+\left(1\!-\!\beta_2\right)\left(\nabla \tilde{F}_n(\boldsymbol{w}_n^{l,t})\right)^2+\epsilon\right)^{-\frac{1}{2}}\left(\beta_2\check{\boldsymbol{v}}^{l,t}\right.\right.\nonumber\\&\left.\left.\!+\left(1\!-\!\beta_2\right)\left(\nabla F(\check{\boldsymbol{w}}^{l,t})\right)^2\!+\!\epsilon\right)^{-\frac{1}{2}}\right\Vert\Bigg\Vert\bigg(\beta_2\boldsymbol{v}_n^{l,t}+\left(1-\left.\!\beta_2\right)\!\left(\nabla \tilde{F}_n(\boldsymbol{w}_n^{l,t})\right)^2\!+\!\epsilon\right)^{\frac{1}{2}}\!\!\nonumber\\&+\bigg(\beta_2\check{\boldsymbol{v}}^{l,t}+\left(1-\beta_2\right)\left(\nabla F(\check{\boldsymbol{w}}^{l,t})\right)^2+\epsilon\bigg)^{\frac{1}{2}}\Bigg\Vert^{-1}\Bigg\Vert\bigg(\left(1-\beta_2\right)\!\left(\nabla \tilde{F}_n(\boldsymbol{w}_n^{l,t})\right)^2\nonumber\\&+\beta_2\boldsymbol{v}_n^{l,t}+\epsilon\bigg)-\bigg(\beta_2\check{\boldsymbol{v}}^{l,t}+\left(1-\beta_2\right)\left(\nabla F(\check{\boldsymbol{w}}^{l,t})\right)^2+\epsilon\bigg)\Bigg\Vert,
	\end{alignat}\end{small}which holds by Minkowski's and Cauchy-Schwarz inequalities. Substituting the bounds on moment estimates and gradient deviation from (\ref{equ3:theorem1}), (\ref{equ4:theorem1}) and (\ref{equ5:theorem1}) into (\ref{equ6:theorem1}), we can derive that\begin{small}\begin{alignat}{1}\label{equ9:theorem1}
			&\Bigg\Vert\frac{\boldsymbol{m}_n^{l+1,t}}{\sqrt{\boldsymbol{v}_n^{l+1,t}\!+\!\epsilon}}\!-\!\frac{\check{\boldsymbol{m}}^{l+1,t}}{\sqrt{\check{\boldsymbol{v}}^{l+1,t}\!+\!\epsilon}}\Bigg\Vert\!\leq\! \frac{\beta_1}{\sqrt{\beta_2}}\Bigg\Vert\frac{\boldsymbol{m}_n^{l,t}}{\sqrt{\boldsymbol{v}_n^{l,t}\!+\!\epsilon}}\!-\!\frac{\check{\boldsymbol{m}}^{l,t}}{\sqrt{\check{\boldsymbol{v}}^{l,t}\!+\!\epsilon}}\Bigg\Vert+\frac{\beta_1\left(1\!-\!\sqrt{\beta_2}\right)}{\epsilon\sqrt{\epsilon\beta_2}}\nonumber\\&\times\bigg(\left\Vert\boldsymbol{m}_n^{l,t}\right\Vert\left\Vert\left(\nabla \tilde{F}_n(\boldsymbol{w}_n^{l,t})\right)^2+\epsilon\right\Vert\!+\!\left\Vert\check{\boldsymbol{m}}^{l,t}\right\Vert\left\Vert\left(\nabla F(\check{\boldsymbol{w}}^{l,t})\right)^2+\epsilon\right\Vert\bigg)\!+\!\frac{1-\beta_1}{\sqrt{\epsilon}}\nonumber\\&\times\left(\frac{\sigma_l}{\sqrt{\tilde{D}_n}}+\sigma_g+\rho\left\Vert\boldsymbol{w}_n^{l,t}-\check{\boldsymbol{w}}^{l,t}\right\Vert\right)+\frac{\left(1-\beta_1\right)\left(1-\beta_2\right)}{2\epsilon\sqrt{\epsilon}}\left(\left\Vert\nabla F(\check{\boldsymbol{w}}^{l,t})\right\Vert\right.\nonumber\\&\left.+\left\Vert \nabla \tilde{F}_n(\boldsymbol{w}_n^{l,t})\right\Vert\right)\!\bigg(\!\frac{\sigma_l}{\sqrt{\tilde{D}_n}}\!+\!\sigma_g\!+\!\rho\left\Vert\boldsymbol{w}_n^{l,t}\!-\!\check{\boldsymbol{w}}^{l,t}\right\Vert\!\bigg)\!\left\Vert\nabla F(\check{\boldsymbol{w}}^{l,t})\right\Vert\!+\!\frac{\left(1\!-\!\beta_1\right)\beta_2}{2\epsilon\sqrt{\epsilon}}\nonumber\\&\times\left(\left\Vert\check{\boldsymbol{v}}^{l,t}\right\Vert\!+\!\left\Vert\boldsymbol{v}_n^{l,t}\right\Vert\right)\left\Vert\nabla F(\check{\boldsymbol{w}}^{l,t})\right\Vert\leq\! \frac{\beta_1}{\sqrt{\beta_2}}\Bigg\Vert\frac{\boldsymbol{m}_n^{l,t}}{\sqrt{\boldsymbol{v}_n^{l,t}\!+\!\epsilon}}\!-\!\frac{\check{\boldsymbol{m}}^{l,t}}{\sqrt{\check{\boldsymbol{v}}^{l,t}\!+\!\epsilon}}\Bigg\Vert\!+\!\frac{1-\beta_1}{\sqrt{\epsilon}}\nonumber\\&\times\left(1\!+\!\frac{1-\beta_2}{\epsilon}dG^2\right)\left(\frac{\sigma_l}{\sqrt{\tilde{D}_n}}+\sigma_g+\rho\left\Vert\boldsymbol{w}_n^{l,t}-\check{\boldsymbol{w}}^{l,t}\right\Vert\right)+\frac{2\beta_1\left(1-\sqrt{\beta_2}\right)}{\epsilon\sqrt{\epsilon\beta_2}}\nonumber\\&\times dG\left(G^2+\epsilon\right)+\frac{\left(1-\beta_1\right)\beta_2}{\epsilon\sqrt{\epsilon}}dG^3.
	\end{alignat}\end{small}According to the update rule in (3), it can be derived that\begin{small}\begin{alignat}{1}\label{equ7:theorem1}
			&\left\Vert\boldsymbol{w}_n^{l,t}-\check{\boldsymbol{w}}^{l,t}\right\Vert=\Bigg\Vert\boldsymbol{w}_n^{l-1,t}-\eta\frac{\boldsymbol{m}_n^{l,t}}{\sqrt{\boldsymbol{v}_n^{l,t}+\epsilon}}-\check{\boldsymbol{w}}^{l-1,t}+\eta\frac{\check{\boldsymbol{m}}^{l,t}}{\sqrt{\check{\boldsymbol{v}}^{l,t}+\epsilon}}\Bigg\Vert\nonumber\\&\leq\left\Vert\boldsymbol{w}_n^{l-1,t}-\check{\boldsymbol{w}}^{l-1,t}\right\Vert+\eta\Bigg\Vert\frac{\boldsymbol{m}_n^{l,t}}{\sqrt{\boldsymbol{v}_n^{l,t}+\epsilon}}-\frac{\check{\boldsymbol{m}}^{l,t}}{\sqrt{\check{\boldsymbol{v}}^{l,t}+\epsilon}}\Bigg\Vert,
	\end{alignat}\end{small}\begin{small}\begin{alignat}{1}\label{equ8:theorem1}
			&\Bigg\Vert\frac{\boldsymbol{m}_n^{l,t}}{\sqrt{\boldsymbol{v}_n^{l,t}+\epsilon}}-\frac{\check{\boldsymbol{m}}^{l,t}}{\sqrt{\check{\boldsymbol{v}}^{l,t}+\epsilon}}\Bigg\Vert=\frac{1}{\eta}\left\Vert\boldsymbol{w}_n^{l,t}-\boldsymbol{w}_n^{l-1,t}-\check{\boldsymbol{w}}^{l,t}+\check{\boldsymbol{w}}^{l-1,t}\right\Vert\nonumber\\&\leq\frac{1}{\eta}\left\Vert\boldsymbol{w}_n^{l-1,t}-\check{\boldsymbol{w}}^{l-1,t}\right\Vert+\frac{1}{\eta}\left\Vert\boldsymbol{w}_n^{l,t}-\check{\boldsymbol{w}}^{l,t}\right\Vert.
	\end{alignat}\end{small}Substituting (\ref{equ7:theorem1}) and (\ref{equ8:theorem1}) into (\ref{equ9:theorem1}), we then obtain (\ref{equ:lemma4}). This concludes the proof of \textbf{Lemma} \ref{lemma4}.
\end{IEEEproof}

\begin{lemma}\label{lemma5:theorem1}
	Given \textbf{Assumptions} 1, 2 and 3, it follows that 
	\begin{small}\begin{alignat}{1}\label{equ1:lemma5}
			&\left\Vert\boldsymbol{w}_n^{1,t}-\check{\boldsymbol{w}}^{1,t}\right\Vert\leq\left(1+\frac{\eta\rho}{\sqrt{\epsilon}}\left(1-\beta_1\right)+\frac{dG^2\eta\rho}{\epsilon\sqrt{\epsilon}}\left(1-\beta_2\right) \right)\left\Vert\boldsymbol{W}^{t}-\breve{\boldsymbol{W}^t}\right\Vert\nonumber\\&+\frac{\eta\beta_1}{\sqrt{\epsilon}}\left\Vert\boldsymbol{M}^{t}-\breve{\boldsymbol{M}^t}\right\Vert+\frac{\sqrt{d}G\eta\beta_2}{2\epsilon\sqrt{\epsilon}}\left\Vert\boldsymbol{V}^{t}-\breve{\boldsymbol{V}^t}\right\Vert+\left(\frac{\eta}{\sqrt{\epsilon}}\left(1-\beta_1\right)+\frac{dG^2\eta}{\epsilon\sqrt{\epsilon}}\right.\nonumber\\&\left.\times\left(1-\beta_2\right) \right)\bigg(\frac{\sigma_l}{\sqrt{\tilde{D}_n}}+\sigma_g\bigg).
	\end{alignat}\end{small}
\end{lemma}\begin{IEEEproof}
	From (\ref{equ7:theorem1}), it can be derived that\begin{small}\begin{alignat}{1}\label{equ2:lemma5}
			&\frac{1}{\eta}\left(\left\Vert\boldsymbol{w}_n^{1,t}-\check{\boldsymbol{w}}^{1,t}\right\Vert\!-\!\left\Vert\boldsymbol{W}^{t}-\breve{\boldsymbol{W}^t}\right\Vert\right)\!\leq\!\Bigg\Vert\frac{\beta_1\boldsymbol{M}^{t}+\left(1\!-\!\beta_1\right)\nabla \tilde{F}_n(\boldsymbol{W}^{t})}{\sqrt{\beta_2\boldsymbol{V}^{t}\!+\!\left(1\!-\!\beta_2\right)\left(\nabla \tilde{F}_n(\boldsymbol{W}^{t})\right)^2\!+\!\epsilon}}\nonumber\\&-\frac{\beta_1\breve{\boldsymbol{M}^t}+\left(1-\beta_1\right)\nabla F(\breve{\boldsymbol{W}^t})}{\sqrt{\beta_2\breve{\boldsymbol{V}^t}+\left(1-\beta_2\right)\left(\nabla F(\breve{\boldsymbol{W}^t})\right)^2+\epsilon}}\Bigg\Vert\leq\frac{\beta_1}{\sqrt{\epsilon}}\!\left\Vert\boldsymbol{M}^{t}\!-\!\breve{\boldsymbol{M}^t}\right\Vert\!+\!\frac{1}{2\epsilon\sqrt{\epsilon}}\Big(\beta_1\nonumber\\&\times\left\Vert\breve{\boldsymbol{M}^t}\right\Vert\!+\!\left(1-\beta_1\right)\!\left\Vert\nabla F\left(\breve{\boldsymbol{W}^t}\right)\right\Vert\Big)\bigg(\beta_2\left\Vert\breve{\boldsymbol{V}^t}\!-\!\boldsymbol{V}^{t}\right\Vert\!+\!\left(1-\beta_2\right)\Big\Vert\left(\nabla F(\breve{\boldsymbol{W}^t})\right)^2\nonumber\\&-\left(\nabla \tilde{F}_n(\boldsymbol{W}^{t})\right)^2\Big\Vert\bigg)+\frac{1-\beta_1}{\sqrt{\epsilon}}\Big(\left\Vert \nabla \tilde{F}_n\left(\boldsymbol{W}^{t}\right)-\nabla F_n\left(\boldsymbol{W}^{t}\right)\right\Vert+\left\Vert\nabla F_n\left(\boldsymbol{W}^{t}\right)\right.\nonumber\\&\left.-\nabla F\left(\boldsymbol{W}^{t}\right)\right\Vert+\left\Vert\nabla F\left(\boldsymbol{W}^{t}\right)-\nabla F\left(\breve{\boldsymbol{W}^t}\right)\right\Vert\Big),
	\end{alignat}\end{small}where the last inequality follows from the bounds on moment estimates in (\ref{equ3:theorem1}) and (\ref{equ4:theorem1}). Substituting (\ref{equ5:theorem1}) into (\ref{equ2:lemma5}) yields\begin{small}\begin{alignat}{1}
			&\frac{1}{\eta}\left(\left\Vert\boldsymbol{w}_n^{1,t}-\check{\boldsymbol{w}}^{1,t}\right\Vert-\left\Vert\boldsymbol{W}^{t}-\hat{\boldsymbol{W}}^{t}\right\Vert\right)\leq\frac{\beta_1}{\sqrt{\epsilon}}\left\Vert\boldsymbol{M}^{t}-\hat{\boldsymbol{M}}^{t}\right\Vert+\frac{\beta_2\sqrt{d}G}{2\epsilon\sqrt{\epsilon}}\left\Vert\boldsymbol{V}^{t}\right.\nonumber\\&\left.-\hat{\boldsymbol{V}}^{t}\right\Vert+\left(\frac{\left(1-\beta_2\right)dG^2}{\epsilon\sqrt{\epsilon}}+\frac{1-\beta_1}{\sqrt{\epsilon}}\right)\left(\rho\left\Vert\boldsymbol{W}^{t}\!-\!\hat{\boldsymbol{W}}^{t}\right\Vert\!+\!\frac{\sigma_l}{\sqrt{\tilde{D}_n}}\!+\!\sigma_g\right).\!\!
	\end{alignat}\end{small}This concludes the proof of \textbf{Lemma} \ref{lemma5:theorem1}.
\end{IEEEproof}

Define the recurrence relations for all $l\in\mathcal{L}$ as\begin{small}\begin{alignat}{1}\label{recurrence relation}
		x_{l+2}=\psi x_{l+1}+\phi x_{l}+\chi,\end{alignat}\end{small}where the constants $\psi$, $\phi$, and $\chi$ are given in (20), (19), and (21), respectively. Let $x_0=\left\Vert\boldsymbol{w}_n^{0,t}-\check{\boldsymbol{w}}^{0,t}\right\Vert=\left\Vert\boldsymbol{W}^{t}-\breve{\boldsymbol{W}^t}\right\Vert$, and $x_1$ be the upper bound on $\left\Vert\boldsymbol{w}_n^{1,t}-\check{\boldsymbol{w}}^{1,t}\right\Vert$ in (\ref{equ1:lemma5}). Solving the homogeneous linear recurrence relations in (\ref{recurrence relation}) yields\begin{small}\begin{alignat}{1}
		x_l=\Gamma\left\Vert\boldsymbol{W}^{t}-\breve{\boldsymbol{W}^t}\right\Vert+\Lambda\left\Vert\boldsymbol{M}^{t}-\breve{\boldsymbol{M}^t}\right\Vert+\Theta\left\Vert\boldsymbol{V}^{t}-\breve{\boldsymbol{V}^t}\right\Vert+\Phi,\end{alignat}\end{small}where the constants $\Gamma$, $\Lambda$, $\Theta$, $\Phi$ are given in (15), (16), (17) and (18), respectively. Note that $\psi$, $\phi$, and $\chi>0$, and recall that $\left\Vert\boldsymbol{w}_n^{l+2,t}-\check{\boldsymbol{w}}^{l+2,t}\right\Vert\leq \psi\left\Vert\boldsymbol{w}_n^{l+1,t}-\check{\boldsymbol{w}}^{l+1,t}\right\Vert+\phi\left\Vert\boldsymbol{w}_n^{l,t}-\check{\boldsymbol{w}}^{l,t}\right\Vert+\chi$ in (\ref{equ:lemma4}). It is clear that for all $l\in\mathcal{L}$, $\left\Vert\boldsymbol{w}_n^{l,t}-\check{\boldsymbol{w}}^{l,t}\right\Vert\leq x_l$, i.e.,\begin{small}\begin{alignat}{1}
		\left\Vert\boldsymbol{w}_n^{l,t}-\check{\boldsymbol{w}}^{l,t}\right\Vert\leq\Gamma\left\Vert\boldsymbol{W}^{t}\!-\!\breve{\boldsymbol{W}^t}\right\Vert\!+\!\Lambda\left\Vert\boldsymbol{M}^{t}\!-\!\breve{\boldsymbol{M}^t}\right\Vert\!+\!\Theta\left\Vert\boldsymbol{V}^{t}\!-\!\breve{\boldsymbol{V}^t}\right\Vert+\Phi.
\end{alignat}\end{small}

Next, leveraging the update rules of ${\boldsymbol{M}^t}$, ${\boldsymbol{V}^t}$, and ${\boldsymbol{W}^t}$ along with $\breve{\boldsymbol{M}^t}$, $\breve{\boldsymbol{V}^t}$, and $\breve{\boldsymbol{W}^t}$, it can be derived that\begin{small}\begin{alignat}{1}
		&\Gamma\left\Vert\boldsymbol{W}^{t}-\breve{\boldsymbol{W}^t}\right\Vert+\Lambda\left\Vert\boldsymbol{M}^{t}-\breve{\boldsymbol{M}^t}\right\Vert+\Theta\left\Vert\boldsymbol{V}^{t}-\breve{\boldsymbol{V}^t}\right\Vert\nonumber\\&\leq\sum_{n=1}^N\frac{\tilde{D}_n}{\sum_{n=1}^N\tilde{D}_n}\left(\Gamma\left\Vert\Delta\boldsymbol{W}_n^{t-1}\odot\left(1-\mathbbm{1}_{\text{SSM}_n^{t-1}}\right)\right\Vert\right.+\Lambda\left\Vert\Delta\boldsymbol{M}_n^{t-1}\right.\nonumber\\&\left.\left.\odot\left(1-\mathbbm{1}_{\text{SSM}_n^{t-1}}\right)\right\Vert+\Theta\left\Vert\Delta\boldsymbol{V}_n^{t-1}\odot\left(1-\mathbbm{1}_{\text{SSM}_n^{t-1}}\right)\right\Vert\right),
\end{alignat}\end{small}where the inequality holds by Minkowski's inequality. This completes the proof of \textbf{Theorem} 1.

\section{Proof of Proposition 1}
Before we show the main proof of \textbf{Proposition} 1, we first give \textbf{Lemma} \ref{lemma6:proposition_1} below.
\begin{lemma}\label{lemma6:proposition_1}
	For all $l\in\mathcal{L}$, it follows that 
	\begin{small}\begin{alignat}{1}\label{equ1:lemma6}
			\frac{\left(\frac{\psi+\sqrt{\psi^2+4\phi}}{2}\right)^{l+1}-\left(\frac{\psi-\sqrt{\psi^2+4\phi}}{2}\right)^{l+1}}{\left(\frac{\psi+\sqrt{\psi^2+4\phi}}{2}\right)^{l}-\left(\frac{\psi-\sqrt{\psi^2+4\phi}}{2}\right)^{l}}\ge\psi.
	\end{alignat}\end{small}
\end{lemma}\begin{IEEEproof}
	From (19) and (20), we can derive that $\phi>0$ and $\psi>0$. Clearly, it follows that $-1<\frac{\psi-\sqrt{\psi^2+4\phi}}{\psi+\sqrt{\psi^2+4\phi}}<0$. Then, for all $l\in\mathcal{L}$, we have\begin{small}\begin{alignat}{1}\label{equ2:lemma6}
			\left(\frac{\psi-\sqrt{\psi^2+4\phi}}{\psi+\sqrt{\psi^2+4\phi}}\right)^l\ge\frac{\psi-\sqrt{\psi^2+4\phi}}{\psi+\sqrt{\psi^2+4\phi}}
	\end{alignat}\end{small}From (\ref{equ2:lemma6}), it can be derived that\begin{small}\begin{alignat}{1}
			&\frac{\left(\frac{\psi+\sqrt{\psi^2+4\phi}}{2}\right)^{l+1}-\left(\frac{\psi-\sqrt{\psi^2+4\phi}}{2}\right)^{l+1}}{\left(\frac{\psi+\sqrt{\psi^2+4\phi}}{2}\right)^{l}-\left(\frac{\psi-\sqrt{\psi^2+4\phi}}{2}\right)^{l}}=\frac{\sqrt{\psi^2+4\phi}}{1-\left(\frac{\psi-\sqrt{\psi^2+4\phi}}{\psi+\sqrt{\psi^2+4\phi}}\right)^l}\nonumber\\&+\frac{\psi-\sqrt{\psi^2+4\phi}}{2}\ge\frac{\psi-\sqrt{\psi^2+4\phi}}{2}+\frac{\sqrt{\psi^2+4\phi}}{1-\frac{\psi-\sqrt{\psi^2+4\phi}}{\psi+\sqrt{\psi^2+4\phi}}}=\psi.
	\end{alignat}\end{small}This concludes the proof of \textbf{Lemma} \ref{lemma6:proposition_1}.\end{IEEEproof}Given $\beta_2<1-\frac{1}{1+2G\rho\sqrt{d}}$, it can be shown that\begin{small}\begin{alignat}{1}\label{equ3:p1}
		&\phi-\frac{dG^2\eta\rho}{\epsilon\sqrt{\epsilon}}\beta_1\left(1-\beta_2\right)+\frac{\sqrt{d}G\eta\beta_2}{2\epsilon\sqrt{\epsilon}}-\psi=-\frac{dG^2\eta\rho}{\epsilon\sqrt{\epsilon}}\left(1-\beta_2\right)\nonumber\\&+\frac{\sqrt{d}G\eta\beta_2}{2\epsilon\sqrt{\epsilon}} - 1-\frac{\eta\rho}{\sqrt{\epsilon}}\left(1-\beta_1\right)\leq - 1-\frac{\eta\rho}{\sqrt{\epsilon}}\left(1-\beta_1\right)< 0.
\end{alignat}\end{small}Substituting (\ref{equ1:lemma6}) into (\ref{equ3:p1}), we have\begin{small}\begin{alignat}{1}\label{equ4:p1}
		&\phi-\frac{dG^2\eta\rho}{\epsilon\sqrt{\epsilon}}\beta_1\left(1-\beta_2\right)+\frac{\sqrt{d}G\eta\beta_2}{2\epsilon\sqrt{\epsilon}}\nonumber\\&<\frac{\left(\frac{\psi+\sqrt{\psi^2+4\phi}}{2}\right)^{l+1}-\left(\frac{\psi-\sqrt{\psi^2+4\phi}}{2}\right)^{l+1}}{\left(\frac{\psi+\sqrt{\psi^2+4\phi}}{2}\right)^{l}-\left(\frac{\psi-\sqrt{\psi^2+4\phi}}{2}\right)^{l}}.
\end{alignat}\end{small}Note that $\left(\frac{\psi+\sqrt{\psi^2+4\phi}}{2}\right)^{l}-\left(\frac{\psi-\sqrt{\psi^2+4\phi}}{2}\right)^{l}>0$ for all $l\in\mathcal{L}$. Rearranging (\ref{equ4:p1}), it follows that\begin{small}\begin{alignat}{1}\label{equ5:p1}
		&\frac{\sqrt{d}G\eta\beta_2}{2\epsilon\sqrt{\epsilon}}\bigg(\bigg(\frac{\psi+\sqrt{\psi^2+4\phi}}{2}\bigg)^{l}-\bigg(\frac{\psi-\sqrt{\psi^2+4\phi}}{2}\bigg)^{l}\bigg)\nonumber\\&<\bigg(\frac{\psi+\sqrt{\psi^2+4\phi}}{2}\bigg)^{l}\Bigg(\frac{\psi+\sqrt{\psi^2+4\phi}}{2}-\phi+\frac{dG^2\eta\rho}{\epsilon\sqrt{\epsilon}}\beta_1\left(1-\beta_2\right)\Bigg)\nonumber\\&+\bigg(\!\frac{\sqrt{\psi^2\!+\!4\phi}-\psi}{2}+\phi-\frac{dG^2\eta\rho}{\epsilon\sqrt{\epsilon}}\beta_1\left(1-\beta_2\right)\!\bigg)\!\bigg(\frac{\psi-\sqrt{\psi^2+4\phi}}{2}\bigg)^{l}\!.\!\!\!\!
\end{alignat}\end{small}Therefore, we have $\Gamma>\Theta$ when $\beta_2<1-\frac{1}{1+2G\rho\sqrt{d}}$. Given $\epsilon>0$ is a small number, it can be derived that\begin{small}\begin{alignat}{1}\label{equ6:p1}
		&\Theta-\Lambda=\frac{\bigg(\frac{\psi+\sqrt{\psi^2+4\phi}}{2}\bigg)^l\!\!-\!\bigg(\frac{\psi-\sqrt{\psi^2+4\phi}}{2}\bigg)^l}{\sqrt{\epsilon}\sqrt{\psi^2+4\phi}}\bigg(\beta_1\!-\!\frac{\sqrt{d}G\beta_2}{2\epsilon}\bigg)\eta>0.\end{alignat}\end{small}This concludes the proof of \textbf{Proposition} 1.

\section{Proof of Theorem 2}
For ease of exposition, we rewrite the sparse updates of local moment estimates and model parameters in FedAdam-SSM as $\Delta\hat{\boldsymbol{M}}_n^t =\text{SSM}_k\left(\Delta\boldsymbol{M}_n^t\right)$, $\Delta\hat{\boldsymbol{V}}_n^t =\text{SSM}_k\left(\Delta\boldsymbol{V}_n^t\right)$, and $\Delta\hat{\boldsymbol{W}}_n^t =\text{SSM}_k\left(\Delta\boldsymbol{W}_n^t\right)$. First, the $\rho$-Lipschitz continuity assumption gives us  
\begin{small}\begin{alignat}{1}
		&F\left(\boldsymbol{W}^{t+1}\right)-F\left(\boldsymbol{W}^t\right)\leq \frac{\rho}{2}\Bigg\Vert\sum_{n=1}^N\frac{\tilde{D}_n}{\sum_{n=1}^N\tilde{D}_n}\text{SSM}_k\bigg(-\eta\sum_{l=1}^L\frac{\boldsymbol{m}_n^{l,t}}{\sqrt{\boldsymbol{v}_n^{l,t}+\epsilon}}\bigg)\Bigg\Vert^2\nonumber\\&+\Bigg\langle\nabla F(\boldsymbol{W}^t),\sum_{n=1}^N\frac{\tilde{D}_n}{\sum_{n=1}^N\tilde{D}_n}\text{SSM}_k\Bigg(-\eta\sum_{l=1}^L\frac{\boldsymbol{m}_n^{l,t}}{\sqrt{\boldsymbol{v}_n^{l,t}+\epsilon}}\Bigg)\Bigg\rangle\nonumber\\&=-\frac{\eta}{2}\Bigg(\gamma\bigg\Vert\sum_{n=1}^N\frac{\tilde{D}_n}{\sum_{n=1}^N\tilde{D}_n}\text{SSM}_k\bigg(\sum_{l=1}^L\frac{\boldsymbol{m}_n^{l,t}}{\sqrt{\boldsymbol{v}_n^{l,t}+\epsilon}}\bigg)\bigg\Vert^2+\frac{1}{\gamma}\left\Vert\nabla F(\boldsymbol{W}^t)\right\Vert^2\nonumber\\&-\gamma\bigg\Vert-\frac{1}{\gamma}\nabla F(\boldsymbol{W}^t)+\sum_{n=1}^N\frac{\tilde{D}_n}{\sum_{n=1}^N\tilde{D}_n}\text{SSM}_k\bigg(\sum_{l=1}^L\frac{\boldsymbol{m}_n^{l,t}}{\sqrt{\boldsymbol{v}_n^{l,t}+\epsilon}}\bigg)\bigg\Vert^2\Bigg)\nonumber\\&+\frac{\eta^2\rho}{2}\sum_{n=1}^N\frac{\tilde{D}_n}{\sum_{n=1}^N\tilde{D}_n}\Bigg\Vert \text{SSM}_k\Bigg(\sum_{l=1}^L\frac{\boldsymbol{m}_n^{l,t}}{\sqrt{\boldsymbol{v}_n^{l,t}+\epsilon}}\Bigg)\Bigg\Vert^2.\end{alignat}\end{small}Given $\gamma=1$, it can be derived that\begin{small}\begin{alignat}{1}&F\left(\boldsymbol{W}^{t+1}\right)\!-\!F\left(\boldsymbol{W}^t\right)\leq\left(\eta^2\rho\!-\!\eta\right)\!\sum_{n=1}^N\!\frac{\tilde{D}_n}{\sum_{n=1}^N\tilde{D}_n}\Bigg(\Bigg\Vert\text{SSM}_k\Bigg(\!\sum_{l=1}^L\!\frac{\boldsymbol{m}_n^{l,t}}{\sqrt{\boldsymbol{v}_n^{l,t}+\epsilon}}\Bigg)\nonumber\\&-\sum_{l=1}^L\frac{\boldsymbol{m}_n^{l,t}}{\sqrt{\boldsymbol{v}_n^{l,t}+\epsilon}}\Bigg\Vert^2+\Bigg\Vert\sum_{l=1}^L\frac{\boldsymbol{m}_n^{l,t}}{\sqrt{\boldsymbol{v}_n^{l,t}+\epsilon}}\Bigg\Vert^2\Bigg)+3\eta\sum_{n=1}^N\!\frac{\tilde{D}_n}{\sum_{n=1}^N\!\tilde{D}_n}\Bigg(\Bigg\Vert\text{SSM}_k\Bigg(\nonumber\\&\sum_{l=1}^L\!\frac{\boldsymbol{m}_n^{l,t}}{\sqrt{\boldsymbol{v}_n^{l,t}\!+\!\epsilon}}\Bigg)\!-\!\sum_{l=1}^L\!\frac{\boldsymbol{m}_n^{l,t}}{\sqrt{\boldsymbol{v}_n^{l,t}\!+\!\epsilon}}\Bigg\Vert^2\!+\!\Bigg\Vert\sum_{l=1}^L\frac{\boldsymbol{m}_n^{l,t}}{\sqrt{\boldsymbol{v}_n^{l,t}\!+\!\epsilon}}\!-\!\sum_{l=1}^L\frac{\boldsymbol{m}_n^{l,t}}{\sqrt{(\beta_2)^l\boldsymbol{V}^t+\epsilon}}\Bigg\Vert^2\nonumber\\&+\left\Vert\sum_{l=1}^L\frac{\boldsymbol{m}_n^{l,t}}{\sqrt{(\beta_2)^l\boldsymbol{V}^t+\epsilon}}-\sum_{l=1}^L\frac{\nabla \tilde{F}_n(\boldsymbol{w}_n^{l,t})}{\sqrt{(\beta_2)^l\boldsymbol{V}^t+\epsilon}}\right\Vert^2+\left\Vert\sum_{l=1}^L\frac{\nabla \tilde{F}_n(\boldsymbol{w}_n^{l,t})}{\sqrt{(\beta_2)^l\boldsymbol{V}^t+\epsilon}}\right.\nonumber\\&\left.-\sum_{l=1}^L\frac{\nabla \tilde{F}_n(\boldsymbol{w}_n^{l,t})}{L}\right\Vert^2+\left\Vert\sum_{l=1}^L\frac{\nabla \tilde{F}_n(\boldsymbol{w}_n^{l,t})}{L}-\sum_{l=1}^L\frac{\nabla F(\boldsymbol{w}_n^{l,t})}{L}\right\Vert^2\nonumber\\&\left.+\left\Vert\sum_{l=1}^L\frac{\nabla F(\boldsymbol{w}_n^{l,t})}{L}-\nabla F(\boldsymbol{W}^t)\right\Vert^2\right)-\frac{\eta}{2}\left\Vert\nabla F(\boldsymbol{W}^t)\right\Vert^2,\!
\end{alignat}\end{small}which holds by Jensen's inequality. The $k$-contraction property in (6) yields that\begin{small}\begin{alignat}{1}\label{equ0:theorem2}&F\left(\boldsymbol{W}^{t+1}\right)-F\left(\boldsymbol{W}^t\right)\leq\left(\left(\eta^2\rho+2\eta\right)\left(1-\frac{k}{d}\right)+\eta^2\rho-\eta\right)\sum_{n=1}^N\frac{\tilde{D}_n}{\sum_{n=1}^N\!\tilde{D}_n}\nonumber\\&\times\Bigg\Vert\!\sum_{l=1}^L\!\frac{\boldsymbol{m}_n^{l,t}}{\sqrt{\boldsymbol{v}_n^{l,t}\!+\!\epsilon}}\Bigg\Vert^2\!\!+\!3\eta\!\sum_{n=1}^N\!\frac{\tilde{D}_n}{\sum_{n=1}^N\!\tilde{D}_n}\Bigg(\Bigg\Vert\!\sum_{l=1}^L\!\!\frac{\boldsymbol{m}_n^{l,t}}{\sqrt{\boldsymbol{v}_n^{l,t}\!+\!\epsilon}}\!-\!\!\sum_{l=1}^L\!\!\frac{\boldsymbol{m}_n^{l,t}}{\sqrt{(\beta_2)^l\boldsymbol{V}^t\!+\!\epsilon}}\Bigg\Vert^2\nonumber\\&+\left\Vert\sum_{l=1}^L\!\frac{\boldsymbol{m}_n^{l,t}}{\sqrt{(\beta_2)^l\boldsymbol{V}^t\!+\!\epsilon}}\!-\!\sum_{l=1}^L\!\frac{\nabla \tilde{F}_n(\boldsymbol{w}_n^{l,t})}{\sqrt{(\beta_2)^l\boldsymbol{V}^t+\epsilon}}\right\Vert^2\!+\!\Bigg\Vert\sum_{l=1}^L\frac{\nabla \tilde{F}_n(\boldsymbol{w}_n^{l,t})}{\sqrt{(\beta_2)^l\boldsymbol{V}^t+\epsilon}}-\!\sum_{l=1}^L\nonumber\\&\frac{\nabla \tilde{F}_n(\boldsymbol{w}_n^{l,t})}{L}\Bigg\Vert^2+\left\Vert\sum_{l=1}^L\frac{\nabla \tilde{F}_n(\boldsymbol{w}_n^{l,t})}{L}-\sum_{l=1}^L\frac{\nabla F(\boldsymbol{w}_n^{l,t})}{L}\right\Vert^2+\Bigg\Vert\sum_{l=1}^L\frac{\nabla F(\boldsymbol{w}_n^{l,t})}{L}\nonumber\\&-\nabla F(\boldsymbol{W}^t)\Big\Vert^2\Big)-\frac{\eta}{2}\left\Vert\nabla F(\boldsymbol{W}^t)\right\Vert^2.
\end{alignat}\end{small}From the update rule of $\boldsymbol{m}_n^{l,t}$ in (4), we have\begin{small}\begin{alignat}{1}\label{sum_m}
		\sum_{l=1}^L\boldsymbol{m}_n^{l,t}\!=\!\frac{\beta_1\big(1\!-\!(\beta_1)^L\big)}{1\!-\!\beta_1}\boldsymbol{M}^t\!+\!\sum_{l=1}^L\!\big(1\!-\!(\beta_1)^{L-l+1}\big)\nabla \tilde{F}_n(\boldsymbol{w}_n^{l-1,t}).
\end{alignat}\end{small}From (\ref{sum_m}), it can be derived that\begin{small}\begin{alignat}{1}\label{equ1:theorem2}
		&\bigg\Vert \sum_{l=1}^L\frac{\boldsymbol{m}_n^{l,t}}{\sqrt{\boldsymbol{v}_n^{l,t}+\epsilon}}\bigg\Vert^2\leq\frac{1}{\epsilon}\bigg\Vert\sum_{l=1}^L\boldsymbol{m}_n^{l,t}\bigg\Vert^2\leq\frac{L\beta_1\left(1-(\beta_1)^L\right)G^2d}{\epsilon(1-\beta_1)}+\frac{LG^2d}{\epsilon}\nonumber\\&\times\bigg(L-\frac{\beta_1\big(1-(\beta_1)^L\big)}{1-\beta_1}\bigg)=\frac{L^2G^2d}{\epsilon},\!\!\!\!
\end{alignat}\end{small}where the last inequality holds by \textbf{Lemma} \ref{lemma:upper bound on M&V}. Based on the update rule of $\boldsymbol{v}_n^{l,t}$ in (5), we can derive that\begin{small}\begin{alignat}{1}\label{sum_v}
		\boldsymbol{v}_n^{l,t}=(\beta_2)^l\boldsymbol{V}^t\!+\!(1-\beta_2)\!\sum_{\tau=1}^l(\beta_2)^{l-\tau}\!\left(\nabla \tilde{F}_n(\boldsymbol{w}_n^{\tau-1,t})\right)^2\!.\!\!\!
\end{alignat}\end{small}From (\ref{sum_v}), it can be derived that\begin{small}\begin{alignat}{1}
		&\Bigg\Vert\sum_{l=1}^L\Bigg(\frac{\boldsymbol{m}_n^{l,t}}{\sqrt{\boldsymbol{v}_n^{l,t}+\epsilon}}-\frac{\boldsymbol{m}_n^{l,t}}{\sqrt{(\beta_2)^l\boldsymbol{V}^t+\epsilon}}\Bigg)\Bigg\Vert^2\nonumber\\&=\Bigg\Vert\sum_{l=1}^L\frac{\boldsymbol{m}_n^{l,t}\left((\beta_2)^l\boldsymbol{V}^t-\boldsymbol{v}_n^{l,t}\right)}{\sqrt{\boldsymbol{v}_n^{l,t}+\epsilon}\sqrt{(\beta_2)^l\boldsymbol{V}^t+\epsilon}\Big(\sqrt{\boldsymbol{v}_n^{l,t}+\epsilon}+\sqrt{(\beta_2)^l\boldsymbol{V}^t+\epsilon}\Big)}\Bigg\Vert^2\nonumber\\&\leq\frac{L}{4\epsilon^3}\sum_{l=1}^L\left\Vert\boldsymbol{m}_n^{l,t}\left((1-\beta_2)\!\sum_{\tau=1}^l(\beta_2)^{l-\tau}\left(\nabla \tilde{F}_n(\boldsymbol{w}_n^{\tau-1,t})\right)^2\right)\right\Vert^2,
\end{alignat}\end{small}where the last inequality holds by Jensen's inequality. The Cauchy-Schwarz inequality along with \textbf{Lemma} \ref{lemma:upper bound on M&V} yields\begin{small}\begin{alignat}{1}\label{equ2:theorem2}
		&\Bigg\Vert\sum_{l=1}^L\!\Bigg(\!\frac{\boldsymbol{m}_n^{l,t}}{\sqrt{\boldsymbol{v}_n^{l,t}\!+\!\epsilon}}\!-\!\frac{\boldsymbol{m}_n^{l,t}}{\sqrt{(\beta_2)^l\boldsymbol{V}^t\!+\!\epsilon}}\!\Bigg)\Bigg\Vert^2\!\!\!\leq\!\frac{LG^6d^2}{4\epsilon^3}\Bigg(L\!-\!\frac{\beta_2\big(1\!-\!(\beta_2)^L\big)}{1\!-\!\beta_2}\Bigg).\!\!
\end{alignat}\end{small}From (\ref{sum_v}), it can be derived as
\begin{small}\begin{alignat}{1}\label{equ3:theorem2}
		&\left\Vert\sum_{l=1}^L\left(\frac{\boldsymbol{m}_n^{l,t}}{\sqrt{(\beta_2)^l\boldsymbol{V}^t\!+\!\epsilon}}\!-\!\frac{\nabla \tilde{F}_n(\boldsymbol{w}_n^{l-1,t})}{\sqrt{(\beta_2)^l\boldsymbol{V}^t\!+\!\epsilon}}\right)\right\Vert^2\!\leq\!\frac{1}{\epsilon}\Bigg\Vert\sum_{l=1}^L(\beta_1)^l\boldsymbol{M}^t\!+\!\sum_{l=1}^L(1\!-\!\beta_1)\nonumber\\&\times\sum_{\tau=1}^{l-1}(\beta_1)^{l-\tau}\nabla \tilde{F}_n(\boldsymbol{w}_n^{\tau-1,t})-\sum_{l=1}^L\nabla \tilde{F}_n(\boldsymbol{w}_n^{l-1,t})\Bigg\Vert^2\leq\frac{2\beta_1}{\epsilon\left(1-\beta_1\right)}\sum_{l=1}^L\nonumber\\&(\beta_1)^l\left(\left\Vert\boldsymbol{M}^t\right\Vert^2+\left\Vert\nabla \tilde{F}_n(\boldsymbol{w}_n^{L-l,t})\right\Vert^2\right)\leq\frac{4\beta_1\left(1-(\beta_1)^L\right)G^2d}{\epsilon\left(1-\beta_1\right)^2},
\end{alignat}\end{small}where the second inequality holds by Cauchy-Schwarz inequality, and the last inequality holds by \textbf{Lemma} \ref{lemma:upper bound on M&V}. It follows from the Cauchy–Schwarz inequality that\begin{small}\begin{alignat}{1}\label{equ4:theorem2}
		&\left\Vert\sum_{l=1}^L\left(\frac{\nabla \tilde{F}_n(\boldsymbol{w}_n^{l-1,t})}{\sqrt{(\beta_2)^l\boldsymbol{V}^t+\epsilon}}-\frac{\nabla \tilde{F}_n(\boldsymbol{w}_n^{l-1,t})}{L}\right)\right\Vert^2\leq\frac{L^2G^2d}{\epsilon}+G^2d.
\end{alignat}\end{small}It follows from the Cauchy-Schwarz inequality and \textbf{Assumption} 3 and 2 that\begin{small}\begin{alignat}{1}\label{equ5:theorem2}
		\left\Vert\sum_{l=1}^L\left(\frac{\nabla \tilde{F}_n(\boldsymbol{w}_n^{l-1,t})}{L}-\frac{\nabla F(\boldsymbol{w}_n^{l-1,t})}{L}\right)\right\Vert^2\leq\left(\frac{\sigma_l}{\sqrt{\tilde{D}_n}}+\sigma_g\right)^2.
\end{alignat}\end{small}Next, we can derive that\begin{small}\begin{alignat}{1}\label{equ6:theorem2}
		&\left\Vert\sum_{l=1}^L\left(\frac{\nabla F(\boldsymbol{w}_n^{l-1,t})}{L}-\frac{\nabla F(\boldsymbol{W}^t)}{L}\right)\right\Vert^2\leq\frac{\rho^2}{L}\sum_{l=1}^L\left\Vert\boldsymbol{w}_n^{l-1,t}-\boldsymbol{W}^t\right\Vert^2\nonumber\\&\leq\frac{\rho^2}{L}\sum_{l=1}^{L-1}\bigg\Vert\sum_{\tau=1}^{l}\frac{\boldsymbol{m}_n^{\tau,t}}{\sqrt{\boldsymbol{v}_n^{\tau,t}+\epsilon}}\bigg\Vert^2\leq\frac{\beta_1\rho^2}{L\epsilon(1-\beta_1)}\sum_{l=1}^{L-1}l\left(1-(\beta_1)^l\right)\left\Vert\boldsymbol{M}^t\right\Vert^2\nonumber\\&+\frac{\rho^2G^2d}{L\epsilon}\sum_{l=1}^{L-1}l\Bigg(l-\frac{\beta_1\big(1-(\beta_1)^l\big)}{1-\beta_1}\Bigg)\leq\frac{\rho^2\beta_1G^2d}{L\epsilon}\left(-\frac{(\beta_1)^{L-1}}{1-\frac{1}{\beta_1}}\left(-\frac{1}{\beta_1}\right.\right.\nonumber\\&\times\frac{1-\frac{1}{(\beta_1)^{L-1}}}{1-{1}/{\beta_1}}+L-1\Bigg)+\frac{L(L-1)}{2}\Bigg)+\frac{\rho^2G^2d}{L\epsilon}\left(\frac{(L-1)L(2L-1)}{6}\right.\nonumber\\&-\frac{\beta_1L(L-1)}{2(1-\beta_1)}-\frac{(\beta_1)^{L-1}}{\left(1-{1}/{\beta_1}\right)^2}\Bigg(L-1+\frac{1-\frac{1}{(\beta_1)^{L-1}}}{1-\beta_1}\Bigg)\Bigg)\nonumber\\&=\frac{\rho^2G^2dL(2L-1)(L-1)}{6L\epsilon}.
\end{alignat}\end{small}Substituting (\ref{equ1:theorem2}), (\ref{equ2:theorem2}), (\ref{equ3:theorem2}), (\ref{equ4:theorem2}), (\ref{equ5:theorem2}) and (\ref{equ6:theorem2}) to (\ref{equ0:theorem2}) yields\begin{small}\begin{alignat}{1}\label{68}&F\left(\boldsymbol{W}^{t+1}\right)-F\left(\boldsymbol{W}^t\right)\leq\left(\left(\eta\rho+2\right)\left(1-\frac{k}{d}\right)+\eta\rho-1\right)\frac{\eta ^2G^2dL^2}{\epsilon}-\frac{\eta}{2}\nonumber\\&\times\left\Vert\nabla F(\boldsymbol{W}^t)\right\Vert^2+3\eta G^2d\left(\frac{G^4dL}{4\epsilon^3}\left(L-\frac{\beta_2\big(1-(\beta_2)^L\big)}{1-\beta_2}\right)+\frac{L^2}{\epsilon}+\frac{\rho^2L^2}{3\epsilon}\right.\nonumber\\&\left.+1+\frac{4\beta_1\big(1-(\beta_1)^L\big)}{\epsilon\left(1-\beta_1\right)^2}\right)+3\eta\sum_{n=1}^N{\tilde{D}_n\left(\frac{\sigma_l}{\sqrt{\tilde{D}_n}}+\sigma_g\right)^2}/{\sum_{n=1}^N\tilde{D}_n},
\end{alignat}\end{small}By summing up (\ref{68}) from $t=0$ to $T-1$, we have (27). This concludes the proof of \textbf{Theorem} 2.

\section{Proof of Theorem 3}
From (\ref{68}) and applying the P$\L$ condition, we have\begin{small}\begin{alignat}{1}\label{69}&F\left(\boldsymbol{W}^{t+1}\right)-F(\boldsymbol{w}^*)\leq\left(1-\eta\mu\right)\left(F(\boldsymbol{W}^t)-F(\boldsymbol{w}^*)\right)+\left((\eta\rho+2)\left(1-\alpha\right)\right.\nonumber\\&\left.+\eta\rho-1\right)\frac{\eta ^2G^2dL^2}{\epsilon}+3\eta G^2d\left(\frac{G^4dL}{4\epsilon^3}\Bigg(L-\frac{\beta_2\big(1-(\beta_2)^L\big)}{1-\beta_2}\Bigg)\!+\!\frac{L^2}{\epsilon}+1\right.\nonumber\\&\left.+\frac{\rho^2L^2}{3\epsilon}+\frac{4\beta_1\big(1-(\beta_1)^L\big)}{\epsilon\left(1-\beta_1\right)^2}\right)\!+\!3\eta\!\sum_{n=1}^N{\tilde{D}_n\Big(\frac{\sigma_l}{\sqrt{\tilde{D}_n}}+\sigma_g\Big)^2}/{\sum_{n=1}^N\tilde{D}_n}.\!\!
\end{alignat}\end{small}Solving the recurrence over $T$ communication rounds yields\begin{small}\begin{alignat}{1}&F\left(\boldsymbol{W}^{T}\right)\!-\!F(\boldsymbol{w}^*)\!\leq\!\left(1\!-\!\eta\mu\right)^T\big(F(\boldsymbol{W}^0)\!-\!F(\boldsymbol{w}^*)\big)\!+\!\left(\left((\eta\rho\!+\!2)\left(1\!-\!\alpha\right)\!+\!\eta\rho\right.\right.\nonumber\\&\left.\!-1\right)\frac{\eta ^2G^2dL^2}{\epsilon}\!+\!3\eta G^2d\left(\frac{G^4dL}{4\epsilon^3}\bigg(L\!-\!\frac{\beta_2\big(1\!-\!(\beta_2)^L\big)}{1\!-\!\beta_2}\bigg)\!+\!\frac{L^2}{\epsilon}\!+\!1+\!\frac{\rho^2L^2}{3\epsilon}\right.\nonumber\\&\left.+\frac{4\beta_1\big(1\!-\!(\beta_1)^L\big)}{\epsilon\big(1\!-\!\beta_1\big)^2}\right)\!+\!3\eta\sum_{n=1}^N\!\frac{\tilde{D}_n\Big(\frac{\sigma_l}{\sqrt{\tilde{D}_n}}+\sigma_g\Big)^2}{\sum_{n=1}^N\tilde{D}_n}\Bigg)\left(\sum_{t=0}^{T-1}\left(1-\eta\mu\right)^t\right).\!\!\!\!
\end{alignat}\end{small}This concludes the proof of \textbf{Theorem} 3.

\section{Proof of Proposition 3}
Let $\eta=\frac{\eta_0\ln T}{TL^2}$. Given $T\to\infty$, $\mu\eta\to0$. Thus, we have\begin{small}\begin{alignat}{1}\log\left(\left(1-\mu\eta\right)^T\right)=T\log\left(1-\mu\eta\right)\approx -T\mu\eta
\end{alignat}\end{small}Then, it follows that\begin{small}\begin{alignat}{1}&F\left(\boldsymbol{W}^T\right)-F(\boldsymbol{w}^*)=\widetilde{\mathcal{O}}\left(\exp\left(-T\mu\eta\right)\left(F(\boldsymbol{W}^0)-F(\boldsymbol{w}^*)\right)\right)+\mathcal{O}\Big(\!\left((2-\alpha)\right)\nonumber\\&\times\frac{\eta^2L^2\rho G^2d}{\mu\epsilon}\Big)+\mathcal{O}\left((1-\alpha)\frac{\eta L^2G^2d}{\mu\epsilon}\right)=\widetilde{\mathcal{O}}\left(\exp\left(-\frac{\mu\ln(T)}{L^2}\right)\left(F(\boldsymbol{W}^0)\right.\right.\nonumber\\&\left.\left.-F(\boldsymbol{w}^*)\right)\right)+\widetilde{\mathcal{O}}\left(\left((2-\alpha)\right)\frac{\rho G^2d}{\mu\epsilon L^2T^2}\right)+\widetilde{\mathcal{O}}\left((1-\alpha)\frac{G^2d}{\mu\epsilon T}\right).
\end{alignat}\end{small}With $\exp\left(-\frac{\mu\ln(T)}{L^2}\right)=\mathcal{O}\left(\frac{1}{T}\right)$, we obtain (32). This concludes the proof of \textbf{Proposition} 3.

\vfill

\end{document}